\documentclass[10pt,twocolumn,twoside]{IEEEtran}

\usepackage{romannum}
\usepackage{amssymb}
\usepackage{amsmath,amsthm}
\usepackage{graphicx}
\usepackage{dblfloatfix} 
\usepackage{multicol}
\usepackage{lipsum}
\usepackage{algorithm}
\usepackage{algpseudocode}
\usepackage{booktabs}
\usepackage{tabularx}
\usepackage{ragged2e}
\usepackage{csquotes}
\usepackage{helvet}
\usepackage{textcomp}
\usepackage[T1]{fontenc}
\usepackage{bm}
\usepackage{verbatim} 
\usepackage{mathtools}
\usepackage{caption}
\usepackage{graphicx} 
\usepackage{threeparttable}
\usepackage{textcomp}
\usepackage{commath}
\usepackage{float}
\usepackage{svg}
\usepackage{tabularx}
\usepackage{amsmath}
\usepackage{color}
\usepackage{bm}
\usepackage{mathrsfs,float} 
\usepackage{epstopdf}
\usepackage{float}
\usepackage{gensymb,stmaryrd}
\usepackage{algorithmicx}
 \usepackage{enumerate}
 \usepackage{enumitem}
  \newcolumntype{P}[1]{>{\centering\arraybackslash}p{#1}}
  \newcolumntype{M}[1]{>{\centering\arraybackslash}m{#1}}

\usepackage{hyperref}
\usepackage{textcomp}
\usepackage[normalem]{ulem}
\usepackage{xr}

\usepackage{romannum}

\usepackage{color}
\definecolor{lightgray}{gray}{0.95}

\usepackage{hyperref}

\usepackage{xr}
\makeatletter

\newcommand*{\addFileDependency}[1]{
\typeout{(#1)}
\@addtofilelist{#1}
\IfFileExists{#1}{}{\typeout{No file #1.}}
}\makeatother

\newcommand*{\myexternaldocument}[1]{%
\externaldocument{#1}%
\addFileDependency{#1.tex}%
\addFileDependency{#1.aux}%
}

\myexternaldocument{Supplementary}

\usepackage{booktabs, multirow} 
\usepackage{soul}

\usepackage{caption}
\usepackage{subcaption}
\usepackage{cite}

\ifCLASSINFOpdf
\else
\fi
\usepackage{array}
\usepackage{fixltx2e}
\usepackage{url}

\begin{document}
%
\title{RAMAN: A Re-configurable and Sparse tinyML Accelerator for Inference on Edge}

%

\author{Adithya Krishna, Srikanth Rohit Nudurupati, Chandana D G, Pritesh Dwivedi,  Andr\'e van Schaik, Mahesh Mehendale and Chetan Singh Thakur* 
\thanks{A. Krishna, S. R. Nudurupati, D G Chandana, P. Dwivedi, M. Mehendale and C. S. Thakur (Email: csthakur@iisc.ac.in) are with the Department of Electronic Systems Engineering, Indian Institute of Science, Bangalore - 560012, India;  and A. van Schaik is with the International Centre for Neuromorphic Systems, The MARCS Institute, Western Sydney University, Australia. A. Krishna is currently with the International Centre for Neuromorphic Systems, The MARCS Institute, Western Sydney University, Australia.
\newline  This work was supported by the Department of Science and Technology of India under Grant DST/IMP/2018/000550 and Pratiksha Trust Grant FG/SMCH-22-2106.
\newline *Corresponding author}} 
\maketitle

\begin{abstract}
    Deep Neural Network (DNN) based inference at the edge is challenging as these compute, and data-intensive algorithms need to be implemented at low cost and low power while meeting the latency constraints of the target applications. Sparsity, in both activations and weights inherent to DNNs, is a key knob to leverage. In this paper, we present RAMAN, a \textbf{\emph{R}}e-configurable and sp\textbf{\emph{A}}rse {tiny\textbf{\emph{M}}L} \textbf{\emph{A}}ccelerator for infere\textbf{\emph{N}}ce on edge, architected to exploit the sparsity to reduce area (storage), power as well as latency. RAMAN can be configured to support a wide range of DNN topologies - consisting of different convolution layer types and a range of layer parameters (feature-map size and the number of channels). RAMAN can also be configured to support accuracy vs. power/latency tradeoffs using techniques deployed at compile-time and run-time. We present the salient features of the architecture, provide implementation results and compare the same with the state-of-the-art. RAMAN employs novel dataflow inspired by Gustavson’s algorithm that has optimal input activation (IA) and output activation (OA) reuse to minimize memory access and the overall data movement cost. The dataflow allows RAMAN to locally reduce the partial sum (Psum) within a processing element array to eliminate the Psum writeback traffic. Additionally, we suggest a method to reduce peak activation memory by overlapping IA and OA on the same memory space, which can reduce storage requirements by up to 50\%. RAMAN was implemented on a low-power and resource-constrained Efinix Ti60 FPGA with 37.2K LUTs and 8.6K register utilization. RAMAN processes all layers of the MobileNetV1 model at 98.47 GOp/s/W and the DS-CNN model at 79.68 GOp/s/W by leveraging both weight and activation sparsity.

\end{abstract}


\section{Introduction}
Deep neural networks (DNNs) have become ubiquitous in various cognition and learning problems \cite{Deng2010,Krizhevsky2012,Deng2014,Lecun2015}. DNNs are often computed in the cloud, and the inferred result is delivered back to an edge node, introducing delay owing to constrained communication bandwidth. Thus, the deployment of DNNs directly on edge has recently attracted more attention since it offers many inherent benefits, including privacy, bandwidth savings, and latency reductions. However, the computation on an edge device poses numerous challenges due to power, memory, and resource constraints. GPUs and CPUs conventionally used in cloud platforms are extremely power intensive and area inefficient and thus cannot be directly deployed on edge. A promising avenue to pursue in this respect is the development of an accelerator tailored for neural computations on edge. A customized hardware design for neural networks provides an opportunity to optimize dataflow, memory access and exploit network sparsity to overcome edge computing bottlenecks.  
\par Sparsity is an inherent attribute in most DNNs which can be leveraged on hardware. It is estimated that approximately $40\%$ of the input activations (IAs) and $50\%$ of weights (Ws) are sparse in the MobileNet model \cite{Howard2017,Sandler2019} trained on the Imagenet dataset \cite{Deng2010} with the hardware-aware pruning strategy presented in this paper. Other well-established networks like AlexNet \cite{Krizhevsky2012}, VGG-16 \cite{SimonyanVGG16} , and ResNet-50 \cite{HeResnet50} show similar sparsity statistics. Thus sparsity induces a lot of ineffectual zero computations that can be skipped. Aggressive pruning strategies can further reduce computations if a slight reduction in inference accuracy is acceptable. In this direction, several attempts have been made in the literature to maximize the sparsity by zeroing out low-magnitude weights \cite{Han2016,Ding2018,Yao2019,Tanaka2020}. In addition to weight sparsity, the commonly used rectified linear unit (ReLU) activation function clamps all negative activation values to zero, resulting in sparse output activations (OAs), which become IAs to the subsequent layer. Even though exploiting weight sparsity in hardware has been thoroughly investigated, leveraging activation sparsity in hardware efficiently is a topic of research and needs further exploration. This disparity is primarily because of the fact that it is possible to enforce structured sparsity in weights during training by pruning in a hardware-aware fashion (by knowing the underlying hardware architecture and the dataflow) that maximizes the overall hardware utilization and efficiency. However, the activation sparsity is unstructured and highly challenging to leverage on hardware as the data varies dynamically and depends on the environment \cite{Yousefzadeh2021}.

\subsection{Related Work}
Early work in this domain used the indirection principle to exploit sparsity in one of the operands meaning in either weights or activations, but not both. Cnvlutin \cite{Albericio2016} exploits sparsity in IA by storing them in a compressed format as value and index pairs. The index information of the non-zero activations is used to perform in-direct memory access to extract dense weights. In another work, Cambricon-X\cite{Zhang2016Cambricon} uses the same indirection principle, assuming sparse synaptic connections. The input neurons with non-zero synaptic connections are transferred to a computational unit to perform MAC (multiply-accumulate) operations. These architectures are inefficient as they are intended to exploit sparsity present in only one of the operands (W or IA). The dense processing core can quickly adapt to accommodate one operand sparsity by in-direct memory access.
\par EyerissV1 \cite{Chen2017Eyeriss} is one of the early works investigating activation sparsity to save power by employing data-gating logic. When the IA is zero, the data gating logic disables the weight read from a local spad and prevents the MAC datapath from switching. Although this method improves energy efficiency, it does not reduce latency. EyerissV2 \cite{Chen2019Eyeriss} exploits sparsity further in both IA and W by preserving the data in compressed form all the way to the computational element and adopting a row stationary dataflow like EyerissV1. They employ a two-step search technique by retrieving non-zero activations and then utilizing the IA's channel index to fetch non-zero weights, requiring a deeper pipeline and complex processing element (PE) architecture. EyerissV2 employs an external DRAM for storing the activations and parameters and requires a complex hierarchical mesh network to route the data to different computational elements on-chip. We avoid such complexities in RAMAN as we target tinyML edge applications with all on-chip memory implementation.

\par Computer architects face two significant challenges in designing a sparse neural network accelerator leveraging both IA and W sparsity. First is the front-end challenge, where a sufficient number of non-zero IA and W pairs stored in a compressed format must be transferred to the computational unit to keep the MAC utilization high. Maintaining high MAC utilization is often tricky because it requires channel index matching to guarantee that the valid W-IA from the same channel is multiplied. Second, the back-end challenge where the partial-sum (Psum) addresses have to be aligned and immediately reduced within the computational element before a writeback. If the addresses are not aligned, then the Psums cannot be reduced, leading to significant writeback traffic and access contention. SCNN \cite{Parashar2017SCNN} and Sticker\cite{Yuan2020STICKER} try to overcome the front-end challenge; however, these works are plagued by the back-end problem. SCNN \cite{Parashar2017SCNN} uses channel-last dataflow to maximize the multiplier utilization at the cost of high output traffic and access contention. In the channel-last dataflow, the non-zero W and IA values are first organized in the pixel dimension, followed by the channel dimension in memory. Since any non-zero IA can be multiplied with any W in the same channel, it's easy to form non-zero W-IA pairs for computation resulting in high MAC utilization. However, this approach makes the Psum reduction (accumulation) highly challenging. As a result, the output writeback traffic is very high and requires a crossbar switch to arbitrate the data movement. Sticker \cite{Yuan2020STICKER} also uses channel-last dataflow and adopts a two-way set associative PEs to reduce memory access contention and collisions. However, since this strategy requires data re-ordering to prevent collisions, which are done offline using the CPU, it ultimately defeats the whole purpose of latency and energy reductions that sparsity offers. SNAP \cite{Zhang2021} is one of the latest DNN accelerators that tries to overcome the back-end problem by employing channel-first dataflow. Non-zero W and IA data are organized and processed in the channel dimension first and subsequently in the pixel dimension in the channel-first dataflow. This ensures that the Psums calculated by the multiplier array are locally reduced before writeback, thus decreasing the writeback traffic. However, pairs of non-zero IA and W of matching channels have to be extracted, and they utilize address matching units (AMU) and sequence decoders to do the same. The AMU uses $N\times N$ comparators to match W and IA channel indices, making it highly inefficient as the area and power grow quadratically with N. EIE\cite{Han2016EIE} exploits both IA and W sparsity but only supports fully connected (FC) layer making it incompatible to run convolution operations. 
\par In addition, several FPGA-based implementations have been proposed in the literature \cite{Aimar2019, Shah2018, McDanel2019, Zhang2021mdpi, Nguyen2019, Ma2018, Yu2020}. NullHop\cite{Aimar2019} exploits the activation sparsity of the neurons in CNNs to accelerate the computation and reduce storage. However, this work does not exploit weight sparsity. Shah et al. \cite{Shah2018} propose a run-time programmable coprocessor architecture to accelerate CNNs. However, programmability is limited to the standard convolution-type networks, and the sparsity optimizations have not been studied. McDanel et al. \cite{McDanel2019} propose a full-stack optimization framework to speed up CNNs by leveraging weight sparsity by column-combine strategy and reducing power by clock gating when activations are zero. Zhang et al. \cite{Zhang2021mdpi} and Nguyen et al. \cite{Nguyen2019} propose CNN accelerators for YOLOv2 networks, and their architectures are hardwired to support a particular topology without any programming flexibility. Ma et al. \cite{Ma2018}, and Yu et al. \cite{Yu2020} provide FPGA CNN implementations without exploiting sparsity and have limited programming capability.

\subsection{Our Contributions}
\par This work presents a re-configurable and sparse deep neural network accelerator that exploits both IA and W sparsity. To address the front-end challenge, we employ Gustavson's inspired dataflow \cite{Zhang2021Gamma} with the hardware-aware balanced weight pruning strategy to keep workload uniform across all the PEs and maintain high MAC utilization. The back-end issue is resolved by reducing Psum locally within a processing element (PE) array; this reduces the writeback bandwidth and eliminates memory access contention because only the final result is sent back to memory rather than every intermediate result. 

\par 
Latest advancements in quantization techniques \cite{Jacob2017,Yao2021HAWQV3} in DNNs have eliminated the need for floating-point arithmetic (during inference), and simple energy-efficient fixed-point arithmetic has proven adequate to achieve reasonable accuracy. Despite having a highly efficient computational realm, today's design has a memory bottleneck which is evident from \cite{Horowitz2014}. Memory access (especially DRAM) is orders of magnitude more expensive than conventional arithmetic operations. The most state-of-the-art DNN accelerators, except for EIE, utilize an external DRAM to store IAs and Ws and pay the penalty of unprecedented memory access cost. Since our architecture is targeted at tinyML edge computing applications with stringent energy budgets, using external DRAM and paying high energy costs becomes untenable. Thus, in this work, we do away with the requirement for a DRAM and only use the on-chip SRAM to store the model weights and its activations. However, the memory size of the on-chip SRAM is limited due to area constraints, which necessitates model optimizations to fit modern networks such as MobileNets \cite{Howard2017,Sandler2019} in SRAM. In this work, we introduce a hardware-aware pruning strategy to shrink the model size and intelligent memory scheduling to minimize peak activation memory to accommodate both model and activations on-chip.  
\par Furthermore, prior works have limited flexibility regarding layers supported, and unsupported layers are often executed offline, making the overall system-level computation inefficient. This work supports a wide range of layer computations from the conventional CNN networks to the modern depth-wise separable convolutions constituting depth-wise (DW) and point-wise (PW) layers. Our design also allows max pooling, average pooling, fully connected layers, and residual additions on-chip, eliminating the need to offload work to an external CPU. In summary, the following are the contributions and features of the RAMAN architecture presented in this paper:
\begin{itemize}[leftmargin=*]
\item \textbf{Memory Hierarchy:}  RAMAN employs a three-level memory hierarchy comprising low-cost levels such as reg-file (RF) and cache to amortize the data access cost to the high-cost on-chip global memory (GLB-MEM) (c.f. Section \ref{sec:glb_mem}). The reg-file is used for Psum reduction locally inside the PE to overcome the back-end challenges.
\item\textbf{Sparsity:} RAMAN exploits both IA and W sparsity to achieve higher throughput and energy efficiency. Sparsity is leveraged in storage, computation and data movement. Input activations are stored in compressed form in a cache, and a simplified version of the compressed sparse row (CSR) format \cite{Strout2004} is used for storing weights in both on-chip global memory and cache. Since RAMAN adopts a hardware-aware balanced pruning technique, it is not necessary to keep bounds/row index pointers, and thus we have further optimized the CSR format based on our requirements to minimize W storage (c.f. Section \Romannum{5} of the supplementary document). In computation, sparsity is leveraged in two ways; for the layers with the maximum computational density (such as PW), RAMAN can skip the processing cycles with zero data, improving throughput and energy efficiency. However, skipping processing cycles won't reap any benefit in the layers with relatively low computational density (such as DW); in such cases, the design only data-gates the cycles with zero data but does not skip them. This strategy minimizes architectural complexity and improves overall energy efficiency. Finally, sparsity is utilized in data movement as only non-zero data is fetched and propagated in the network, lowering memory bandwidth and increasing energy efficiency even further.
\item \textbf{Programmability:} RAMAN supports both traditional CNN models and modern separable convolutional models that constitute depth-wise and point-wise layers. In addition, it supports the execution of max pooling, average pooling, and fully connected layers. To enable this model and layer flexibility, we propose a network-on-chip (NoC) that can adapt to a wide range of bandwidth requirements. It can provide a high bandwidth IA data from the cache to keep the PEs busy when there is limited IA reuse (for DW layers); when the IA reuse is high (in PW and FC layers), it reduces IA data bandwidth and increases W bandwidth. In addition, RAMAN offers an instruction memory and a dedicated instruction set to store and program different network topologies. 
\item \textbf{Dataflow:} RAMAN incorporates novel dataflow inspired by Gustavson's algorithm \cite{Zhang2021Gamma} to reduce memory access for the PW layer, which is the most computationally intensive layer. Additionally, the NoC is reconfigured to support weight stationary dataflow for DW and standard convolutional (CONV) layers. This hybrid dataflow architecture, enabled by dynamic NoC reconfiguration, maps the computation of a particular layer to its optimum dataflow to achieve maximum energy efficiency and minimum data movement cost.
\item \textbf{Peak-Memory Reduction} The state-of-the-art accelerators logically partition the IAs and OAs inside the memory, where the total activation memory is the sum of IA and OA memory spaces. This logical partition is eliminated in our work, and the OAs are directly overwritten onto the IA memory space, lowering the peak activation memory requirements. RAMAN employs an intelligent memory scheduling scheme presented in Section \ref{sec:peak_memory_reduction} to prevent memory collision issues that ensue after removing the logical partition.  
\item \textbf{Run-time Activation Pruning (RAP):} RAMAN performs a hardware-aware activation pruning at run-time to increase the activation sparsity, and the architecture effectively leverages this strategy to increase throughput and energy efficiency. We believe this is the first attempt of its kind, and our offline experiments demonstrate the network's resilience to such pruning strategies. Section \ref{sec:ase} provides a detailed description of RAP.
\item \textbf{Hardware-aware balanced weight pruning:} We propose a hardware-aware balanced weight pruning strategy that reduces memory storage, access, and processing latency. The hardware-aware pruning is an example of software-hardware co-optimization, where a DNN is pruned during the training process, considering the underlying hardware architecture. By ensuring uniform zero/non-zero weight distribution across the weight tiles, the balanced pruning methodology solves the issue of workload imbalance. Without this strategy, non-zero weights would be unevenly distributed over different weight tiles processed by different PEs, resulting in workload imbalance, and the overall performance would be limited by the PE with the heaviest workload. The PEs with low non-zero weight tile distribution completes their execution sooner and must be stalled for the slower ones (the PEs with high non-zero weight tile distribution).


\item \textbf{Flexible quantization:} RAMAN supports variable precision quantization of Ws and IAs supporting 2b, 4b, and 8b precisions. In addition, the architecture is programmable such that the precision of individual layers can vary to meet the required accuracy, storage, and latency target.
\end{itemize}

The rest of the paper is organized as follows: Section \ref{sec:top_level_arch} presents the architecture of the RAMAN accelerator. The architectural features that make RAMAN feasible on edge are highlighted in the Section \ref{sec:features}. Section \ref{sec:impl_results} provides implementation results, and Section \ref{sec:conclusion} concludes this article.

\section{System Architecture}
\label{sec:top_level_arch}
\subsection{Top-Level Architecture} 

Fig. \ref{fig:top_module} shows the top-level architecture of the RAMAN accelerator system. The architecture can be broadly categorized into:

\begin{figure}[h]
	\begin{center}
		\includegraphics[width = \columnwidth]{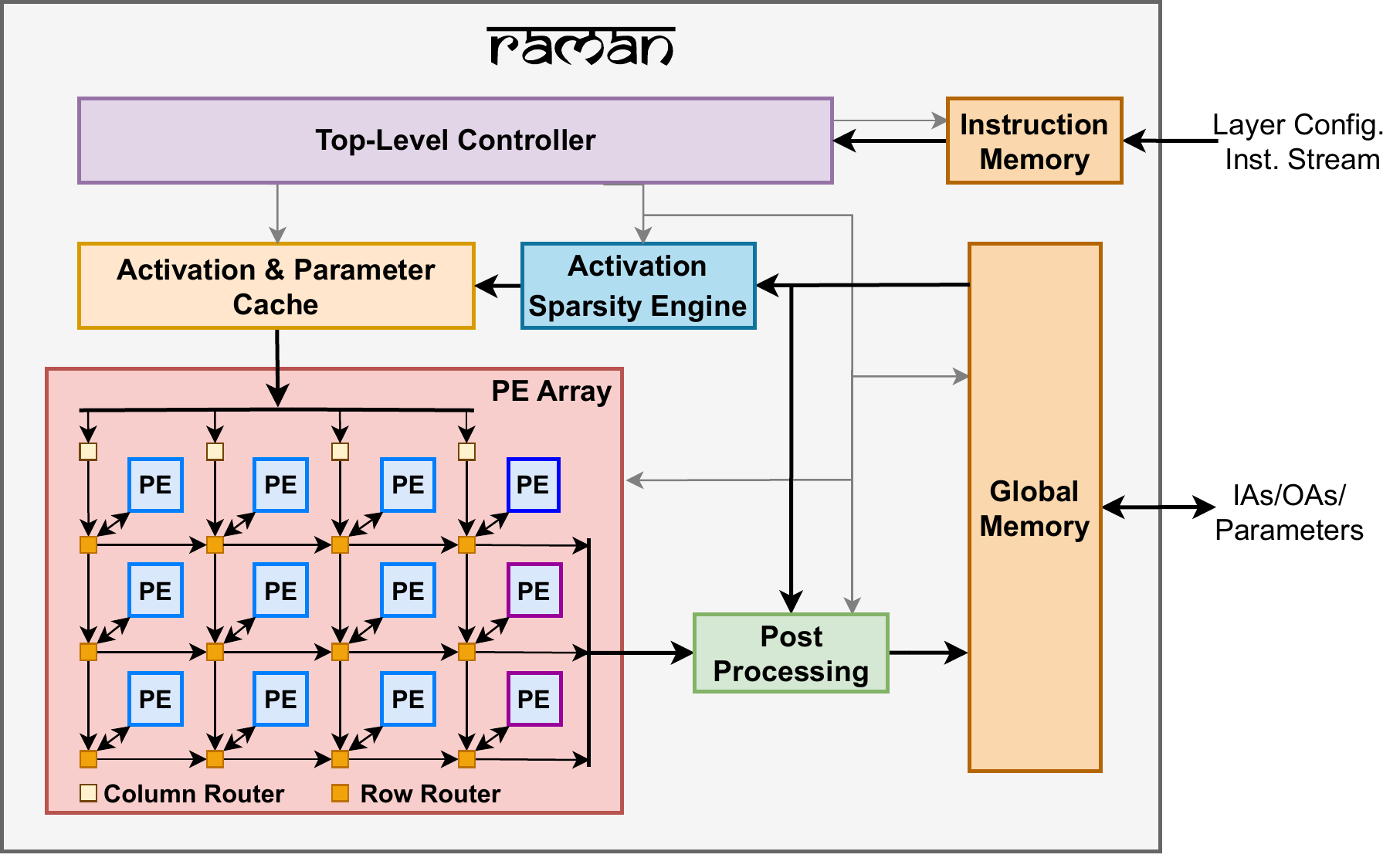}
	\end{center}
	\caption{Top-level architecture.} 
	\label{fig:top_module}
\end{figure}

\begin{itemize}[leftmargin=*]

\item \textbf{Compute:} The compute sub-system comprises a processing element (PE) array, an activation sparsity engine (ASE) and a post-processing module (PPM). The multiply and accumulate (MAC) operations are performed by 12 spatial PEs in the PE array, which are arranged in a 3x4 rectangle. The ASE leverages input activation sparsity to minimize latency and power by skipping ineffectual zero computations. The PPM performs ReLU, quantization, pooling, bias and residual addition operations. Sections \ref{sec:ase}-\ref{sec:ppm} gives a detailed description of the PE array, ASE and post-processing module.
\item \textbf{Memory:} The memory sub-system comprises an on-chip global memory (GLB-MEM), activation and parameter cache, and instruction memory. GLB-MEM stores the parameters of all layers and the input and output activations of a specific layer. Cache exploits temporal reuse in parameters and activations to reduce energy-expensive data access to the large on-chip GLB-MEM. We employ a three-level memory hierarchy composing GLB-MEM, cache and RFs (inside PEs). In addition, we have the instruction memory to store layer configuration instructions of individual layers. A detailed description of the GLB-MEM and cache is provided in Section \ref{sec:glb_mem}.
\item \textbf{Control:} The control sub-system encompasses a top-level controller to coordinate: 1) data transfer between the GLB-MEM and cache; 2) traffic between the cache and PE array utilizing the NoC; 3) traffic between the PE array, PPM and GLB-MEM; 4) operations of the ASE, PE array, NoC, and PPM. We do not use a separate controller for each of the 12 PEs since they are all identical and operate in lockstep, meaning that their processing states are equivalent with regard to one another. The top-level controller is responsible for issuing the control signals to all the 12 PEs.
A detailed description is provided in Section \Romannum{6} of the supplementary document.

\end{itemize}


\begin{figure*}[h]
	\begin{center}
		\includegraphics[width=16cm]{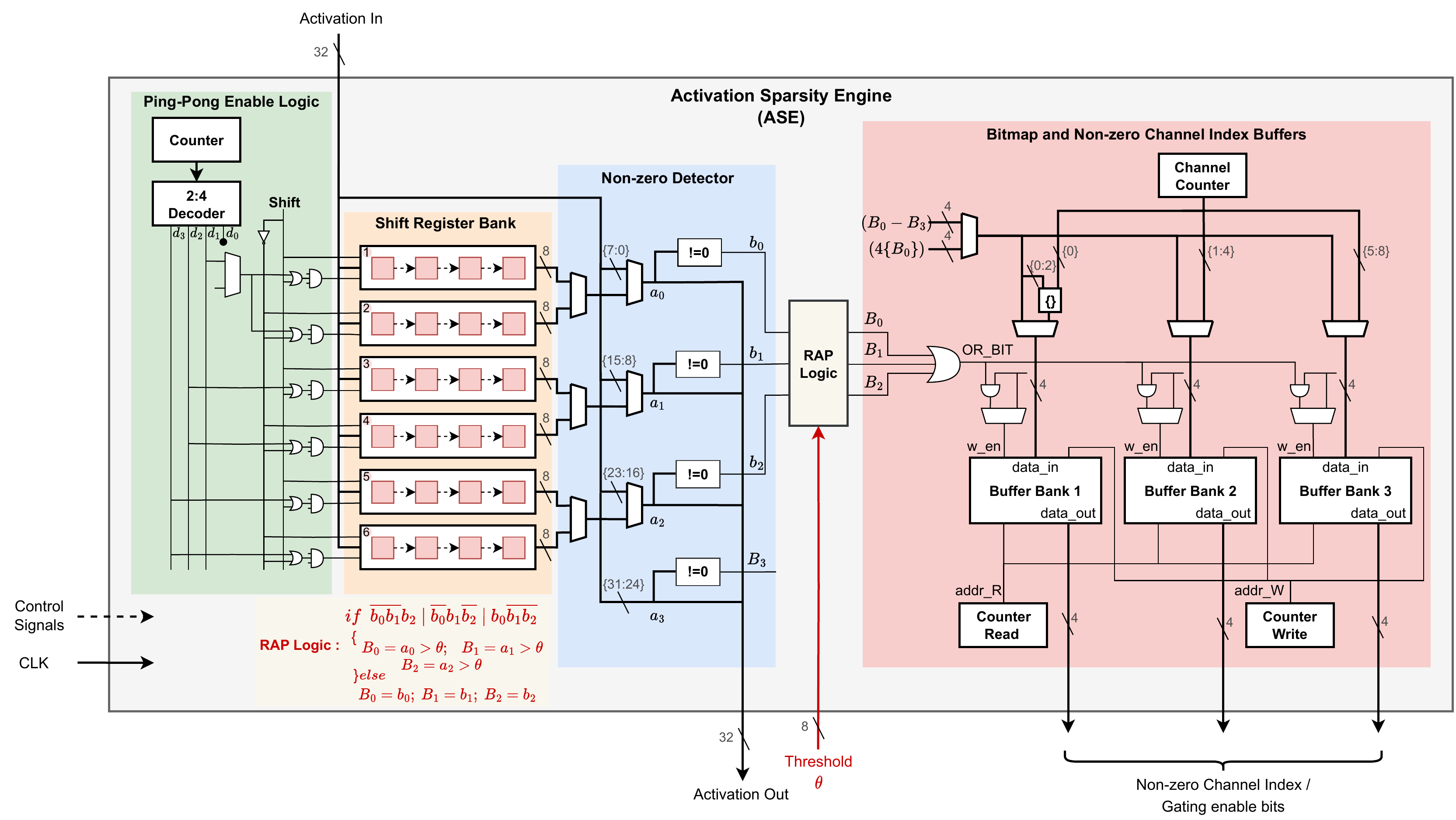}
	\end{center}
	\caption{Activation sparsity engine.} 
	\label{fig:ase}
\end{figure*}




\subsection{Global and cache memory}
\label{sec:glb_mem}
The GLB-MEM comprises of activation and parameter memory banks. We overlay the input and output activations onto the same memory space, as discussed in Section \ref{sec:peak_memory_reduction} to minimize the activation memory storage. The parameter memory is 192b wide with a single port synchronous read/write interface, whereas the activation memory is 32b wide with dual address ports.

\par Cache enables temporal reuse of IA and W, minimizing the global memory accesses, which lower power consumption incurred in accessing global memory. The cache memory is a banked architecture comprising a set of 27 memory banks. Each bank is an 8b wide dual port synchronous read/write memory. Cache banks are dynamically partitioned based on the layer being executed. For instance, three banks are designated for activations in the PW layer, whereas twelve, four, and three banks are set aside for activations in the DW, FC, and CONV layers. W or IA are fetched directly from the GLB-MEM and processed in the PE array skipping cache when reuse is low. The global and cache memory block diagrams are shown in Fig. 8 of the supplementary document.

\subsection{Activation sparsity Engine (ASE)}
\label{sec:ase}
The activation sparsity engine serves two purposes. First, it reduces the cycles needed to write a block of data from the GLB-MEM to cache through ping-pong-based shift registers. Second, it aids in exploiting activation sparsity. Sparsity is leveraged in two ways: 1) by data gating the cycles with zero data and disabling the memory read to prevent the datapath from switching, thereby reducing dynamic power; 2) by skipping the processing cycles entirely to improve energy efficiency and throughput. IAs are routed through the ASE to record the position of zeros, and this information is utilized during computation to either gate or skip zero computation, depending on the layer under execution. The layers with low computing density (e.g., DW) adopt the gating strategy, and the layers with relatively high computational density (e.g., PW) employ the zero skipping technique in addition to data gating. This hybrid approach reduces architectural complexity since imposing zero skipping during DW execution has no positive impact on performance. The design details of the ASE are provided in Section \Romannum{2} of the supplementary document.
\par Area overhead of the ASE in terms of LUTs (lookup tables), registers and memory utilization is insignificant, as demonstrated in Section \ref{sec:impl_results}.


\begin{figure*}[h!]
  \centering
      \begin{subfigure}[b]{1\columnwidth}
    \includegraphics[width=\columnwidth]{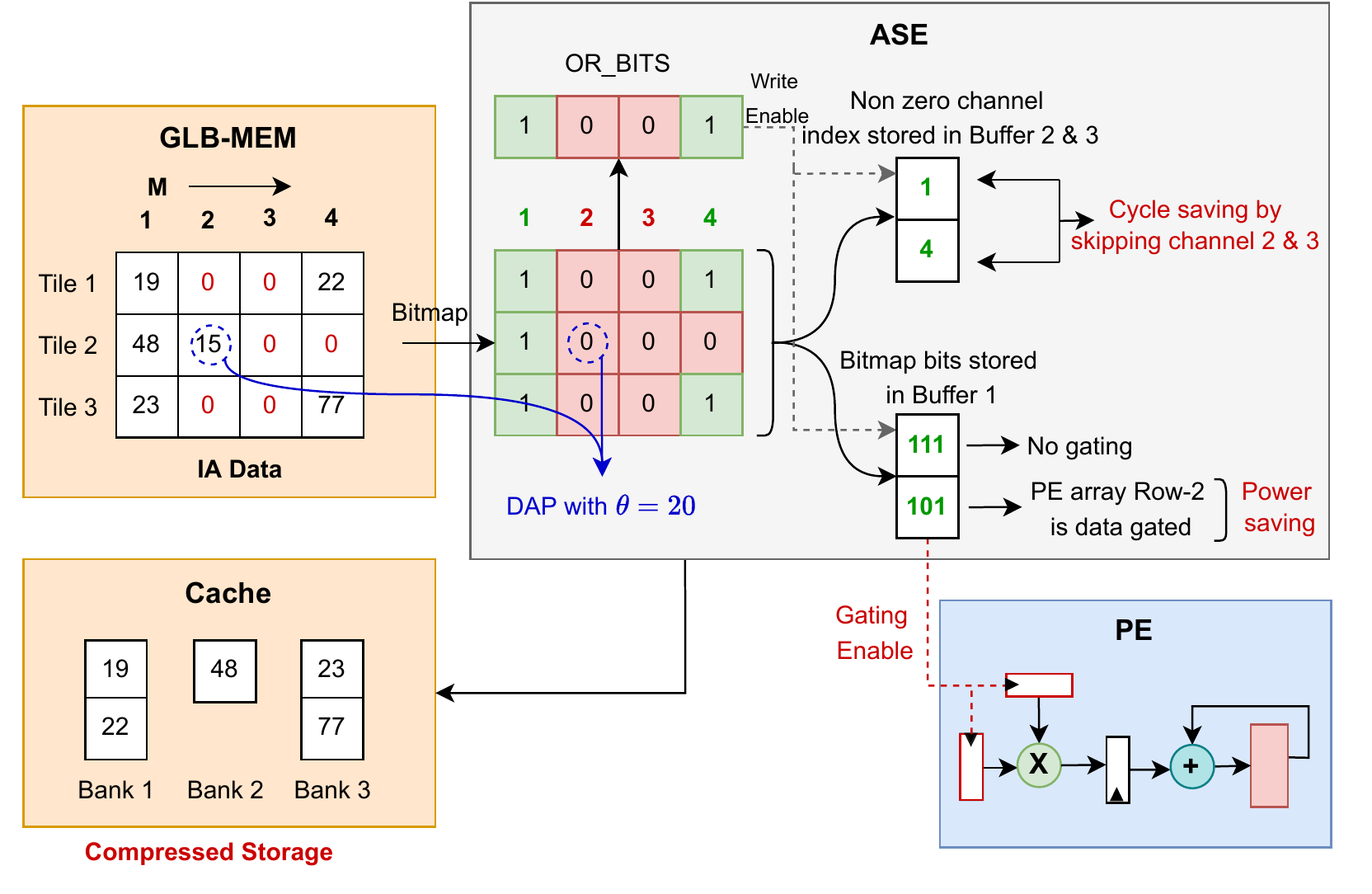}
    \caption{}
  \end{subfigure}
      \begin{subfigure}[b]{1\columnwidth}
    \includegraphics[width=\columnwidth]{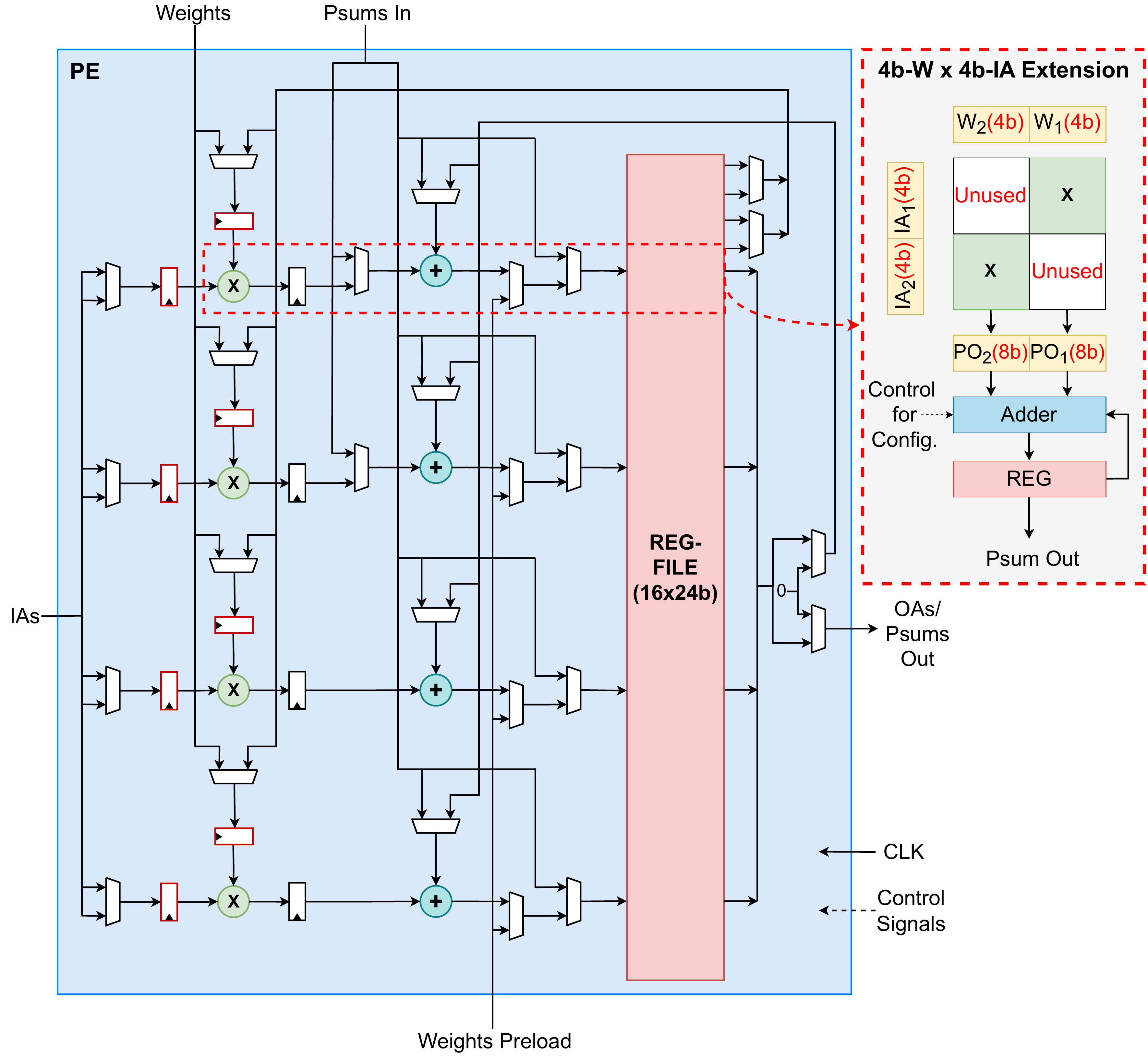}
    \caption{}
  \end{subfigure}
 
  \caption{(a) ASE Illustration for the PW layer. (b) Processing Element (Type-1). }
  \label{fig:pe1}
\end{figure*}


\begin{figure*}[b]
	\begin{center}
		\includegraphics[width = 16cm]{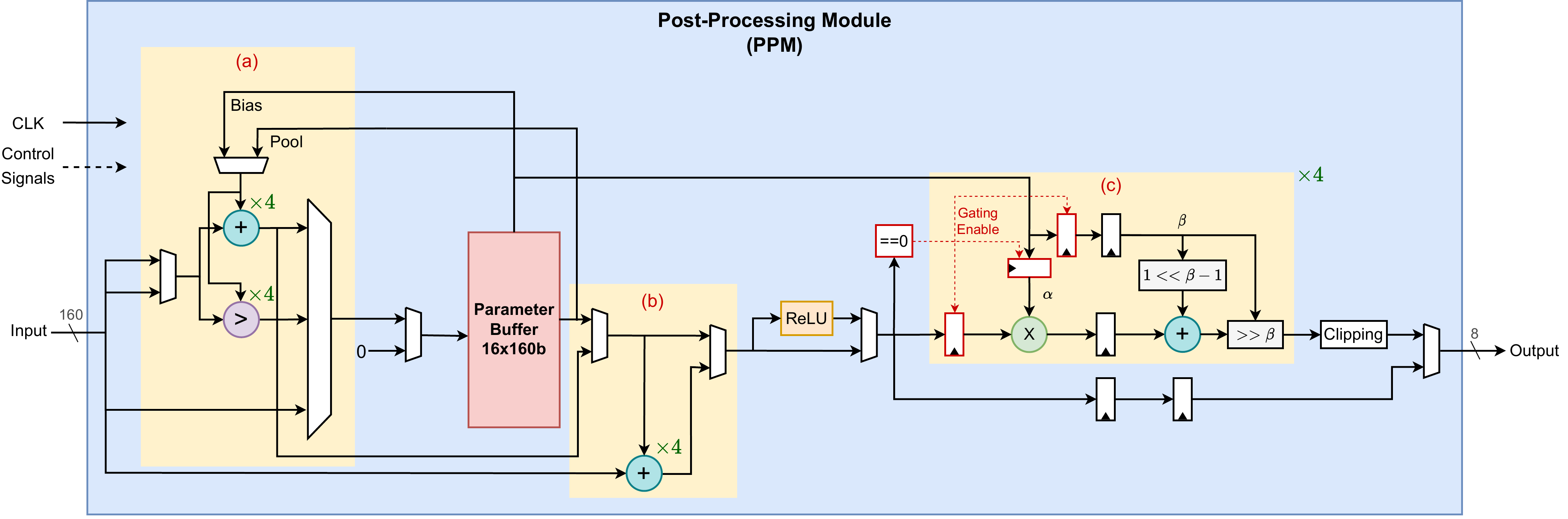}
	\end{center}
	\caption{Post-Processing Module.} 
	\label{fig:ppm}
\end{figure*}

\subsubsection{ASE illustration}

Fig. \ref{fig:pe1}(a) illustrates the working of ASE for the PW layer. The three IA tiles are loaded into the shift register bank, and the non-zero detector module generates the bitmap. In Fig. \ref{fig:pe1}(a), the second input channel of tile-2 is pruned as the value is less than the threshold value of $20$, inside the RAP block.  The bitmap matrix is column-wise ORed to obtain the $OR\_BITS$ vector. The columns ($2$ and $3$) are skipped during computation as all elements are zero post RAP. The channel indices ($1$ and $4$) are cached in buffers $2$ and $3$, and they are subsequently utilized as addresses to retrieve the corresponding input channel weight from the cache during computation. Additionally, the column-wise bitmap is stored in buffer $1$ for non-zero channel indices ($1$ and $4$), which is accessible during computation cycles to activate or disable the PE's data registers as depicted in the figure. In the example shown, for the first input channel, all the rows of the PE array are active as the bitmap is $ '111’$ for that particular column, and in the fourth channel, row-2 of the PE array is deactivated as the bitmap is $ '101’$. Additionally, only the non-zero elements of the IA matrix are saved in cache banks utilizing the bitmap data. Just the bitmap is saved in buffers for other layers, not channel indices, as they only facilitate data gating and not cycle skipping.

\subsection{PE Array}

The PE array comprises 12 processing elements spatially distributed along three rows and four columns. It is coupled with the Network-on-chip to route the data among different PEs.

\subsubsection{Network-on-chip}
\label{sec:noc}
The network-on-chip handles data delivery between the cache, the PE array, and between different PEs. The following are the NoC's responsibilities: 1) Support various data delivery patterns needed by different layers; 2) Provide sufficient data bandwidth for parallel processing to keep the PEs active and improve utilization; 3) Handle various strides and padding, and 4) Leverage spatial data reuse to increase energy efficiency.
\par To accomplish this, we employ a network of row and column routers. The column routers distribute a cached data block across four columns of the PE array, while the row router further splits the incoming packets and delivers input to the PE in a specific row. Furthermore, the output from the PEs is directed to the post-processing module via the row routers. Section \ref{sec:dataflow} goes into greater detail on the various data delivery patterns to perform different layer computations. 

\subsubsection{Processing Element}
\label{pe_dg}
Fig. \ref{fig:pe1}(b) shows the architecture of the PE. We employ three types of PEs to support dataflow flexibility for different layer types. The architectures of the other two PE types constituting the last column of the PE array are presented in Figs. 10, 11 of the supplementary document.   The fundamental elements of all PE configurations are an 8b multiplier, a 24b adder constituting MAC (Multiply and Accumulate Unit), and a reg-file. 8b W and IAs are provided as input and accumulated using the MAC unit, and a 24b Psum is saved in the RF. Offline experiments show that the Psum could fit within the 24b range. The type-2 and type-3 PEs use four-ported RF (4 input \& output ports), while the RF in type-1 PE has four input and eight output ports, each with a width of 24b and a depth of 16. Thus, the RF can be addressed using 4b. The Psums are locally reduced inside the PE array, and the final result is written back to the memory after quantization in PPM, thereby reducing the writeback bandwidth and eliminating memory contention. 
\\ \textbf{Power saving:} PE implements data-gating logic to leverage zeros in the IAs for saving processing power in the layers with low computational density, such as DW. The red-bordered registers in Fig. \ref{fig:pe1}(b) are data-gated, and if a zero IA value is detected (provided by the bitmap bits stored in ASE buffers), then the gating registers are disabled to prevent the MAC datapath from switching. 
\\ \textbf{SIMD support:} PE supports SIMD processing, evident from Fig. \ref{fig:pe1}(b), by performing four MAC operations per cycle, thereby speeding up the processing four times. In addition,  SIMD processing also enables W and IA reuse, thereby reducing the number of memory accesses. RF has four write ports to simultaneously write the Psums from the four MAC units.
\\ \textbf{Variable precision:} The PE datapath supports variable precision Ws and IAs. The reconfiguration of the data path for the 4b-W and 4b-IA using the same 8b datapath is shown in Fig. \ref{fig:pe1} (b). First, two 4b-Ws and 4b-IAs are packed as an input to the 8b multiplier to compute two 4b multiplications doubling the throughput. Next, the accumulator reduces the ensuing partial outputs (PO$_{1}$ and PO$_{2}$), each 8b wide. Similarly, for the 2b-W and 2b-IA case, the multiplier simultaneously does four 2b multiplications boosting throughput by 4x.

\begin{figure*}[h!]
  \centering
      \begin{subfigure}[b]{0.9\textwidth}
    \includegraphics[width=\textwidth]{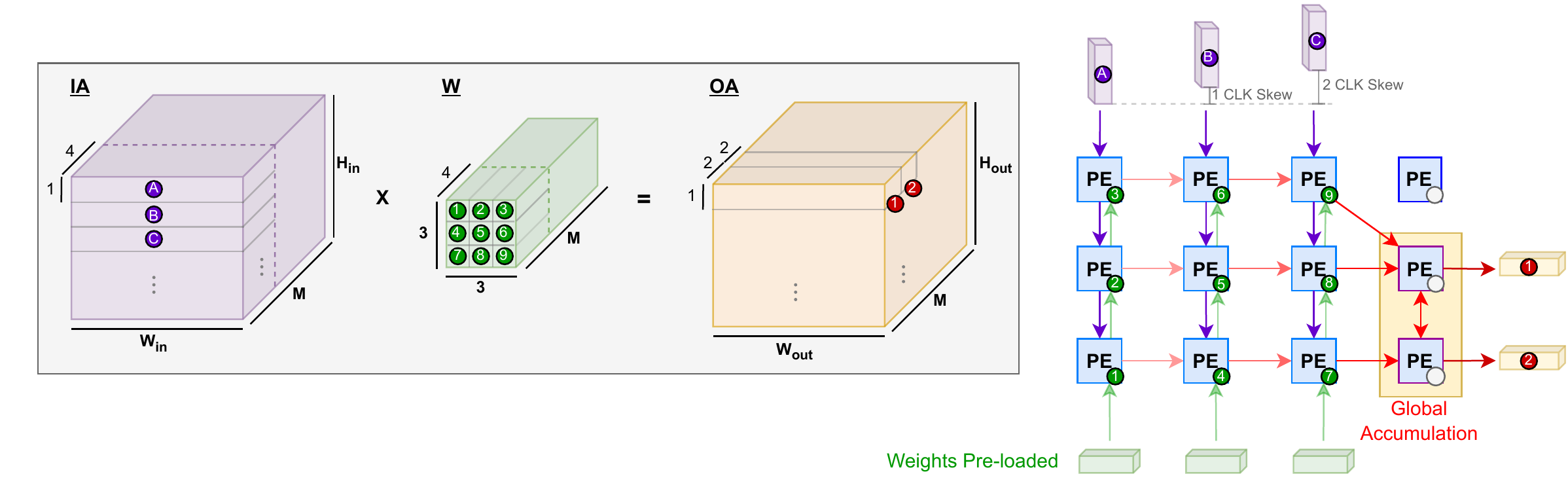}
    \caption{}
  \end{subfigure}
      \begin{subfigure}[b]{0.4\textwidth}
    \includegraphics[width=\textwidth]{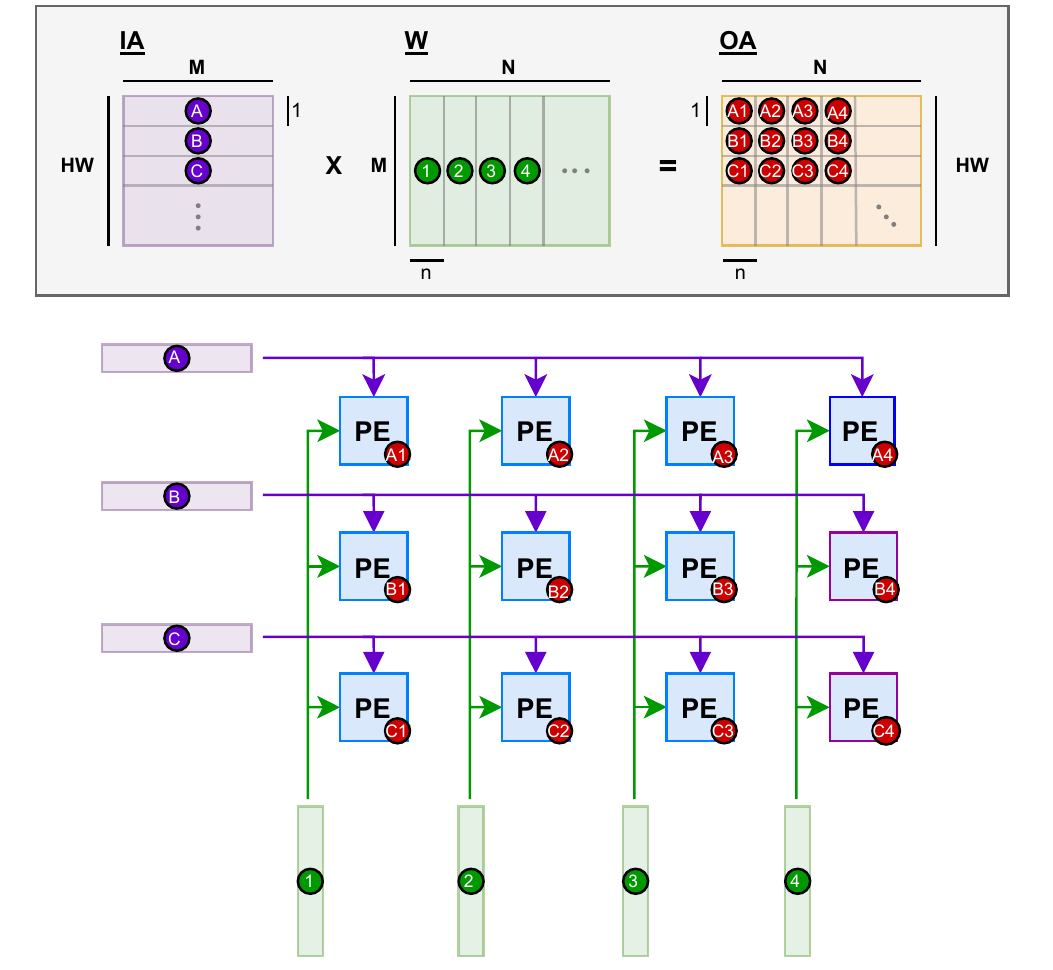}
    \caption{}
  \end{subfigure}
  \begin{subfigure}[b]{0.4\textwidth}
    \includegraphics[width=\textwidth]{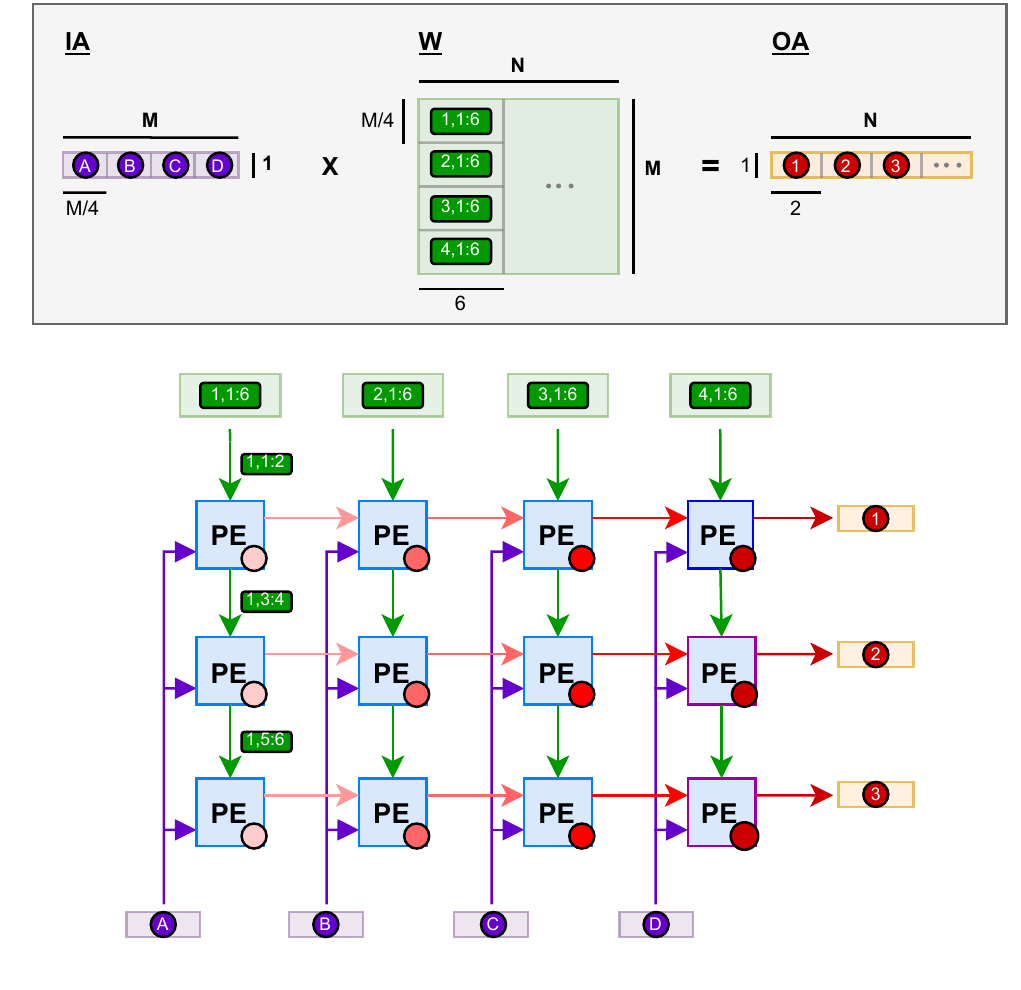}
    \caption{ }
  \end{subfigure}

  \caption{Dataflow configuration across an array of PEs in RAMAN for (a) DW computation (b) PW computation and (c) FC computation. }
  \label{fig:dataflow}
\end{figure*}

\subsection{Post Processing Module}
\label{sec:ppm}
The post-processing module (PPM) is responsible for ReLU activation, bias addition, quantization, residual addition, max pooling, and average pooling operations. The PPM operations can broadly be categorized into memory pre-fetching, computation, and writeback stages. In the pre-fetching phase, the post-processing parameters such as bias ($b$), $\alpha$, and $\beta$ are pre-fetched from the GLB-MEM and cached locally in a 160b wide parameter buffer.  $\alpha$ and $\beta$ represent dyadic scaling parameters computed offline. The pre-fetching is overlapped with the PE array computations to amortize the memory access latency. In the computation phase, the Psum output obtained from the PE array is added with bias in sub-module (a) of Fig. \ref{fig:ppm}. The result is then passed to the ReLU block that clamps the negative values. ReLU is implemented by a sign-bit (MSB) bit comparison. If the MSB bit is $0$, the ReLU block outputs the input; else, it outputs a zero. The output of this block is a ReLU-activated Psum which is then quantized to obtain an 8b representation and written back to the memory. The quantization operation is performed in sub-module (c) of Fig. \ref{fig:ppm}. The implementation detail of the quantization operation is provided in Section \Romannum{3} of the supplementary document.
\par The same hardware can be reused to perform average pooling and max-pooling operations. For average pooling, the adder used for bias addition is re-purposed as an accumulator, and the parameter buffer is re-purposed to store the intermediate accumulation results. Division of the accumulated sum by $HW$ (Height $\times$ Width of the feature map) is performed in the sub-module (c) by expressing $1/HW$ in the dyadic form $\dfrac{\alpha}{2^{\beta}}$. An additional comparator is needed for max-pooling, as shown in sub-module (a). The PPM also supports residual input additions in sub-module (b).
\par PPM uses 4-way SIMD to improve throughput by performing 4 post-processing operations each cycle.

\begin{figure*}[h]
	\begin{center}
		\includegraphics[width =16cm]{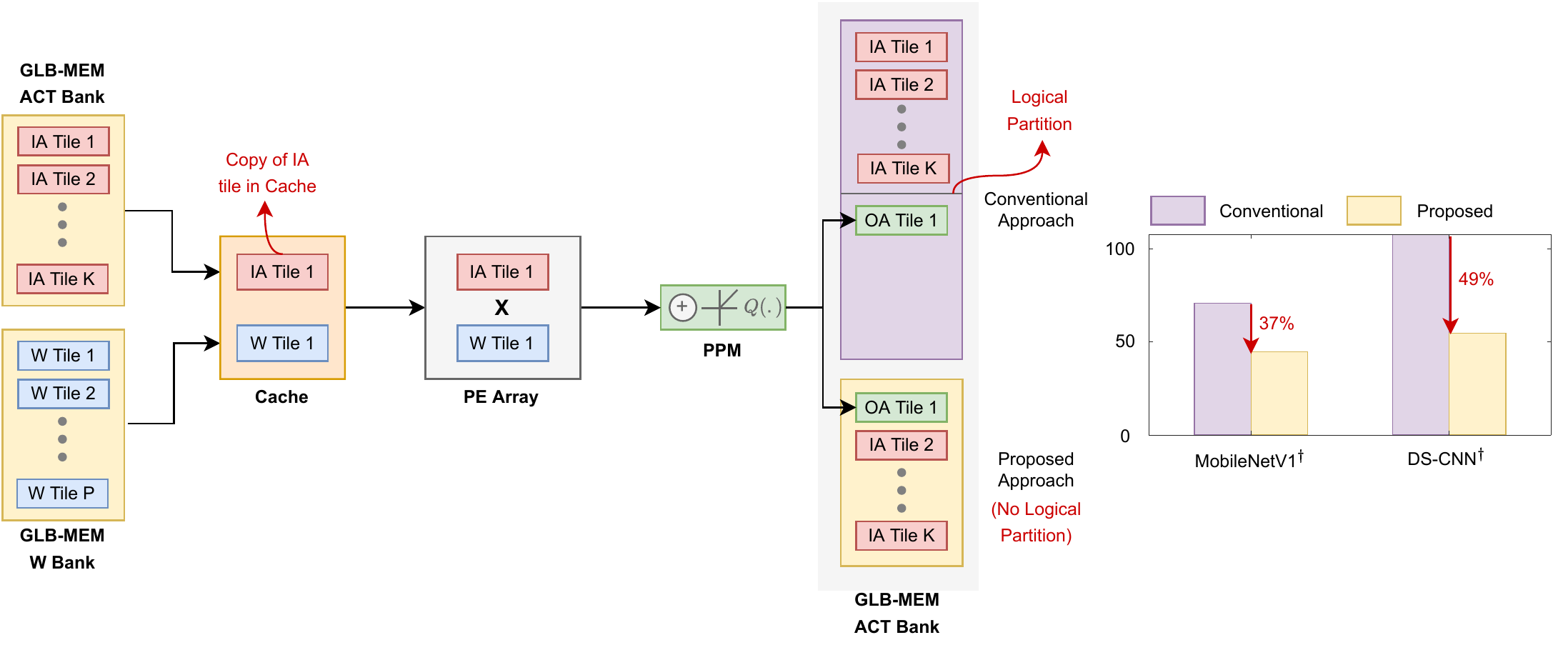}
	\end{center}
	\caption{Illustration of the IA and OA memory space overlapping to reduce peak activation memory.} 
	\label{fig:ip_op_overlapping}
\end{figure*}

\subsection{Dataflow} 
\label{sec:dataflow}

\subsubsection{DW}
DW layer uses a weight stationary systolic dataflow where the W remains static during the computation. IAs of dimension $(H_{in}\times W_{in}\times M)$ are partitioned into tiles of size $(1\times W_{in}\times4)$ and Ws of dimension $(3\times3\times M)$ are partitioned  into tiles of size $(3\times3\times4)$. Then three IA tiles are streamed from the cache to the PE array from top-to-bottom, and the Psum obtained is spatially reduced along the PE columns from left-to-right. A global accumulation is performed by the last column of PEs to obtain a final accumulated result. This mapping allows the reuse of IA along a column and Psums are spatially reduced inside the PE array. Since each PE supports SIMD with four MAC operations per cycle, four input IA channels are convolved with four W channels in every cycle. The systolic data flow necessitates proper synchronization of IA and Psums, as shown in Fig. \ref{fig:dataflow}(a). Each processing run generates a tile of OA of size $(1\times W_{out}\times4)$, and it takes $(H_{out}\times1\times M/4)$ runs to obtain all outputs. The dataflow also allows strided convolution and zero-padding. The CONV layer uses the same dataflow as the DW.

\subsubsection{PW}
The PW layer execution can be represented by 2D matrix multiplication of IA with dimension ($HW \times M$) and weight W with dimension ($M\times N$) to produce OA with dimension ($HW \times N $) as shown in Fig. \ref{fig:dataflow}(b), where M and N are number of input and output channels respectively. The IA matrix is partitioned into $HW$ tiles each sizing ($1\times M$), and the W matrix is partitioned into $N/n$ tiles each sizing ($M\times n$). Since the PE array comprises $3 \times 4$ PEs, three IA tiles and four W tiles are passed to the PE array for computation in a single processing run. An IA tile is broadcasted to four PEs in a row, and a W tile is broadcasted to three PEs in a column. This mapping allows spatial reuse of IA along a row and W along a column. The Psums are locally accumulated and stored in the RF of individual PEs, and the final Psum is transferred to the PPM through row routers of NoC, reducing output writeback traffic to GLB-MEM. The IA and W tiles are sent to the PE array in compressed form for computation. In every processing run, $(3\times 4n)$ tile of OA is generated, and it takes $(HW/3)\times (N/4n)$ runs to construct the entire OA matrix. $n$ depends on the RF depth and is set to $16$ in our implementation. This dataflow ensures the maximum reuse of IA and W with low OA writeback bandwidth.
\subsubsection{FC}
The FC layer can be represented by vector-matrix multiplication, as shown in Fig. \ref{fig:dataflow}(c). The IA is denoted as a vector of size ($1\times M$), and W is denoted as a matrix of size ($M \times N$). The IA vector is divided into four tiles of size ($1\times M/4$), and each tile is broadcasted to three PEs in a column. The W matrix is divided into $4 \times N/6$ tiles of size ($M/4 \times 6$) and is multicasted to three PEs in a column. A single PE receives two weights and a single activation per clock cycle and performs two MAC operations. The Psum reduction is made in two levels: PE-level and core-level. At the PE level, $M/4$ activations and $M/4\times 2$ weights are streamed into each PE, and the resulting Psums are locally accumulated inside PE for $M/4$ cycles. At the core-level, the Psums stored in the RF of each PE are spatially reduced along four columns of the PE array. This two-level Psum reduction reduces the writeback traffic to GLB-MEM. The last column output provides the final result, which is sent to the PPM. Six OAs are generated in every processing run, and $N/6$ runs are required to construct the entire output vector. Each PE only uses two of the four MAC units available due to bandwidth constraints and low-weight reuse in the FC layer.


\section{RAMAN memory optimizations to support deployment at the edge.}
\label{sec:features}
The RAMAN accelerator was developed targeting tinyML edge computing applications, and the following are the features that enable RAMAN to be deployable on edge:

\subsection{Peak activation memory reduction:} 
\label{sec:peak_memory_reduction}

\begin{figure*}[h]
	\begin{center}
		\includegraphics[width = \textwidth]{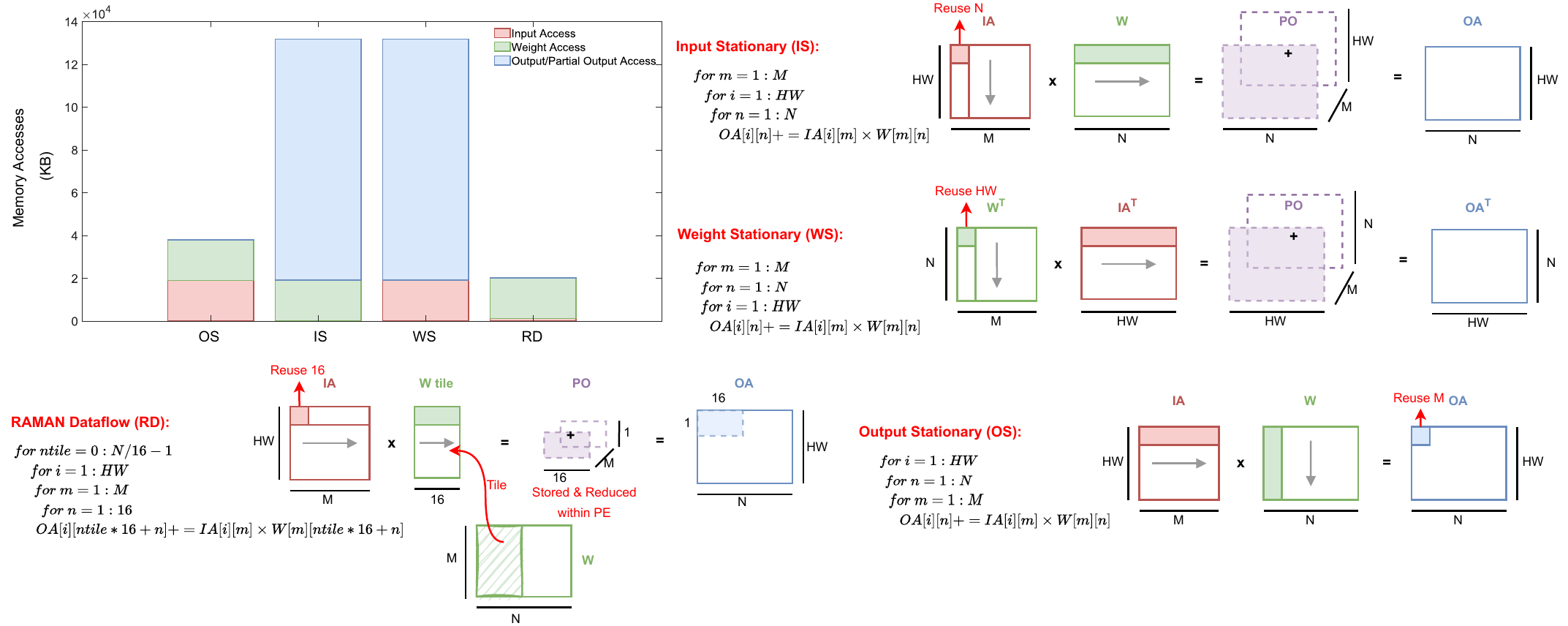}
	\end{center}
	\caption{Comparison of dataflows in terms of memory accesses.} 
	\label{fig:dataflow_analysis}
\end{figure*}

The state-of-the-art (SOA) accelerators use two different memory spaces for storing IAs and OAs in a single memory. The IA and OA memory spaces are logically partitioned to prevent memory collision issue. This conventional approach requires a peak memory of:

\vspace{-5mm}
\begin{equation}
MEM\_size_{(SOA)} = max(\{sum(IA^{l},OA^{l})\}_{l=1}^{L})
\end{equation}

Where $MEM\_size_{(SOA)}$ denotes the peak memory required by the state-of-the-art accelerators, $L$ represents the total number of layers in the network, $IA^{l}, OA^{l}$ represents the IA and OA memory sizes of a particular layer $'l'$. Effectively in the conventional approach, the activation memory size is governed by the layer with the maximum sum of IA and OA memory sizes. 

In our approach, as illustrated in Fig. \ref{fig:ip_op_overlapping}, the logical partition between IA and OA memory spaces is eliminated, meaning that the OAs are overwritten in the same IA memory space. The memory size required by the proposed approach is given by: 
\vspace{-5mm}

\begin{equation}
MEM\_size_{(RAMAN)} = max(\{max(IA^{l},OA^{l})\}_{l=1}^{L})
\end{equation}

Where $MEM\_size_{(RAMAN)}$ denotes the peak memory required by RAMAN. Compared with the conventional approach, the proposed memory reduction scheme reduces the peak activation memory of the MobileNetV1$^{\dagger}$ model trained for visual wake word (VWW) task by $37\%$  and the DS-CNN$^{\dagger}$ model trained for keyword spotting (KWS) application by $49\%$ as shown in Fig. \ref{fig:ip_op_overlapping}. We have made a couple of modifications to the original MobileNetV1 \cite{Howard2017} and DS-CNN models \cite{ZhangKWS2017} as per our requirements, and the modified models are denoted as MobileNetV1$^{\dagger}$ and DS-CNN$^{\dagger}$ models from here on.

However, the memory collision issue arises while overwriting the OA into the IA memory space if the IA is not consumed before the OA writeback. This problem can be solved by storing a copy of the IA tile in the cache and then proceeding with the computation, as shown in Fig. \ref{fig:ip_op_overlapping}. This makes the original copy of the IA tile (e.g., IA tile-1 in Fig. \ref{fig:ip_op_overlapping}) in the GLB-MEM activation bank redundant, allowing the OA tile to occupy that space. The disparity in tile sizes between IA and OA is another challenge with the proposed approach. If the OA tile is larger than the IA tile, it might corrupt the contents of IA tile-2 by overflowing the memory area of IA tile-1 (cf. Fig. \ref{fig:ip_op_overlapping}) and spilling over the subsequent tile (e.g., IA tile-2 in Fig. \ref{fig:ip_op_overlapping}). This is particularly true in the PW layer when N>M, but it's not a concern in the DW layer because OA is always less than (if stride>1) or equal (if stride=1)  to IA. Intelligent data organization inside the memory and pre-fetching the IA tile-2 to the cache prior to OA tile-1 writeback aids in resolving this issue.

\subsection{Dataflow to reduce memory accesses:} 
\label{sec:dataflow_Gustavsons}
Fig. \ref{fig:dataflow_analysis} shows the GLB-MEM memory accesses evaluated for a single PE for different dataflows and their corresponding abstract loop nests. The analysis was carried out on the PW layers of the MobileNetV1$^{\dagger}$ model \cite{Howard2017}, whose operation can be described as a matrix-multiplication of $IA_{(HW\times M)} \times W_{(M\times N)}$, where input channel $M$ is a shared dimension of multiplication. 
\par The output stationary (OS) dataflow, also termed inner product dataflow has the shared dimension in the innermost loop and achieves good output reuse ($M$ times) but has poor input reuse. It computes an OA element one at a time by traversing a row of IA and a column of W. On the other hand, the Input stationary (IS) and the Weight stationary (WS) dataflows achieve good input reuse ($N$ times) and weight reuse ($HW$ times), respectively, but poor output reuse. It computes one partial output matrix (PO) of size ($HW\times N$) at a time by traversing a row of W and a column of IA in IS (or a column of W\textsuperscript{T} and a row of IA\textsuperscript{T} in WS) and M such matrices are generated before the final reduction. The size of the partial output matrix is massive to be stored locally inside the PE and thus has to be saved in the GLB-MEM, creating significant output data traffic evident from Fig. \ref{fig:dataflow_analysis}. Additionally, the bandwidth required by the partial output matrix (24b value) is much higher than the final output (8b value). Thus, moving the partial output matrix from PE to GLB-MEM is costly. 
\par In this work, we employ a RAMAN dataflow (RD) inspired by Gustavson’s algorithm to reduce the overall data-movement cost of the PW layer. It computes a row of $16$ OAs at a time by traversing a row of IA, and a row of $16$ elements from W. Since just 16 partial outputs are generated at a time, it is locally stored and reduced inside the PE. The dataflow is the most efficient as it avoids the two extremes of OS (by reusing IA by a factor of $16$) and IS/WS (by eliminating the partial output traffic) dataflows. Only the quantized 8b accumulation result is sent to the GLB-MEM, drastically decreasing the output traffic. The W accesses from the GLB-MEM are reduced by storing a W tile in the cache and re-using them for $HW$ times in the PW layer. Compared to the OS and IS/WS dataflows, the RAMAN dataflow significantly reduces the PW layer memory access by 1.9x and 6.5x, respectively.


\section{Implementation Results}
\label{sec:impl_results}
This section assesses the RAMAN’s performance on Efinix FPGA \cite{EfinixTi60}.

\begin{figure*}[h]
  \centering
  \begin{subfigure}[b]{0.4\textwidth}
    \includegraphics[width=\textwidth]{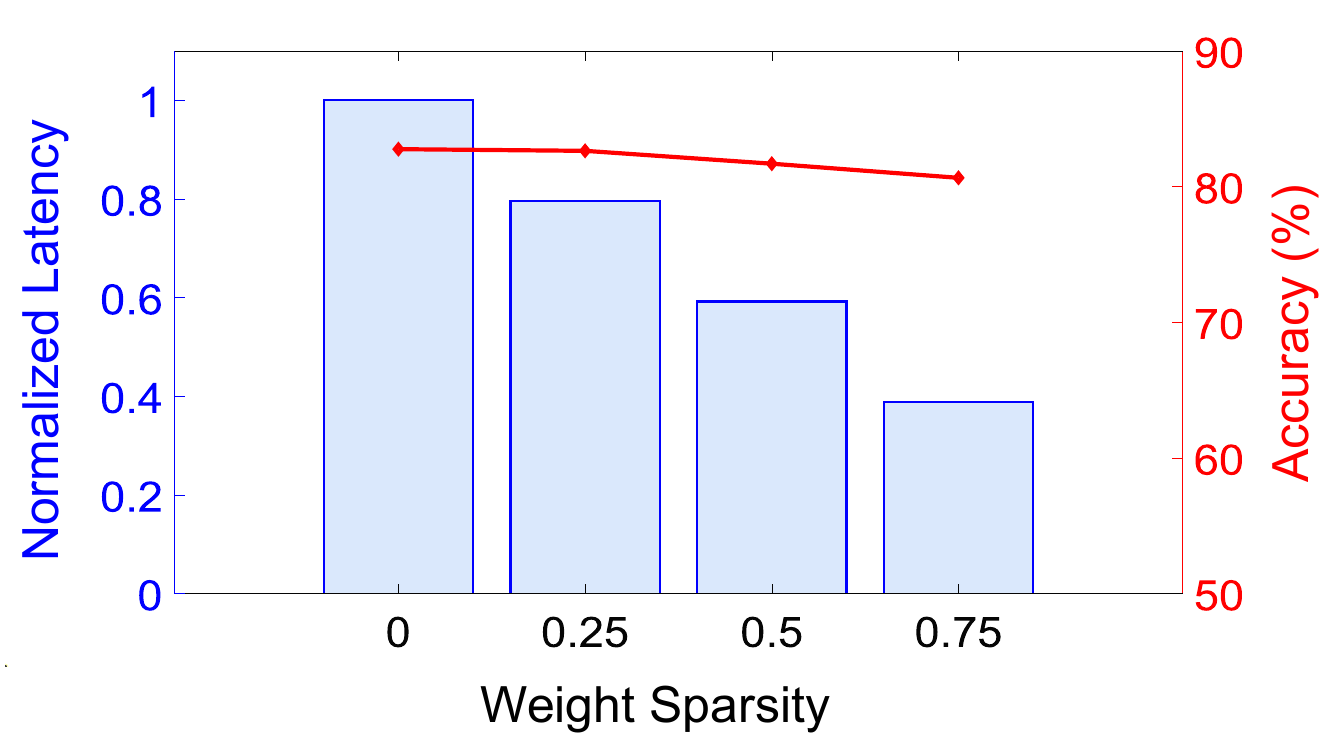}
    \caption{ }
  \end{subfigure}
    \begin{subfigure}[b]{0.4\textwidth}
    \includegraphics[width=\textwidth]{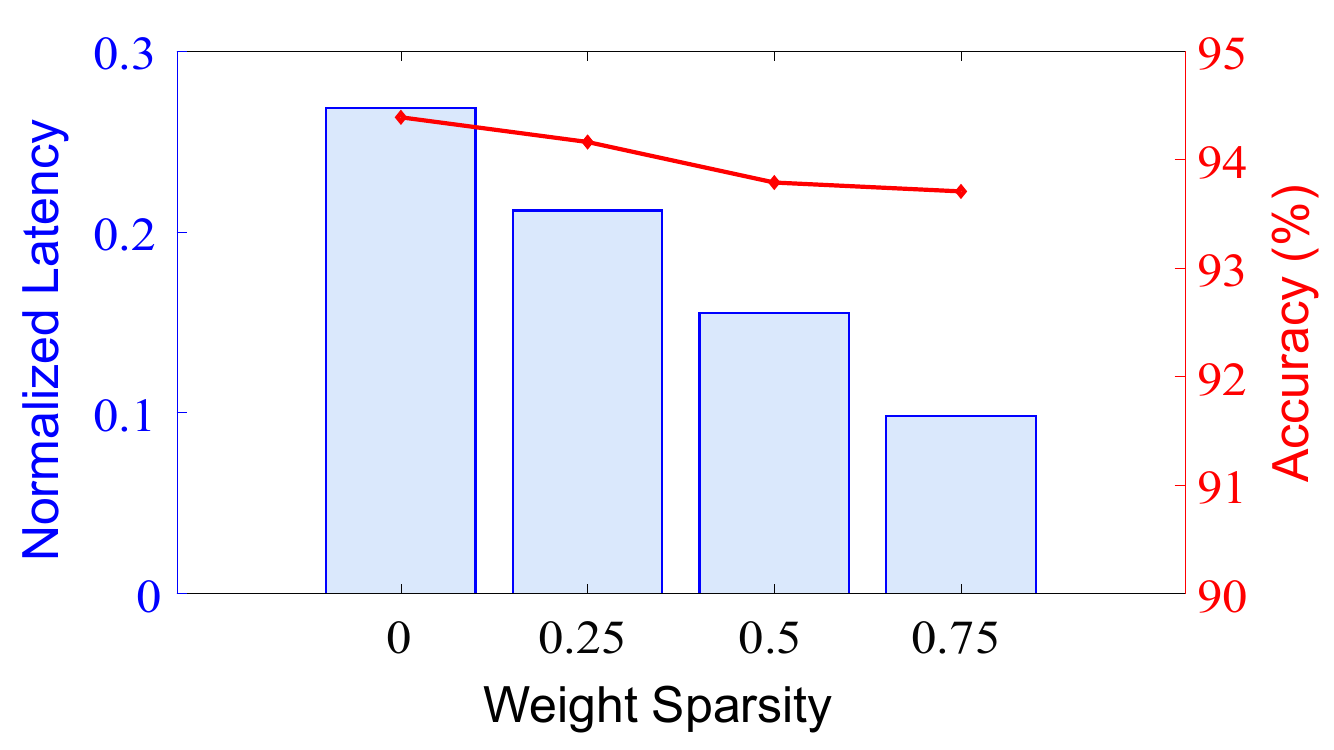}
    \caption{}
  \end{subfigure}
  \begin{subfigure}[b]{0.4\textwidth}
    \includegraphics[width=\textwidth]{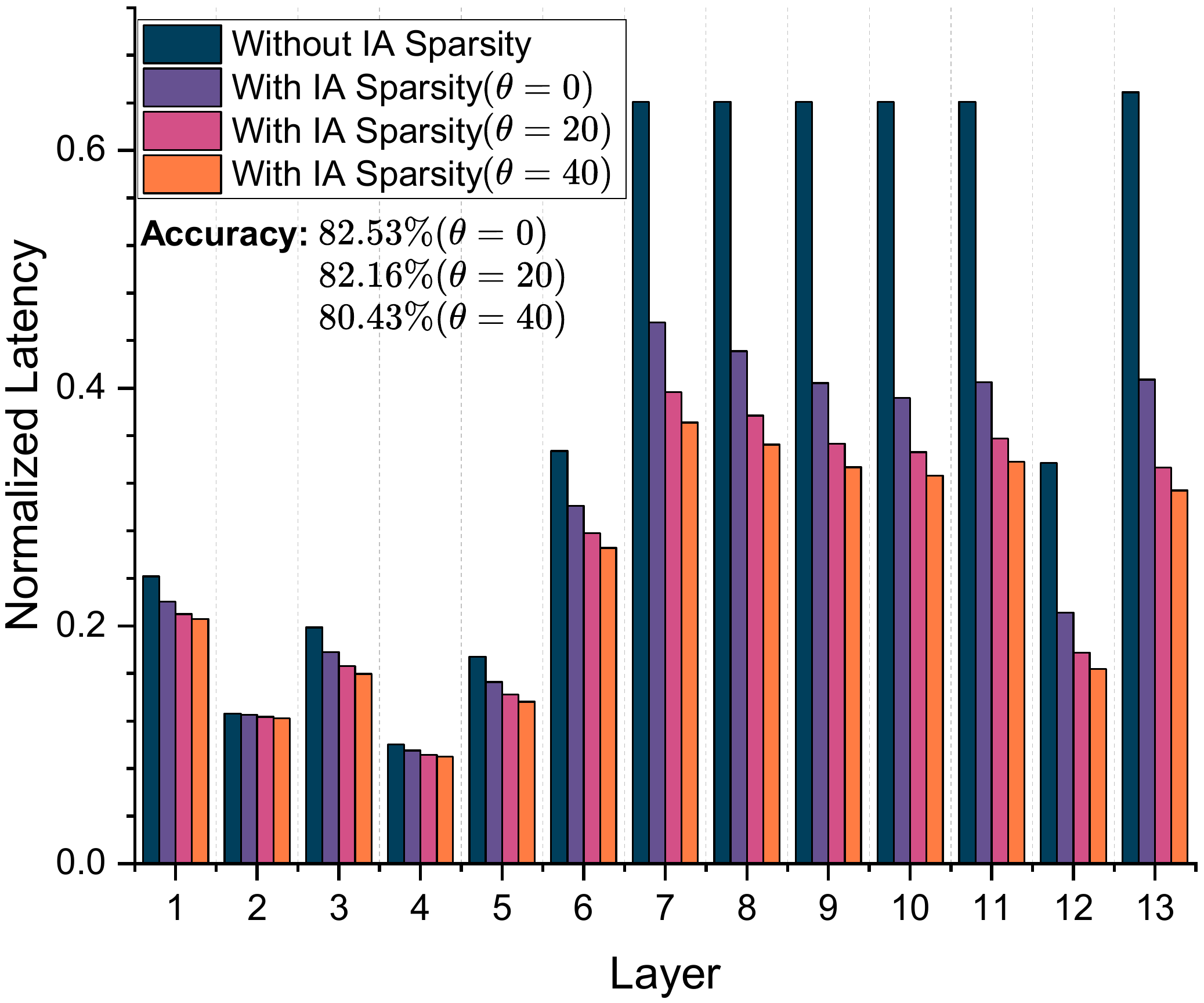}
    \caption{ }
  \end{subfigure}
    \begin{subfigure}[b]{0.4\textwidth}
    \includegraphics[width=\textwidth]{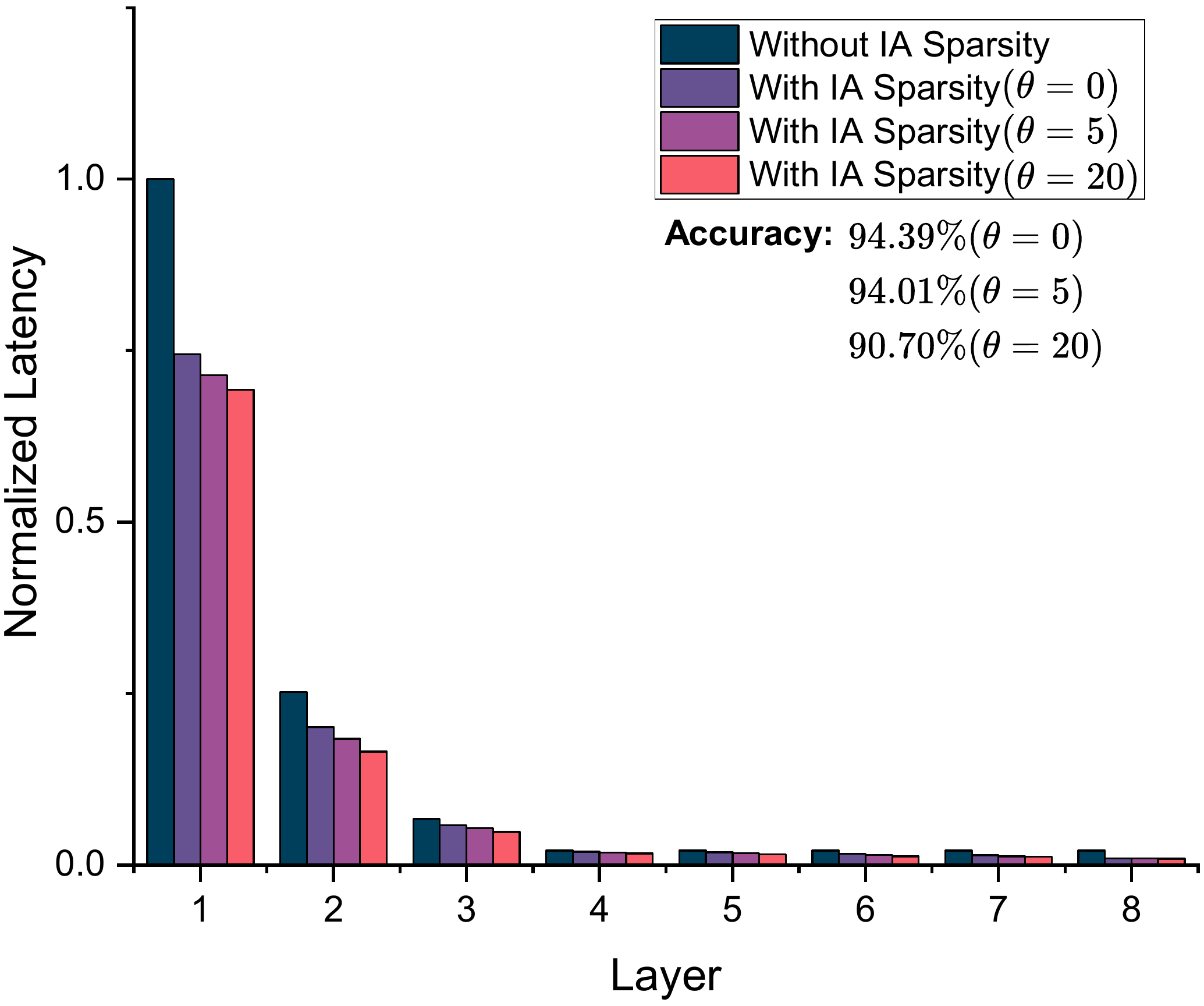}
    \caption{}
  \end{subfigure}

  \caption{Leveraging sparsity in latency reduction: Compile-time and run-time latency vs. accuracy trade-off for MobileNetV1$^{\dagger}$ model (left-panel) and DS-CNN$^{\dagger}$ model (right-panel). Top-panel: Latency of RAMAN for different weight sparsity (pruning) ratios set at compile time. Bottom-panel: Latency distribution of PW layers in RAMAN with run-time activation pruning at different activation pruning thresholds ($\theta$) with accuracy highlighted. The accuracy and latency estimates are for two tasks: keyword spotting using the DS-CNN$^{\dagger}$ model trained on the Google speech command dataset and image classification using the MobileNetV1$^{\dagger}$ model trained on the VWW dataset.}
  \label{fig:latency}
\end{figure*}



 \subsection{Efinix FPGA Implementation Results}
We evaluate RAMAN’s performance on popular networks aimed at tinyML edge computing applications: MobileNetV1 and DS-CNN. The input dimensions are resized as per the requirement: $96 \times 96$ for the MobileNetV1$^{\dagger}$ and $30 \times 32$ for the DS-CNN$^{\dagger}$ model. The DS-CNN$^{\dagger}$ model was trained on the google speech command dataset for the keyword spotting (KWS) application \cite{Warden2018Speech}. The input audio with each $1$s duration was sampled at $16$KHz and fed to the cascade of asymmetric resonators \cite{Xucar2018} to generate a cochleagram. RAMAN used the output cochleagram of size $30 \times 32$ as an input to infer the keyword. The MobileNetV1$^{\dagger}$ model was trained on the Visual Wake Words (VWW) dataset \cite{Chowdhery2019VWW} with input image converted to gray-scale. The specific MobileNetV1$^{\dagger}$ and DS-CNN$^{\dagger}$ network architectures deployed on RAMAN are provided in Section \Romannum{9} of the supplementary document. 
\par
\begin{figure*}[h!]
\centering

  \begin{subfigure}[b]{0.36\textwidth}
    \includegraphics[width=\textwidth]{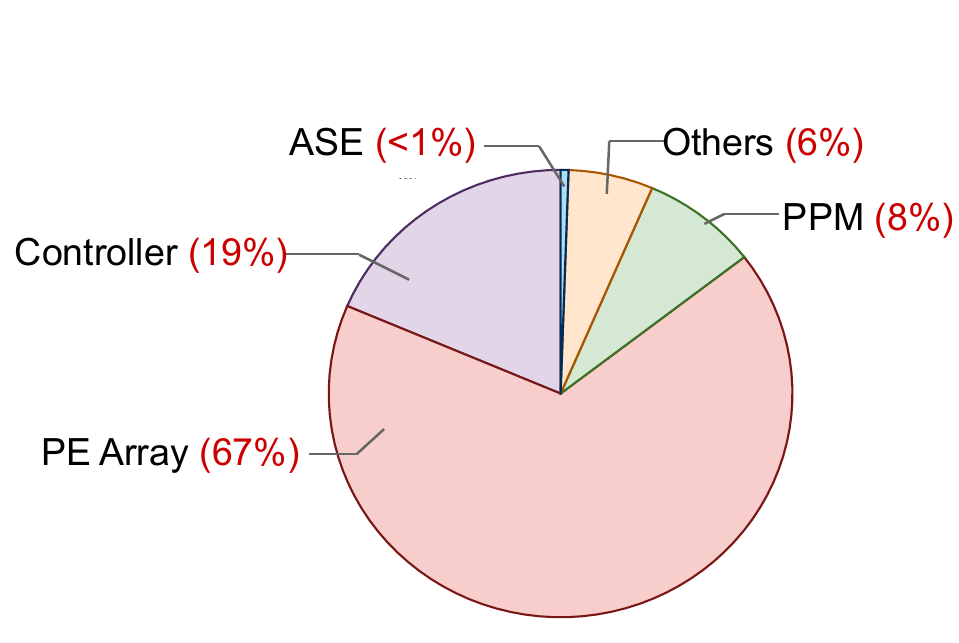}
    \caption{ }
  \end{subfigure}
    \begin{subfigure}[b]{0.38\textwidth}
    \includegraphics[width=\textwidth]{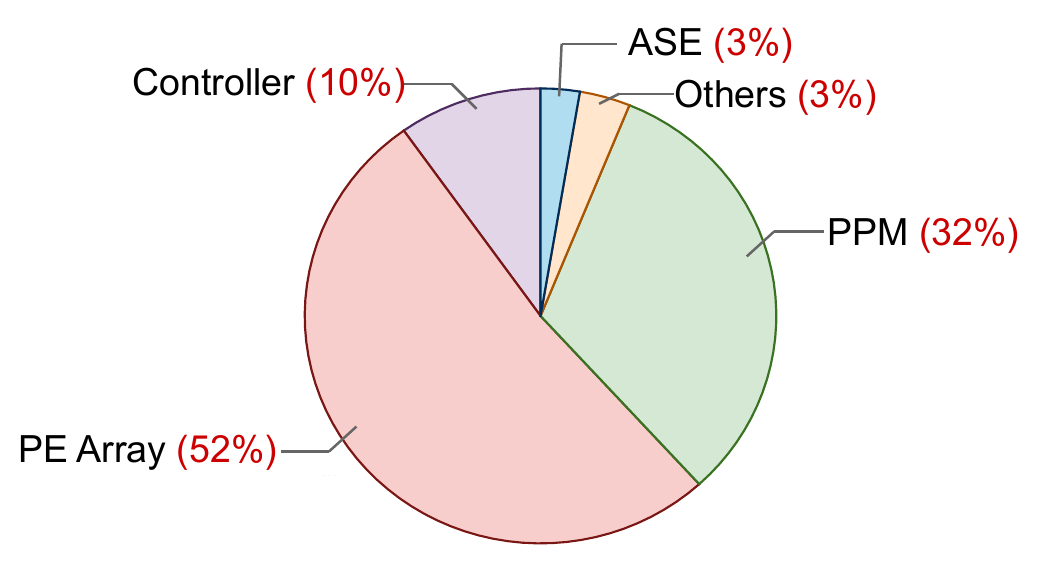}
    \caption{}
  \end{subfigure}

  \caption{Resource utilization breakdown of (a) LUTs and (b) Registers. }
  \label{fig:fpga_utilization}
\end{figure*}

RAMAN was implemented on an Efinix Ti60 FPGA, with parameters, instructions, and pre-processed inputs written into corresponding memories. The specifications of the RAMAN architecture are shown in Table \ref{tab:raman_specs}. The LUT breakdown of the RAMAN architecture is shown in Fig. \ref{fig:fpga_utilization}(a). It is evident that the controller and the PE array consume most of the LUTs in FPGA fabric, accounting for $86\%$ of LUTs, and the ASE utilization is insignificant. Fig. \ref{fig:fpga_utilization}(b) shows the register breakdown of the RAMAN architecture. The PE array utilizes $52\%$ of the registers since each PE comprises $16 \times 24$b RF to store Psums. RAMAN provides the flexibility to downsize the PE RF width to 20b (or lower) depending on the application to reduce register utilization. The PPM stores post-processing parameters in registers leading to $32 \%$ register utilization. The LUTs and registers are inferred as the eXchangeable Logic and Routing (XLR) cells in Efinix FPGA.


\begin{table}[h!]
\renewcommand{\arraystretch}{1.2}
\caption{RAMAN Specifications.}
\begin{center}
\begin{tabular}{|c|c|c|c|c|c|c|c|}
\hline
\textbf{Platform} &  Efinix Ti60\\ \hline
\textbf{Layers Supported} &  \begin{tabular}[c]{@{}c@{}} CONV, DW, PW, \\   FC and Max/Average pooling.   \end{tabular} \\ \hline
\textbf{Number of PEs} & 12 (4 MACs/PE) \\ \hline
\textbf{Reg-file Memory} &   \begin{tabular}[c]{@{}c@{}} PE Array:   $0.576$ KB \\ PPM: $0.32$ KB   \end{tabular}  \\ \hline
\textbf{Clock Rate} & 75 MHz \\ \hline
\textbf{Arithmetic Precision}  & \begin{tabular}[c]{@{}c@{}} W \& IAs: 2b, 4b or 8b fixed point, \\  Psums: 24b fixed point.   \end{tabular}  \\ \hline
\textbf{Power} & \begin{tabular}[c]{@{}c@{}} $137$ mW for MobileNetV1$^{\dagger}$ \\  $132$ mW for DS-CNN$^{\dagger}$  \end{tabular}\\ \hline
\textbf{XLR cells}  & $52261$ ($85.96\%$ util.) \\ \hline
\textbf{DSPs} & $61$ ($38.12\%$ util.) \\ \hline
\textbf{Memory Blocks} &   \begin{tabular}[c]{@{}c@{}}$168$ ($65.62\%$ util.) for MobileNetV1$^{\dagger}$ \\  $118$ ($46.09\%$ util.) for DS-CNN$^{\dagger}$   \end{tabular}   \\ \hline 
\textbf{Theoretical Throughput} & 7.2 GOp/s (3.6 GMACS) \\ \hline
\textbf{Effective Throughput} &  \begin{tabular}[c]{@{}c@{}} 13.5 GOp/s for MobileNetV1$^{\dagger}$ \\ 10.5 GOp/s  for DS-CNN$^{\dagger}$  \end{tabular}    \\ \hline
\textbf{Energy Efficiency} &  \begin{tabular}[c]{@{}c@{}} 2355 Inferences/J for MobileNetV1$^{\dagger}$ \\ 6609 Inferences/J for DS-CNN$^{\dagger}$  \end{tabular}    \\ 
\hline
\end{tabular}
\end{center}
\label{tab:raman_specs}
    
\end{table}

\par 
The RAMAN architecture comprises $48$ MAC units operating at $75$ MHz, which theoretically translates to a throughput of $7.2$ GOp/s. However, since operations involving zero are skipped in the PW layer, we achieve an effective throughput of $13.5$ GOp/s and $10.5$ GOp/s for the MobileNetV1$^{\dagger}$ and DS-CNN$^{\dagger}$  models, respectively by exploiting both activation and weight sparsity. The power consumption of the RAMAN architecture on Efinix Ti60 FPGA is estimated to be $136.96$ mW ($89.37$ mW dynamic power + $47.6$ mW static power) and $131.77$ mW ($84.39$ mW dynamic power + $47.38$ mW static power) for the MobileNetV1$^{\dagger}$ and DS-CNN$^{\dagger}$ models, respectively. Therefore, the effective power efficiency of RAMAN at $75$ MHz and $75\%$ PW weight sparsity is $98.4$ GOp/s/W (or equivalently $ 2355$ Inferences/J) for the MobileNetV1$^{\dagger}$  and $79.68$  GOp/s/W (or equivalently $6609$ Inferences/J) for the DS-CNN$^{\dagger}$  model.  The power and memory breakdown of the RAMAN architecture is provided in Section \Romannum{10} of the supplementary document.
\par The average MAC utilization of the DW and PW layers is around 59$\%$ and 86$\%$, respectively, with overall utilization of 78$\%$. DW layers have a lower MAC utilization due to limited IA re-use and the time spent to fetch the IAs and Ws from the GLB-MEM memory. On the other hand, memory access latency of the PW layers in RAMAN is completely hidden with MAC operations; however, the latency introduced due to data fetch from the RF of PEs cannot be completely hidden as the next tile’s MAC operation can be started only after completely evicting the contents of the RF of the current tile. The MAC units remain idle while fetching the contents of RF and transferring them to the post-processing module. This problem can be solved by double buffering the reg-file in the PEs, increasing the PW layer utilization to 97$\%$. 

\par 

\subsection{Sparsity Results}

\subsubsection{Leveraging sparsity in latency reduction} Fig. \ref{fig:latency} shows the accuracy vs. latency trade-off at compile-time and run-time. Figs. \ref{fig:latency}(a) and \ref{fig:latency}(b) demonstrate the RAMAN processing latency and model accuracy as a function of weight sparsity for the MobileNetV1$^{\dagger}$ and DS-CNN$^{\dagger}$ models, respectively. The required degree of weight sparsity is pre-set at compile time using the hardware-aware balanced weight pruning technique described in Section \Romannum{5} of the supplementary document. The performance is assessed for four weight sparsity levels. It is evident that the latency reduction is almost linear with the degree of weight sparsity, and the accuracy degradation is minimal. Additionally, the latency distribution for the PW layers of the MobileNetV1$^{\dagger}$ and DS-CNN$^{\dagger}$ models are shown in Figs.\ref{fig:latency}(c) and \ref{fig:latency}(d). There is a significant reduction in latency by leveraging IA sparsity. The latency gains are substantial in the final layers of the MobileNetV1$^{\dagger}$ model due to an increase in IA sparsity and number of operations. In contrast, the DS-CNN$^{\dagger}$ model is computationally intensive in the initial layers. 
Furthermore, we compare the latency after run-time activation pruning for different thresholds. Thresholds are set based on the input data distribution. For the MobileNetV1$^{\dagger}$ model, RAP with pruning threshold $40$ reduces latency by an additional $16\%$ compared to no pruning case, with accuracy degradation of ${\approx}2\%$. For the DS-CNN$^{\dagger}$ model, RAP reduces total latency by $12\%$ with the pruning threshold set at $20$. However, the possibility of minimizing latency with RAP is only confined to the initial layers since the final layers of DS-CNN$^{\dagger}$ have low computational intensity.

\subsubsection{Leveraging sparsity in memory access reduction} 
Table \ref{cache_act_breakdown_table} shows the activation cache access breakdown for different layer types of the MobileNetV1$^{\dagger}$ and DS-CNN$^{\dagger}$ models. It is evident from the table that the majority of the cache accesses happen in the DW-PW layers, and the activation cache reads are more than the cache writes, leading to effective cache reuse. Additionally, leveraging IA sparsity reduces the cache accesses by $40-45\%$. Finally, Table \ref{cache_wgt_breakdown_table} presents the weight cache access breakdown for different sparsity or pruning ratios. Again, a similar trend is observed where reads dominate writes, indicating effective cache reuse and the cache accesses reduce with increased sparsity ratio. In addition, it is observed that the weight cache reads are further reduced by $30\%$ with IA sparsity since when the IA value is zero, the corresponding weight is not read from memory. However, the weight cache writes remain the same since all weight values are loaded to the cache initially, irrespective of IA sparsity. In addition, run-time activation pruning reduces IA cache access by $8-10\%$, and parameter cache reads by $13-21\%$.

\begin{table}[H]
\caption{Activation cache access breakdown for (a) DS-CNN$^{\dagger}$ and (b) MobileNetV1$^{\dagger}$  model.} 
\renewcommand{\arraystretch}{1.2}
\begin{subtable}[b]{0.5\textwidth}
\centering
\begin{tabular}{|c|cccccccc|}
\hline
\multirow{3}{*}{\textbf{Layer}}                                                                       
                                & \multicolumn{4}{c|}{\textbf{With IA sparsity (in KB)}}                           & \multicolumn{4}{c|}{\textbf{Without IA sparsity (in KB)}}   \\ \cline{2-9} 
                                & \multicolumn{2}{c|}{\textbf{Read}} & \multicolumn{2}{c|}{\textbf{Write}} & \multicolumn{2}{c|}{\textbf{Read}} & \multicolumn{2}{c|}{\textbf{Write}} \\ 
                                \cline{2-9} 
                                & \multicolumn{1}{c|}{\textbf{(a)}} & \multicolumn{1}{c|}{\textbf{(b)}} & \multicolumn{1}{c|}{\textbf{(a)}} & \multicolumn{1}{c|}{\textbf{(b)}} & \multicolumn{1}{c|}{\textbf{(a)}} & \multicolumn{1}{c|}{\textbf{(b)}} & \multicolumn{1}{c|}{\textbf{(a)}} & \textbf{(b)} \\ \hline
{CONV}           & \multicolumn{1}{c|}{0.5} &  \multicolumn{1}{c|}{9.8}  &  \multicolumn{1}{c|}{0.2} &  \multicolumn{1}{c|}{6.7}    & \multicolumn{1}{c|}{2.7}         & \multicolumn{1}{c|}{13.5}         & \multicolumn{1}{c|}{0.96}  & \multicolumn{1}{c|}{9.2}         \\ \hline
{DW}               & \multicolumn{1}{c|}{171} & \multicolumn{1}{c|}{356 }    & \multicolumn{1}{c|}{77}  &   \multicolumn{1}{c|}{144}  &  \multicolumn{1}{c|}{281}     &  \multicolumn{1}{c|}{630}  &  \multicolumn{1}{c|}{129}  &  \multicolumn{1}{c|}{246}      \\ \hline
{PW}               & \multicolumn{1}{c|}{47} & \multicolumn{1}{c|}{173}  & \multicolumn{1}{c|}{47} & \multicolumn{1}{c|}{173}  & \multicolumn{1}{c|}{76} & \multicolumn{1}{c|}{335}   & \multicolumn{1}{c|}{76} & \multicolumn{1}{c|}{335}      \\ \hline
{Pool}              & \multicolumn{1}{c|}{0}    & \multicolumn{1}{c|}{0}           & \multicolumn{1}{c|}{0}              & \multicolumn{1}{c|}{0}           & \multicolumn{1}{c|}{0}    & \multicolumn{1}{c|}{0}           & \multicolumn{1}{c|}{0}              & \multicolumn{1}{c|}{0}              \\ \hline
{FC}               & \multicolumn{1}{c|}{0.13} & \multicolumn{1}{c|}{0.26}        & \multicolumn{1}{c|}{0.06}  & \multicolumn{1}{c|}{0.26}       & \multicolumn{1}{c|}{0.13}   & \multicolumn{1}{c|}{0.26}      & \multicolumn{1}{c|}{0.06}    & \multicolumn{1}{c|}{0.26}         \\ \hline
\textbf{Total}                  & \multicolumn{1}{c|}{\textbf{219}}      & \multicolumn{1}{c|}{\textbf{539}}       & \multicolumn{1}{c|}{\textbf{124}}      & \multicolumn{1}{c|}{\textbf{323}}  & \multicolumn{1}{c|}{\textbf{360}} & \multicolumn{1}{c|}{\textbf{979}}  & \multicolumn{1}{c|}{\textbf{206}}  & \multicolumn{1}{c|}{\textbf{590}}      \\ \hline
\end{tabular}
\label{kws_cache_act_breakdown_table}
\end{subtable}
\label{cache_act_breakdown_table}
\end {table}

\begin{table}[H]
\caption{Weight cache access breakdown for (a) DS-CNN$^{\dagger}$ and (b) MobileNetV1$^{\dagger}$  model.}
\renewcommand{\arraystretch}{1.2}
\begin{subtable}[b]{0.5\textwidth}
\centering
\begin{tabular}{|c|cccccc|}
\hline
\multirow{3}{*}{\textbf{\begin{tabular}[c]{@{}c@{}}Weight\\ Sparsity  \end{tabular}}}                                                                       
                                & \multicolumn{4}{c|}{\textbf{Read (in KB)}}                           & \multicolumn{2}{c|}{ \textbf{Write (in KB)}}   \\ \cline{2-7} 
                                & \multicolumn{2}{c|}{\textbf{With IA sparsity}} & \multicolumn{2}{c|}{\textbf{Without IA sparsity}} & \multicolumn{2}{c|}{\textbf{-}} \\ 
                                \cline{2-7} 
                                & \multicolumn{1}{c|}{\textbf{(a)}} & \multicolumn{1}{c|}{\textbf{(b)}} & \multicolumn{1}{c|}{\textbf{(a)}} & \multicolumn{1}{c|}{\textbf{(b)}} & \multicolumn{1}{c|}{\textbf{(a)}} & \multicolumn{1}{c|}{\textbf{(b)}} \\ \hline
{0\%}           & \multicolumn{1}{c|}{1727} &  \multicolumn{1}{c|}{6344}  &  \multicolumn{1}{c|}{2451} &  \multicolumn{1}{c|}{9425}    & \multicolumn{1}{c|}{49}         & \multicolumn{1}{c|}{295}         \\ \hline
{25\%}               & \multicolumn{1}{c|}{1295} & \multicolumn{1}{c|}{4908}    & \multicolumn{1}{c|}{1839}  &   \multicolumn{1}{c|}{7250}  &  \multicolumn{1}{c|}{37}     &  \multicolumn{1}{c|}{222}      \\ \hline
{50\%}               & \multicolumn{1}{c|}{863} & \multicolumn{1}{c|}{3472}  & \multicolumn{1}{c|}{1226} & \multicolumn{1}{c|}{5075}  & \multicolumn{1}{c|}{25} & \multicolumn{1}{c|}{149}    \\ \hline
{75\%}              & \multicolumn{1}{c|}{432}    & \multicolumn{1}{c|}{2036}           & \multicolumn{1}{c|}{613}              & \multicolumn{1}{c|}{2900}           & \multicolumn{1}{c|}{12}    & \multicolumn{1}{c|}{76}                 \\ \hline

\end{tabular}


\end{subtable}

\label{cache_wgt_breakdown_table}
\end {table}

\subsubsection{Leveraging sparsity in storage reduction} 
The global memory requirements of the design are tabulated in Table \ref{gmem_table}. The peak activation memory needed is obtained by overwriting OAs in the same IA memory space. The parameter memory needed is the sum of memory needed to store weights and post-processing parameters of all the layers. Taking pruning into account, the parameter memory reduces with an increase in pruning percentage, as shown in Table \ref{gmem_table}. \\

\begin{table}[H]
\caption{Global memory requirements.}
\renewcommand{\arraystretch}{1.2}
\centering
\begin{tabular}{|c|c|cccc|}
\hline
\multirow{2}{*}{\textbf{\begin{tabular}[c]{@{}c@{}}Memory\\ (in KB)\end{tabular}}} & \multirow{2}{*}{\textbf{\begin{tabular}[c]{@{}c@{}}Activation\\ Memory\end{tabular}}} & \multicolumn{4}{c|}{\textbf{\begin{tabular}[c]{@{}c@{}}Parameter Memory with \\ different pruning ratios\end{tabular}}} \\ \cline{3-6} 
                                      &                                             & \multicolumn{1}{c|}{\textbf{0}}    & \multicolumn{1}{c|}{\textbf{0.25}}    & \multicolumn{1}{c|}{\textbf{0.5}}   & \textbf{0.75}   \\ \hline
DS-CNN$^{\dagger}$                                    & 54.72                                       & \multicolumn{1}{c|}{61.968}              & \multicolumn{1}{c|}{49.656}             & \multicolumn{1}{c|}{37.392}           & 25.104           \\ \hline
\begin{tabular}[c]{@{}c@{}}Mobile\\ -NetV1$^{\dagger}$  \end{tabular}                                & 44.56                                       & \multicolumn{1}{c|}{324.288}           & \multicolumn{1}{c|}{251.328}            & \multicolumn{1}{c|}{178.368}          & 105.384           \\ \hline
\end{tabular}
\label{gmem_table}
\end{table}

\subsection{Comparison with prior works}
We compare RAMAN with other state-of-the-art (SOA) implementations in Table \ref{tab:comparsion_soa}. While most SOA accelerators support just CONV, RAMAN supports a wide range of DNN topologies, including CONV (standard convolutions), DW + PW (separable convolutions), FC, max and average pooling layers. RAMAN supports variable precision quantization of IAs and Ws. Unlike other SOA implementations, which use both on-chip and off-chip memories, RAMAN, being a tinyML edge accelerator, exclusively employs an on-chip memory to store activations and weights. 
\par RAMAN's resource utilization and power efficiency are significantly better than other prior implementations making it an ideal candidate for resource-constrained edge devices. Table \ref{tab:comparsion_soa} also compares the architectural features regarding sparsity, peak-memory reduction and programmability. Very few works in the literature leverage sparsity in their design. NullHop \cite{Aimar2019} presents an architecture that exploits activation sparsity to accelerate computation (by zero skipping) and reduce storage requirements. However, the architecture doesn’t exploit weight sparsity. McDanel et al. \cite{McDanel2019} present a method to exploit weight sparsity in the systolic arrays by a column-combining scheme that packs the sparse non-zero weights into a denser format. The activation sparsity is leveraged by clock-gating the logic when the activation is zero to save power. However, this method won’t save computation cycles and memory accesses like in RAMAN. Unlike other prior implementations shown in Table \ref{tab:comparsion_soa}, RAMAN co-optimizes the software and hardware stack by training the model based on the underlying accelerator constraints. Hardware-aware pruning is one such optimization vector, where the network is pruned to reduce the workload imbalance of different PEs and improve utilization. Additionally, RAMAN performs run-time activation pruning and peak-activation memory reduction to make it compatible with tinyML edge applications which none of the other implementation support. Furthermore, most prior implementations are limited to standard convolution layers, and the unsupported ones are processed off-chip. In contrast, our design can support different network topologies eliminating the off-chip computation or FPGA reconfiguration.

\begin{table*}
\caption{Comparison with the state-of-the-art (SOA) FPGA implementations.}

\label{tab:comparsion_soa}
\centering
\renewcommand{\arraystretch}{1.2}
{\begin{tabular}{|c|c|c|c|c|c|c|c|c|} 
\hline
\textbf{} & \textbf{\begin{tabular}[c]{@{}c@{}}NullHop \\\cite{Aimar2019}\end{tabular}}  & \textbf{\begin{tabular}[c]{@{}c@{}}Shah et al. \\\cite{Shah2018}\end{tabular}}  & \textbf{\begin{tabular}[c]{@{}c@{}}McDanel et al. \\\cite{McDanel2019}\end{tabular}} &
\textbf{\begin{tabular}[c]{@{}c@{}}Zhang et al. \\\cite{Zhang2021mdpi}\end{tabular}} & 
\textbf{\begin{tabular}[c]{@{}c@{}}Nguyen et al. \\\cite{Nguyen2019}\end{tabular}} 
& \textbf{\begin{tabular}[c]{@{}c@{}}Ma et al. \\\cite{Ma2018}\end{tabular}} 
& \textbf{\begin{tabular}[c]{@{}c@{}}Yu et al. \\\cite{Yu2020}\end{tabular}} 

& \begin{tabular}[c]
{@{}c@{}}\textbf{RAMAN}\\ \textbf{(This work)}\end{tabular} \\ \hline

\textbf{\begin{tabular}[c]{@{}c@{}}Platform\end{tabular}} & \begin{tabular}[c] {@{}c@{}}Xilinx \\ Zynq-7100\end{tabular}  & \begin{tabular}[c] {@{}c@{}}Xilinx \\ Virtex-7\end{tabular}  &  
\begin{tabular}[c] {@{}c@{}}Xilinx \\ Virtex-7\end{tabular} &  
\begin{tabular}[c] {@{}c@{}}Xilinx \\ Zynq-7000\end{tabular}  &  
\begin{tabular}[c] {@{}c@{}}Xilinx \\ Virtex-7\end{tabular}  &  
\begin{tabular}[c] {@{}c@{}}Intel Arria \\ 10 GX\end{tabular}  & 
\begin{tabular}[c] {@{}c@{}}Xilinx \\ Kintex-7\end{tabular} &  
\begin{tabular}[c] 
{@{}c@{}}Efinix \\ Ti60  \end{tabular} \\ \hline

\textbf{\begin{tabular}[c]{@{}c@{}}Layers\\Supported\end{tabular}} &  CONV  & CONV & {\begin{tabular}[c]{@{}c@{}}CONV+\\ PW+FC\end{tabular}} &\begin{tabular}[c]   {@{}c@{}}CONV\end{tabular} & \begin{tabular}[c]   {@{}c@{}}CONV\end{tabular}  &\begin{tabular}[c]   {@{}c@{}}CONV+\\Pool+FC\end{tabular} & \begin{tabular}[c]   {@{}c@{}}CONV+\\Pool+FC\end{tabular} & \begin{tabular}[c]   {@{}c@{}}CONV+DW+\\PW+Pool+FC\end{tabular} \\ \hline
     
\textbf{\begin{tabular}[c]{@{}c@{}}Num. of\\Mults\end{tabular}}  & 128 &  1024 & N/A & 288 & N/A & 3136 & 1024 & 48  \\ \hline

\textbf{Precision}& 16b & 18b & N/A & 8b & W:1b \& IA:6b & 16b & 8b & 2, 4 or 8b  \\ \hline

\textbf{\begin{tabular}[c]{@{}c@{}}LUTs\end{tabular}}& 229k &  78.32k & 239k & 83.24k & 86k & 138k$^{\S}$ & 94.76k & 37.2k \\ \hline

\textbf{\begin{tabular}[c]{@{}c@{}}Registers\end{tabular}}& 107k & 96.93k & 201k & 109k & 60k &  N/A & 150.85k & 8.6k \\ \hline

\textbf{\begin{tabular}[c]{@{}c@{}}DSPs\end{tabular}}& 128 & 1034 & 112 & 192 & 168 & 1518 & 516 & 61 \\ \hline


\textbf{\begin{tabular}[c]{@{}c@{}}Frequency (MHz)\end{tabular}} & 60  &  150 & 170 & 200 & 200 & 200 & 200 & 75 \\ \hline

\textbf{\begin{tabular}[c]{@{}c@{}}Power Efficiency\\(GOp/s/W)\end{tabular}} & 27.4  &  7.2 & N/A & 18.71 & 53.29 & N/A & 21.45 & 98.47$^{\star}$ \\ \hline

\textbf{\begin{tabular}[c]{@{}c@{}}Leveraging \\ Act. Sparsity\end{tabular}} & \textbf{\begin{tabular}[c]{@{}c@{}}Yes (Zero-\\skipping)\end{tabular}} & No & \textbf{\begin{tabular}[c]{@{}c@{}}Yes (Clock-\\gating)\end{tabular}} & No & No & No & No & \textbf{\begin{tabular}[c]{@{}c@{}}Yes (Zero-\\skipping \& gating)\end{tabular}} \\ \hline

\textbf{\begin{tabular}[c]{@{}c@{}}Leveraging \\ Weight Sparsity\end{tabular}} & No & No & \textbf{Yes} & No & No & No & No & \textbf{Yes}\\  \hline

\textbf{\begin{tabular}[c]{@{}c@{}}Hardware-Aware\\ Pruning\end{tabular}} & No & No & No & No & No & No & No & \textbf{Yes}\\ \hline

\textbf{\begin{tabular}[c]{@{}c@{}}Run-time \\ Act. Pruning \end{tabular}} & No & No & No & No & No & No & No & \textbf{Yes}\\ \hline

\textbf{\begin{tabular}[c]{@{}c@{}}Peak Act.\\ Mem. Reduction  \end{tabular}} & No & No & No & No & No & No & No & \textbf{Yes}\\ \hline

\textbf{\begin{tabular}[c]{@{}c@{}}Variable Precision  \end{tabular}} & No & No & No & No & No & No & No & \textbf{Yes}\\ \hline

\textbf{\begin{tabular}[c]{@{}c@{}}Run-time\\Programmability \end{tabular}} & \begin{tabular}[c]{@{}c@{}}Limited to \\ CONV   \end{tabular} & \begin{tabular}[c]{@{}c@{}}Limited to \\ CONV   \end{tabular} & Limited & No$^{\P}$ & No$^{\P}$  & \begin{tabular}[c]{@{}c@{}}Limited  \end{tabular} & Limited & \textbf{Yes}  \\ \hline

\end{tabular}}

\begin{tablenotes}
      \small
      \item $^{\P}$The architecture is limited to YOLOv2 implementation.
      \item $^{\S}$For Intel FPGA (Logic elements/ALMs).
      \item $^{\star}$Estimated for 8b precision.
       \item Run-time programmability means that the accelerator can support different network topologies without reconfiguring FPGA or synthesizing the design again. 
    \end{tablenotes}

\end{table*}

\section{Conclusions}
\label{sec:conclusion}
Deep neural networks (DNNs) introduce weight and activation sparsity, enabling deep learning applications to operate more efficiently on hardware platforms with constrained resources and energy. However, these sparse models need specialized hardware architectures to fully benefit from the sparsity for storage, latency, and energy gains. In this work, a reconfigurable and sparse neural network accelerator exploiting both weight and activation sparsity is proposed. RAMAN uses an activation sparsity engine to leverage unstructured activation sparsity and a hardware-aware balanced pruning to exploit structured weight sparsity. We propose a novel dataflow inspired by Gustavson’s algorithm that enables the Psum reduction with the PE array and significantly reduces the writeback traffic. The dataflow reduces the PW layer memory accesses by 1.9x compared to output stationary dataflow and 6.5x compared to input/weight stationary dataflow. Furthermore, we propose a technique to lower peak memory activation by overlaying IA and OA on the same memory space, which can reduce storage requirements by up to $50\%$. These memory optimizations in terms of memory accesses and memory storage enable RAMAN to be deployable on edge with a small form factor.  

\par 
RAMAN supports a wide range of DNN topologies from standard CNN layers to modern DS-CNNs and can be configured to support accuracy vs. power/latency tradeoffs using techniques deployed at compile time and run time.  RAMAN architecture was implemented on Efinix FPGA with $37.2$K LUTs using $48$ MAC units distributed across $3 \times 4$ PEs. The design achieves an overall energy efficiency of $2355$ and $6609$ Inference/J for MobileNetV1$^{\dagger}$ and DS-CNN$^{\dagger}$ models at $75$ MHz on Efinix FPGA. The effective power efficiency of the system is estimated to be $98.4$ and $79.68$ GOp/s/W for MobileNetV1$^{\dagger}$ and DS-CNN$^{\dagger}$ models, respectively. A demonstration video of the proposed RAMAN accelerator on the Efinix Ti60 FPGA board for the keyword spotting task, where we control the maze game using the keywords uttered by the user, can be found here \url{https://youtu.be/sCksj7nlBY8}.

\section{Acknowledgements}
The authors would like to express their sincere gratitude and appreciation to their colleagues, Madhuvanthi Srivatsav, Hitesh Pavan, Anand Chauhan, and Shankaranarayanan, for their invaluable help throughout this work.

\bibliographystyle{IEEEtran}
\bibliography{ref.bib}

\end{document}


%

\title{RAMAN: Supplementary Material}

\author{Adithya Krishna, Srikanth Rohit Nudurupati, Chandana D G, Pritesh Dwivedi, Andr\'e van Schaik, Mahesh Mehendale and Chetan Singh Thakur* }

\maketitle

\begin{figure*}[h!]
\centering\includegraphics[width=10.9cm]{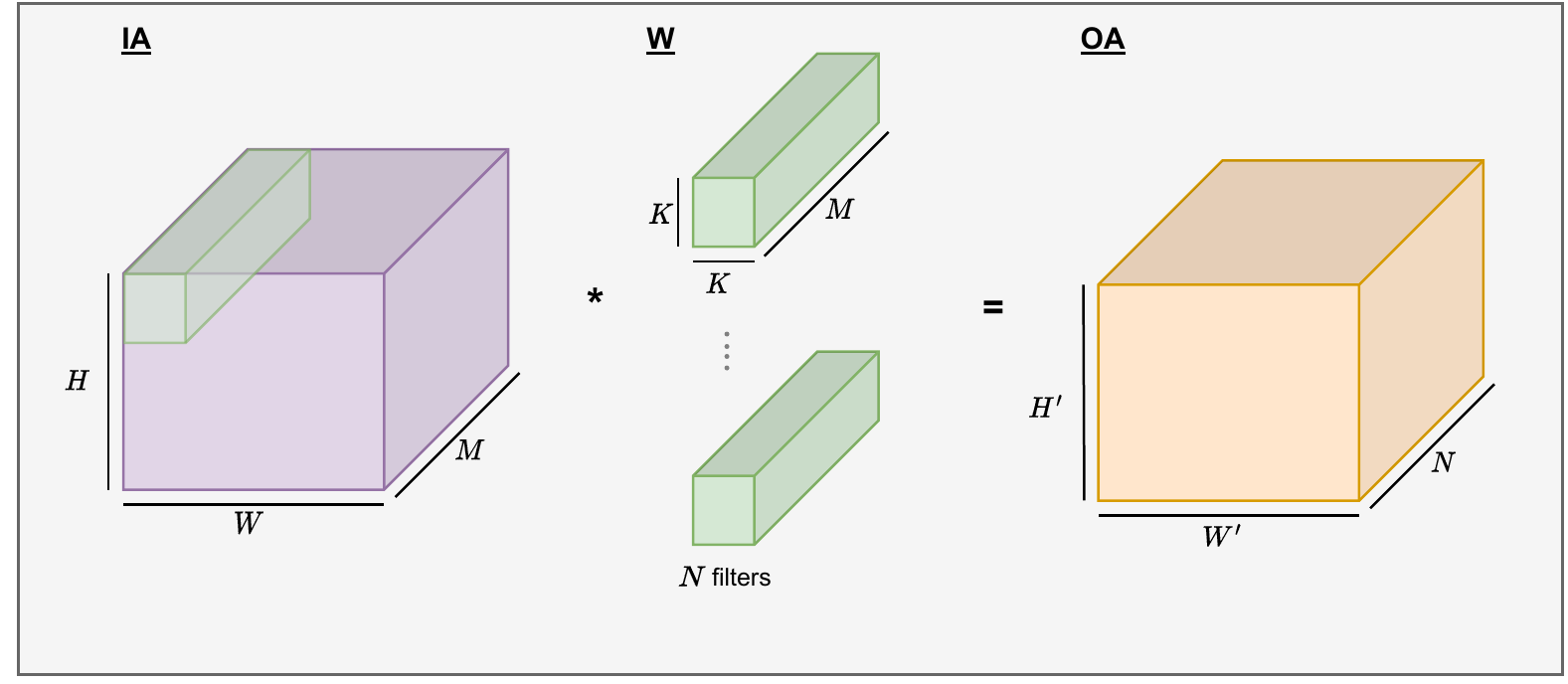}
\centering\caption{Computation of a CNN layer.}
\label{fig:cnn_bd}
\end{figure*}

%


\section{CNN Basics}
\label{chap:CNN_basics}
The CNN algorithm is built by stacking many computation layers for feature extraction and classification. CNN layers apply filters to an input feature map to extract embedded features and generate an output feature map as shown in Fig. \ref{fig:cnn_bd}. By constructing a very deep hierarchy of layers, modern CNNs achieve significantly higher accuracy by transforming the input image data into highly abstract representations known as feature maps. The computation of a layer in CNN is defined by:

\begin{equation}
\begin{split}
\mathbf{OA}_{n,x,y} =  ReLU  \Biggl( \mathbf{b}_{n} +  \sum\limits_{m=0}^{M-1} \sum\limits_{i=0}^{K-1} \sum\limits_{j=0}^{K-1}  \mathbf{IA}_{m,x+i,y+i} \\
 \cdot \mathbf{W}_{n,m,i,j} \Biggl)
\end{split}
\label{eqn:cnn_eqn}
\end{equation}

Where $H/W$ is the height and width of the input feature map denoted as IA, $H'/W'$ is the height and width of the output feature map denoted as OA. $M$ is the number of input channels, and $N$ is the number of filters or output channels. $W$ denotes the weight and $K$ denotes the filter height and width. A rectified linear unit (ReLU) activation function is applied to introduce non-linearity. 
The computational cost of the convolution is:

\begin{equation}
K \cdot K \cdot M \cdot N \cdot H \cdot W
\end{equation}

\par CNN computation can be divided into two phases: 1) Training- where parameters are trained by observing a vast amount of training examples, and 2) Inference- where the network is deployed in the field with the trained parameters. Training is performed on the cloud on GPUs, and inference depends on the application and can employ GPUs, CPUs, MCUs, or customized neural network accelerators. For edge applications with limited area and power budget, customized neural network accelerators are preferred. However, DNNs require millions of operations even during inference, necessitating software and hardware optimizations to be deployable on edge. One such network architecture that seeks to minimize the number of operations is depth-wise separable convolutions (DS-CNNs), rendering it ideal for deployment on the edge. 

\begin{figure*}[h!]
  \centering

  \begin{subfigure}[b]{0.6\textwidth}
    \includegraphics[width=\textwidth]{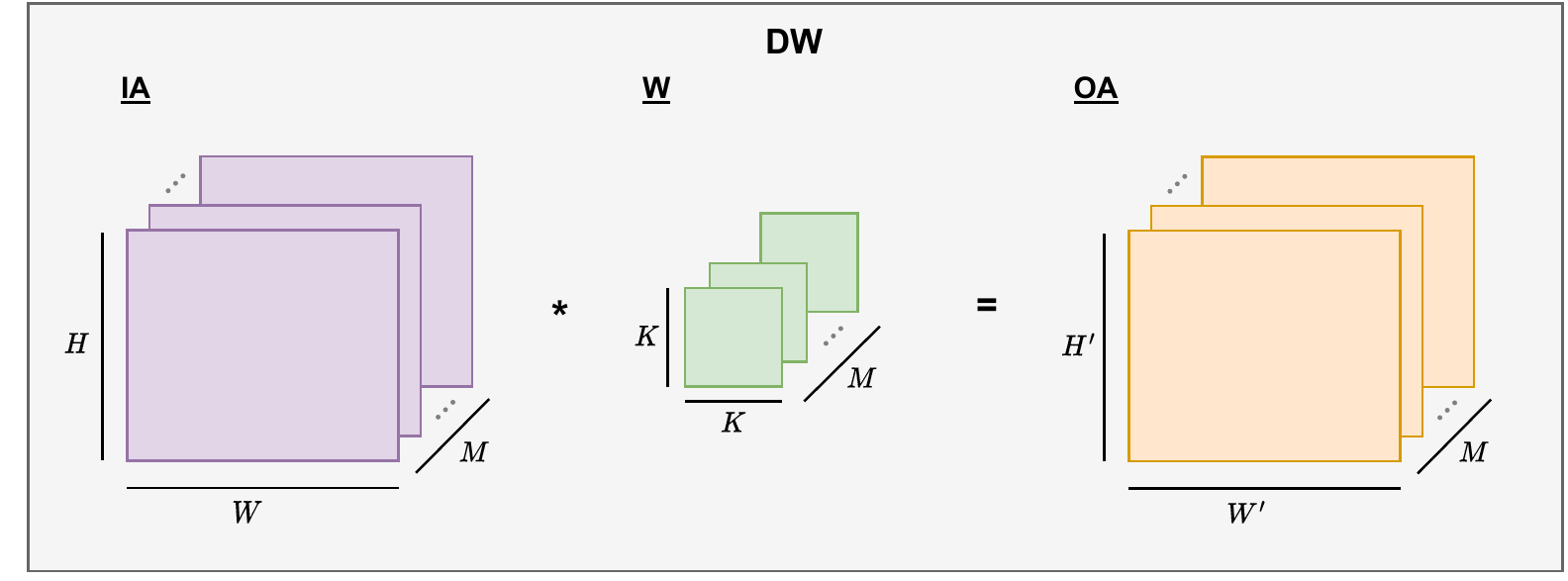}
    \caption{ }
  \end{subfigure}
    \begin{subfigure}[b]{0.6\textwidth}
    \includegraphics[width=\textwidth]{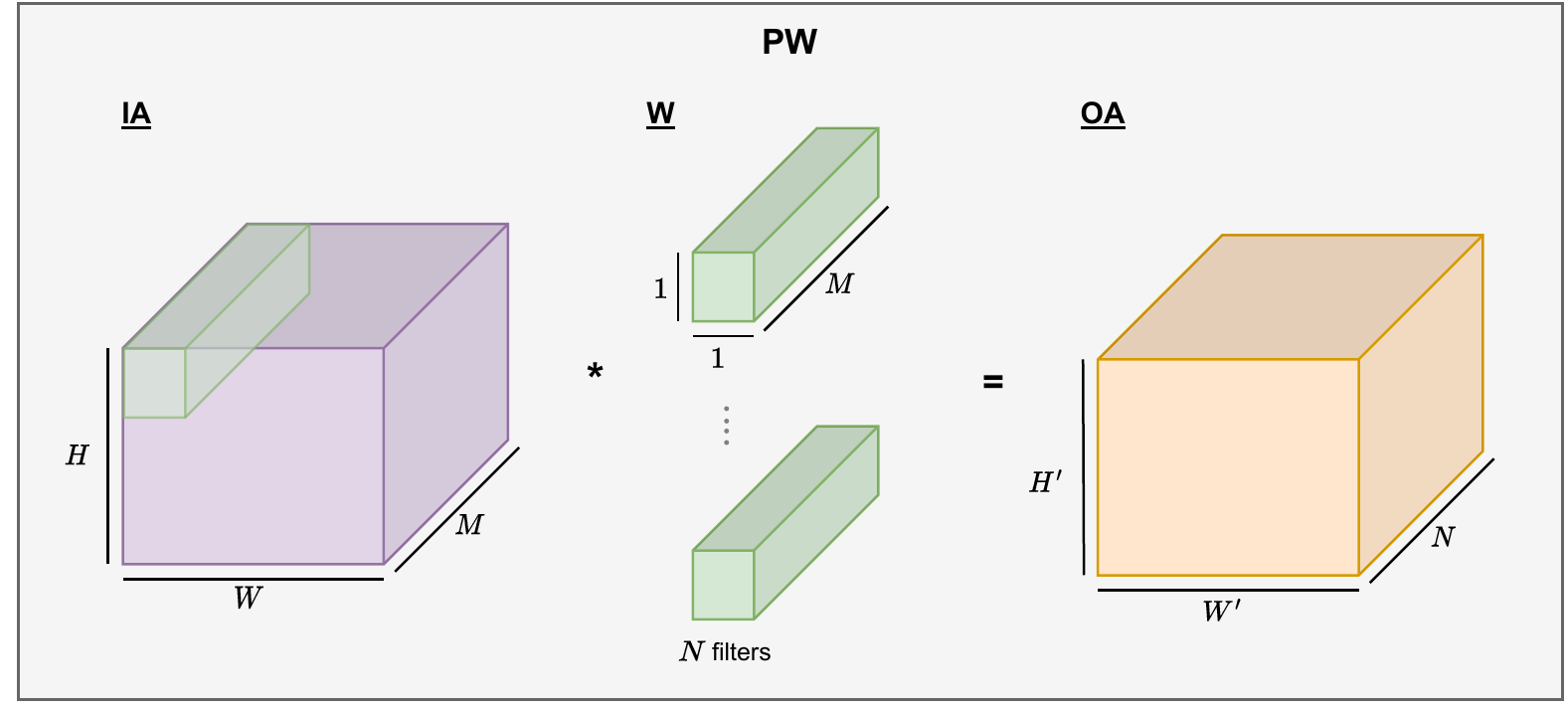}
    \caption{}
  \end{subfigure}

  \caption{DS-CNN computation of a layer. (a) Depth-wise (DW) computation, and (b) Point-wise (PW) computation. }
  \label{fig:ds_cnn}
\end{figure*}

The DS-CNN splits the convolution process into a point-wise and a depth-wise convolution. The depth-wise convolution applies a single filter for every input channel to capture the spatial information as shown in Fig. \ref{fig:ds_cnn}(a). In the point-wise convolution, we apply a 1x1 filter to combine the output of the depth-wise convolution as shown in Fig. \ref{fig:ds_cnn}(b). A standard convolution filters and combines in a single step, whereas the DS-CNN does the same in two stages. As a result, the model size and computation are reduced substantially, which is ideal for tinyML edge applications. 

Depth-wise convolution with one filter per input channel can be written as:

\begin{equation}
{\mathbf{OA}}_{x,y,m} =  \sum\limits_{i=0}^{K-1} \sum\limits_{j=0}^{K-1} {\mathbf{W}}_{i,j,m} \cdot \mathbf{IA}_{x+i,y+j,m}
\end{equation}
where ${\mathbf{W}}$ is the depth-wise convolutional kernel of size $K\times K \times M$ where the $m_{th}$ filter in ${\mathbf{W}}$ is applied to the $m_{th}$ channel in $\mathbf{IA}$ to produce the $m_{th}$ channel of the filtered output feature map ${\mathbf{OA}}$.

Depth-wise convolution has the computational cost of:

\begin{equation}
K \cdot K \cdot M \cdot H \cdot W
\end{equation}

The sum of depth-wise and $1 \times 1$ point-wise convolutions cost:

\begin{equation}  
 K \cdot K \cdot M \cdot H \cdot W +  M \cdot N \cdot  H \cdot W
\end{equation}

By expressing standard convolution as depth-wise separable convolution, we get a reduction in the computation by:

\begin{eqnarray*}
&&\frac{K \cdot K \cdot M \cdot H \cdot W +  M \cdot N \cdot  H \cdot W}{K \cdot K \cdot M \cdot N \cdot H \cdot W} \\
&=&\frac{1}{N} + \frac{1}{K \cdot K}
\end{eqnarray*}

Thus, the DS-CNN model inherently reduces the number of operations carried out during inference.

\begin{figure*}[!h]
	\begin{center}
		\includegraphics[width=\textwidth]{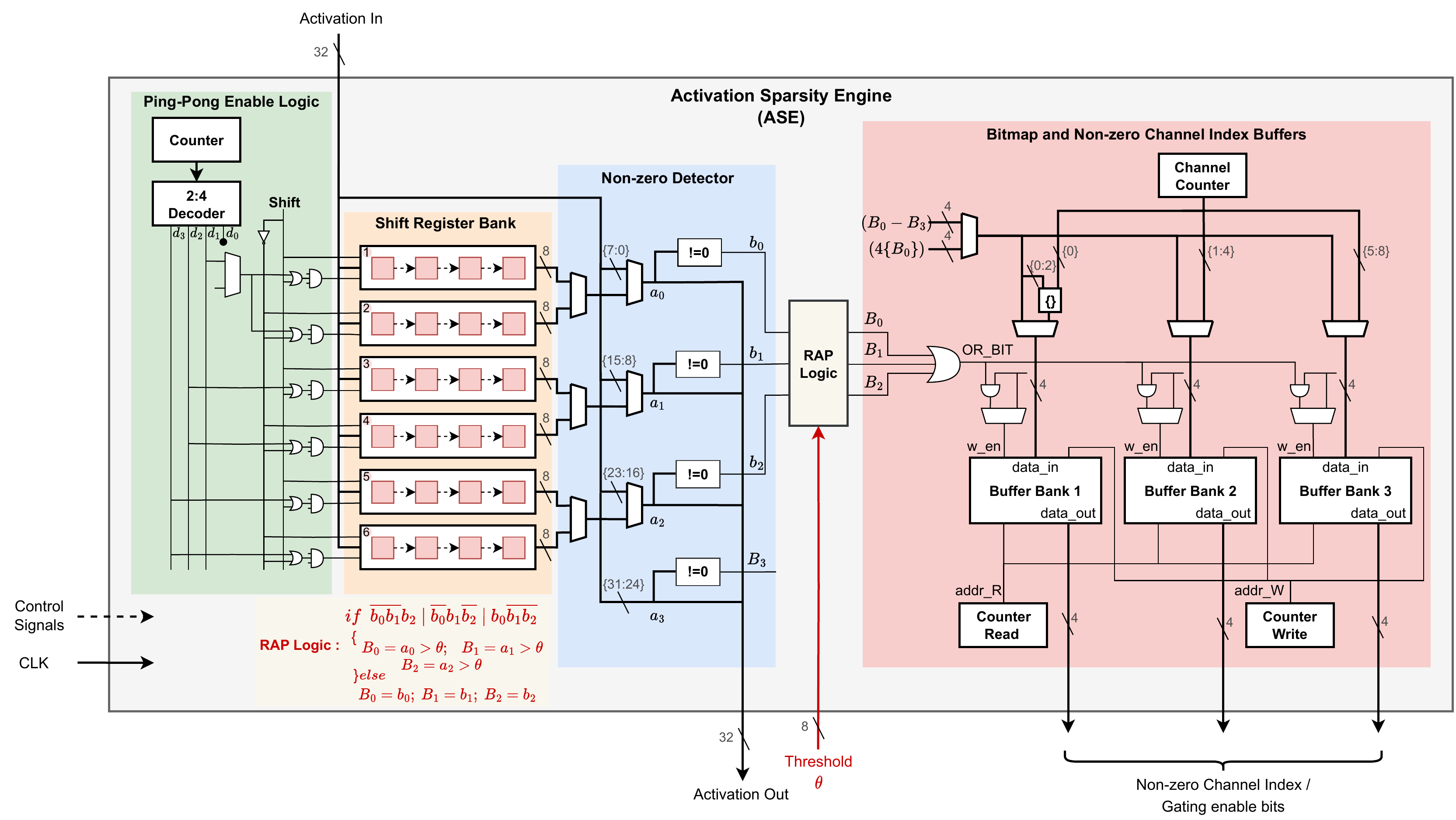}
	\end{center}
	\caption{Activation sparsity engine.} 
	\label{fig:suppl_ase}
\end{figure*}

\section{Activation sparsity engine design}
\label{sec:suppl_ase}

The ASE consists of five major blocks, as shown in Fig. \ref{fig:suppl_ase}. \begin{itemize}[leftmargin=*]
    \item \textbf{Shift Register Bank:} Consists of six parallel shift registers (SRs), with even and odd SR pairs ping-ponging to minimize the time required to write a block of data from the GLB-MEM to cache. To begin with, all even SRs are in input mode, while all odd SRs are in shift/output mode, and the functionality is inverted in the following epoch. In the input mode, we load 32b activations parallelly into four registers of a SR. In the output/shift mode, 8b data is serially shifted into the non-zero detector block.
    \item \textbf{Ping-Pong Enable Logic:} Comprises a counter and 2:4 decoder to activate appropriate SRs in the shift register bank. The contents of the SRs are shifted when the $shift$ control signal is asserted.
    \item \textbf{Non-Zero Detector:} The IAs read from the shift register bank are compared with zero to generate bitmap ($b_{0}$ to $b_{2}$, $B_{3}$), which is $1$ when the IA value is not equal to $0$ and $0$ otherwise, thus recording the position of zeros. The bitmap bits ($b_{0}$ to $b_{2}$) are sent to the run-time activation pruning (RAP) logic to perform activation pruning and generate a new set of bitmap bits ($B_{0}$ to $B_{2}$). Then the bits ($B_{0}$ to $B_{2}$) obtained from RAP are ‘OR’ed to generate $OR\_BIT$, which is required for zero skipping in the PW layer. The bitmap bits, along with the $OR\_BIT$, are fed to the succeeding bitmap and non-zero channel index buffer module. 
    \item \textbf{Run-time Activation Pruning (RAP):} As the name suggests, the RAP performs activation pruning during inference to improve the throughput and minimize computation latency. In our implementation, IA in the PW layer is tiled with each sizing $1\times M$, and three rows of the PE array simultaneously process three such tiles, i.e., $3\times M$ block of IA is processed in a single epoch. However, if all activations in a column of $3\times M$ block are zeros, that particular input channel or column is skipped entirely during the computation. Suppose there is only one non-zero activation in a column and the rest are $0$; that activation value is pruned if it's below the predetermined threshold ($\theta$) so that the column may be skipped during processing. The threshold value is precomputed during the training phase based on the input and hidden layers activation distribution. A pseudo-code of the RAP logic is shown in Fig. \ref{fig:suppl_ase}. When any one of the bitmap bits ($b_{0}$ to $b_{2}$) is $1$, and the remaining are $0s$, then the activation values ($a_{0}$ to $a_{2}$) are compared with the threshold. If the value $a_{k}$ is lesser than the threshold, then the corresponding bitmap bit $B_{k}$ is made $0$. If a column has more than one non-zero element, then the input bitmap bits to the RAP ($b_{0}$ to $b_{2}$) are retained. RAP improves the throughput by ${\approx}(10-15)\%$, and offers accuracy vs energy/latency tradeoff which can be configured at run-time.
    \item \textbf{Bitmap and non-zero channel index buffer:}  It consists of three buffers and a channel counter. When the $ OR\_BIT$ is $1$ in the PW layer, we save the corresponding input channel number obtained from the channel counter in the buffer. This information is useful during the computation as we perform computation only on input channels recorded in the buffer and skip the rest. In other layers (such as DW and CONV), we store the bitmap bits in buffers, which are later used to enable/disable the appropriate data gating registers in PEs (c.f Fig. \ref{fig:suppl_pe1}-\ref{fig:suppl_pe3}).
\end{itemize}
The IAs are stored in a compressed format in the cache using the five blocks discussed above. The bitmap bits and the input channel numbers saved in buffers aid in power saving through data-gating and latency reduction through cycle skipping.

\section{Quantization Operation in Post-processing Module}
\label{sec:suppl_ppm}
The quantization scheme is derived from \cite{Jacob2017} and can be represented by the following equation:
\begin{equation}
Q(x) = round(\dfrac{\alpha\times x}{2^{\beta}})
\label{eqn:quantisation_equation}
\end{equation}

\begin{figure*}[!h]
\centering\includegraphics[width=\textwidth]{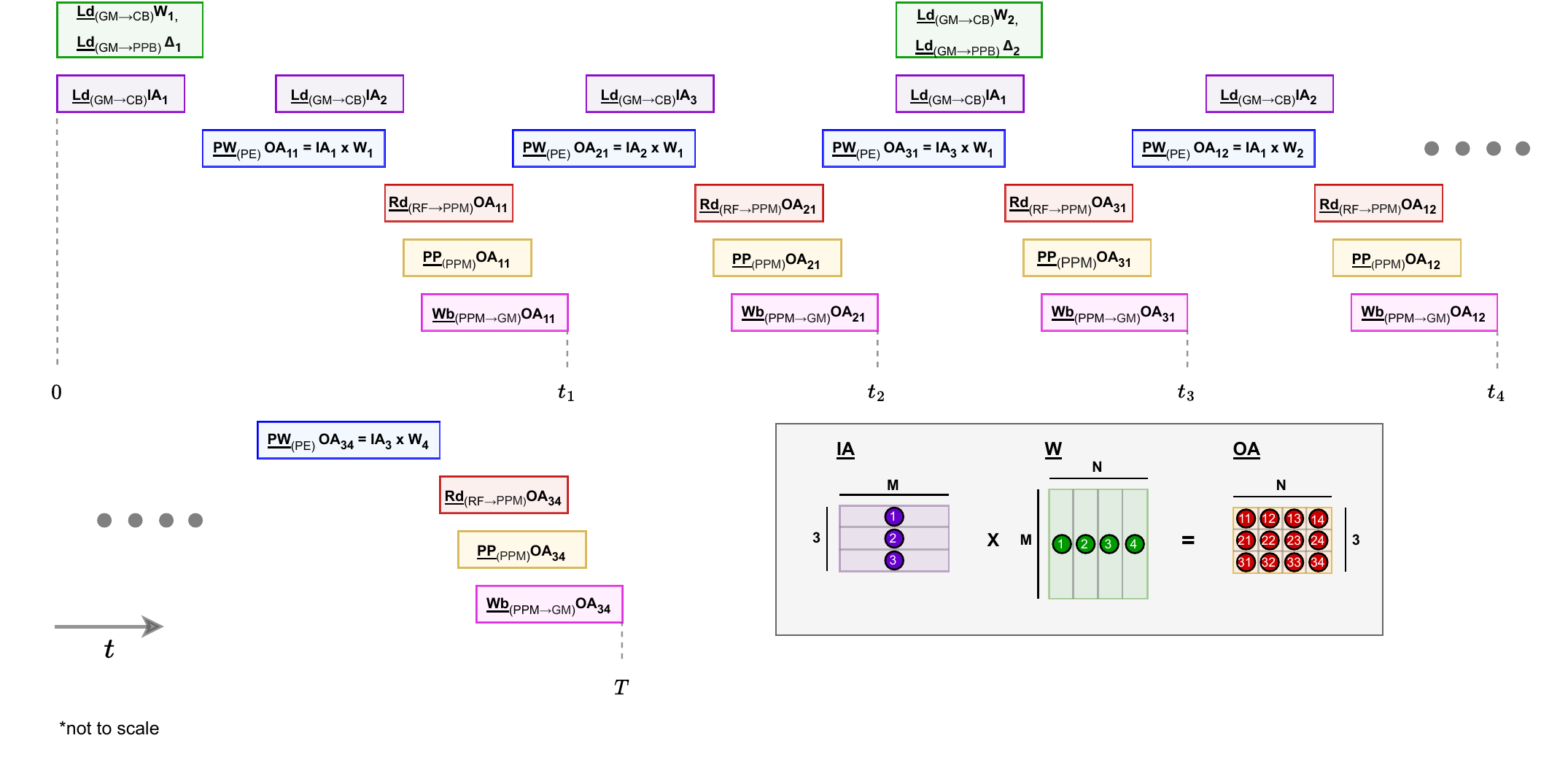}
\centering\caption{PW layer timing diagram. $Ld_{(GM\rightarrow CB)}$ denotes memory transfer from the GLB-MEM to cache,  $Ld_{(GM\rightarrow PPB)}$ denotes memory transfer from the GLB-MEM to the post-processing module buffer, $PW_{PE}$ denotes the point-wise layer computation inside the PE array, $Rd_{(RF\rightarrow PPM)}$ represents moving Psum from RF inside PE to the post-processing module, $PP_{PPM}$ denotes the post-processing operation, and $Wb_{(PPM\rightarrow GM)}$ denotes writeback of the final quantized output to the GLB-MEM. Computation and memory transfers are overlapped to minimize delay and improve MAC utilization. $\Delta$ represents post-processing parameters bias $(b)$, $\alpha$, and $\beta$.}
\label{fig:intra_layer_pipeline}
\end{figure*}

Where $Q$ is the quantized representation of input $x$, $\alpha$ and $\beta$ represent dyadic scaling parameters computed offline. 
\par The rounding operation can be mimicked in a hardware-friendly way as follows:
\begin{equation}
round(\dfrac{\alpha\times x}{2^{\beta}}) \approx floor \left(\frac{\alpha \times (x) + 2^ {\beta -1}}{2^\beta}\right)
\label{eqn:rounding_equation}
\end{equation}
\par Dyadic scaling ensures that $\alpha$ and $\beta$ are fixed point integers; hence, the multiplication and division operations in Eqn. \ref{eqn:rounding_equation} can be efficiently implemented using an integer (fixed point) multiplier and a shifter. From offline accuracy analysis, scales $\alpha$ and $\beta$ are set to 8b. Rounding operation can be visualized as adding 0.5, which is \begin{math} \frac{1}{2^1}\end{math} to the ReLU activated output. This can be thought of adding 1 left shifted by \begin{math} \beta - 1\end{math} times to the ReLU activated outputs, and then right shifting it \begin{math} \beta\end{math} times. 
Post rounding, the result is clamped within the 8b range ( -128 to +127 for 8b signed and 0 to 255 for 8b unsigned). In addition, if the ReLU output is zero, the downstream computations are disabled by data-gating registers to improve energy efficiency. 

\begin{figure*}[h!]
  \centering
  \begin{subfigure}[b]{1\textwidth}
    \includegraphics[width=16cm]{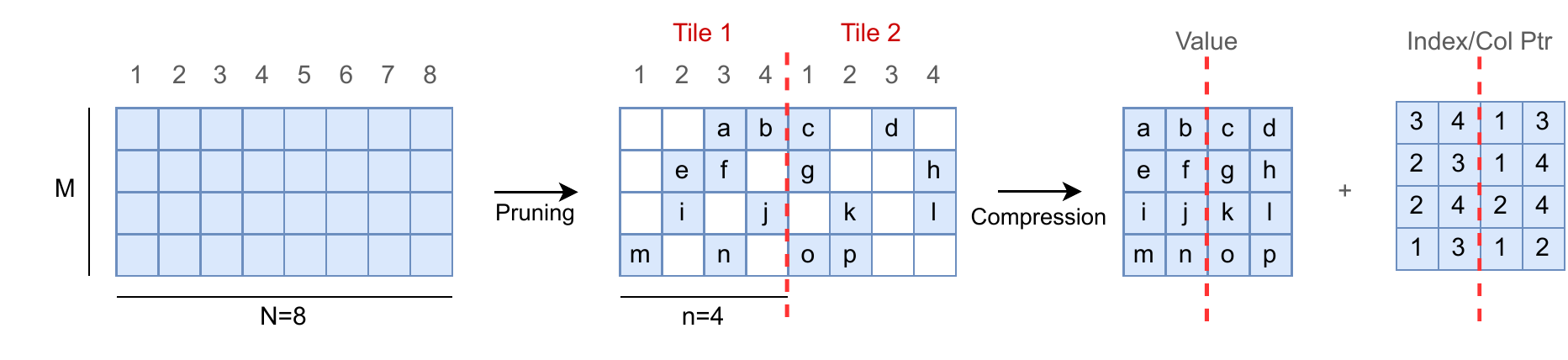}
    \caption{ }
  \end{subfigure}
    \begin{subfigure}[b]{0.8\textwidth}
    \includegraphics[width=13cm]{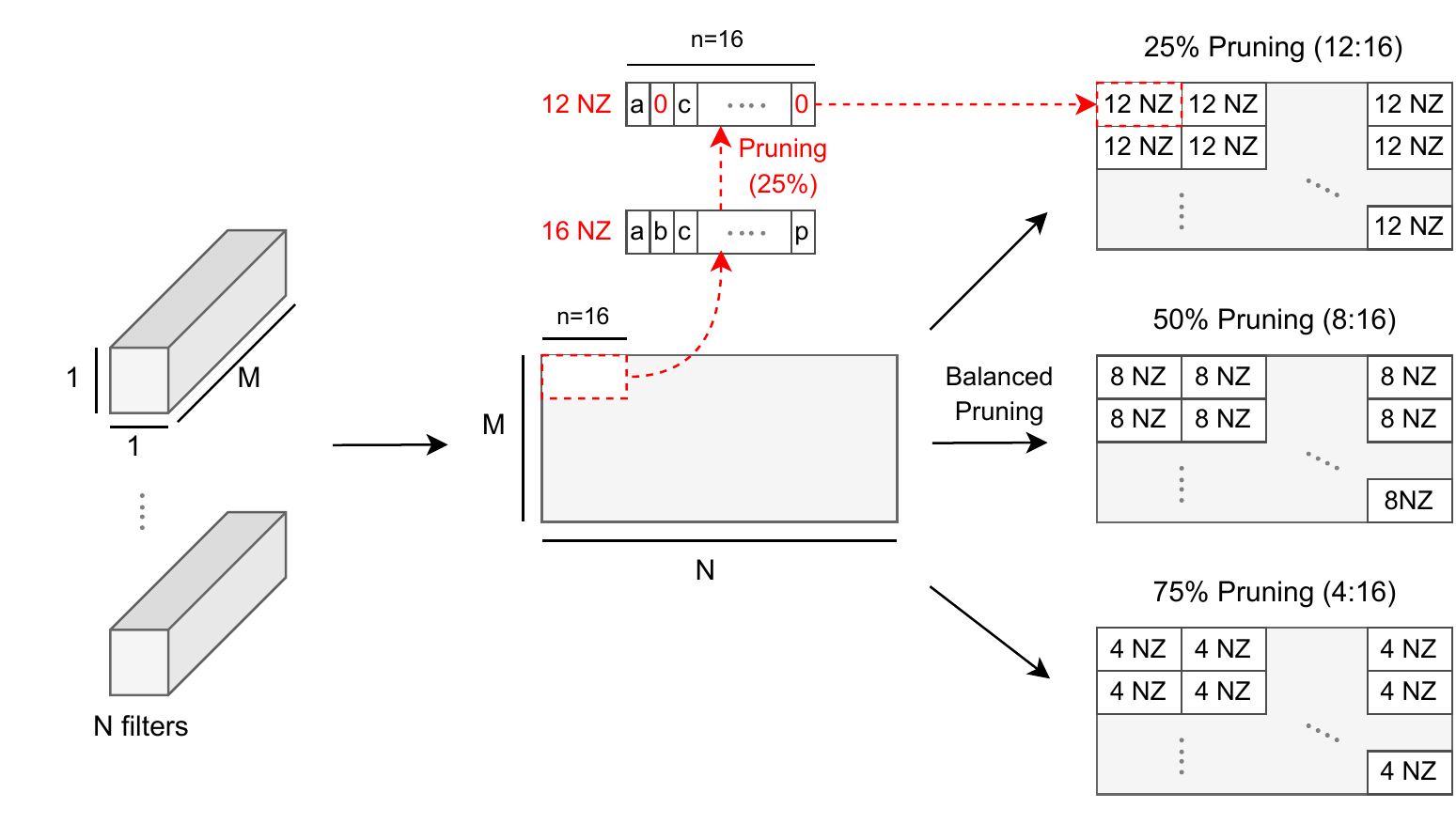}
    \caption{}
  \end{subfigure}
  \caption{Hardware-aware balanced weight pruning illustration. (a) Decomposing a sparse weight matrix into a compressed dense matrix containing only non-zero elements and an index matrix. (b) Balanced pruning strategy for different pruning ratios.}
  \label{fig:w_pruning}
\end{figure*}


\begin{figure}[h]
	\begin{center}
			\includegraphics[width=\columnwidth]{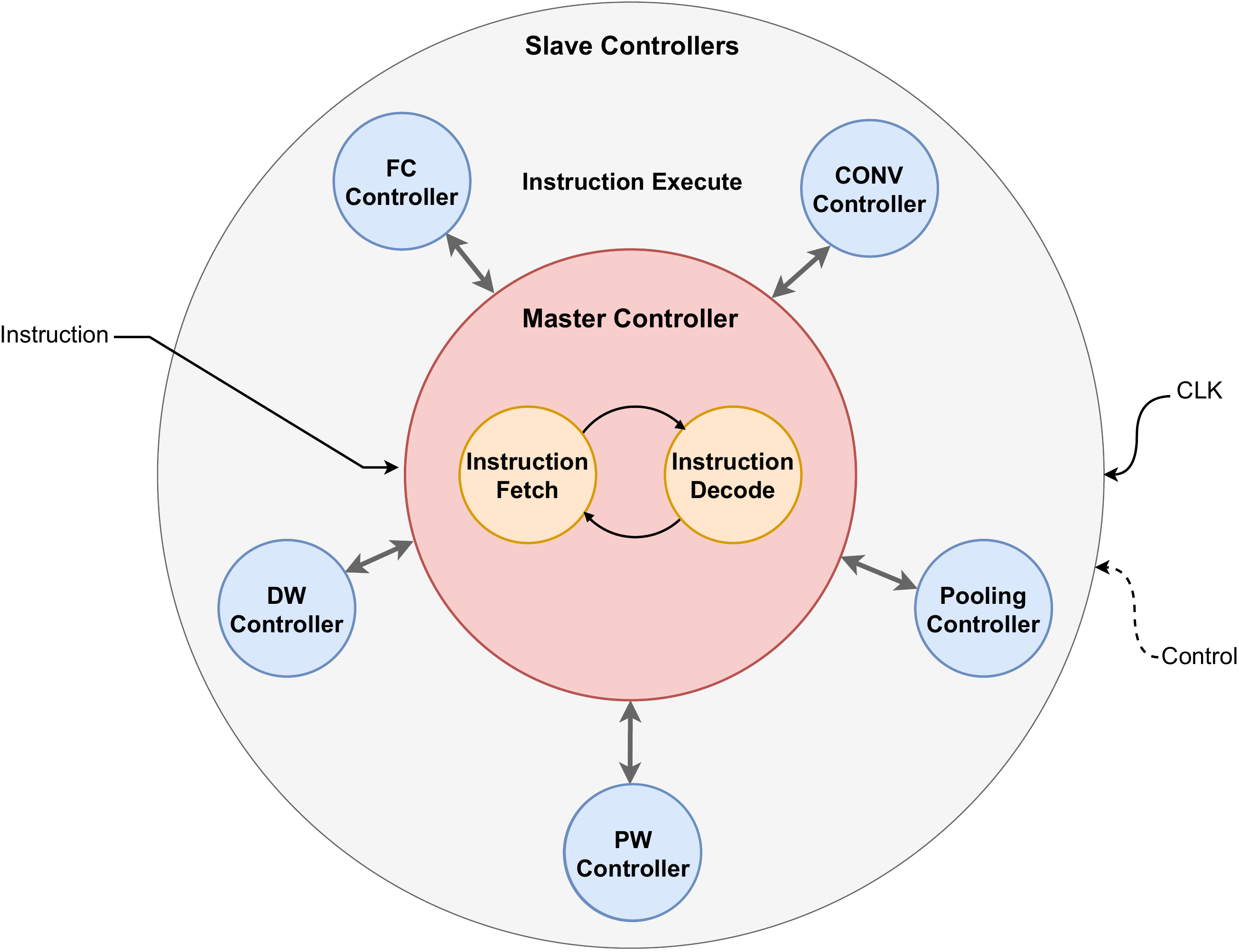}
	\end{center}
	\caption{Top-level controller} 
	\label{fig:master_controller}
\end{figure}

\section{Intra-layer Pipelining}
\label{sec:intra_layer_pipelining_suppl}
Maintaining high MAC utilization is crucial for sustaining low latency and high throughput. This can be accomplished by pipelining the operations of a certain layer, as shown in Fig \ref{fig:intra_layer_pipeline}. $ Ld$ mnemonic denotes the memory access from the GLB-MEM; $PW$ mnemonic indicates the point-wise operation of an IA tile with a W tile in the PE array; $Rd$ denotes transferring the contents of RF from PE to PPM after the final accumulation, $PP$ denotes post-processing, and $Wb$ represents the final writeback of OA after ReLU and quantization inside the PPM, to the GLB-MEM. The IA and W can be loaded into the cache concurrently since there are separate activation and parameter memory banks. After the data tile is loaded into the cache, the PE array starts the MAC computation, and the final accumulated output is sent to the PPM. The data fetch from the PE RF, PPM operation, and final output writeback is pipelined. Furthermore, the $Ld$ operation of the next IA and/or W tiles overlaps with the PW computation of the previous tile to amortize the GLB-MEM access latency. Since activation load (next tile) and store (last tile) can happen simultaneously,  the activation bank is a dual address ported memory. Here we show the timing diagram of the PW layer as it is the most computationally intensive layer. Other layers in the network, such as DW and CONV, employ the same pipelining principle. The overall MAC utilization across the PW layers is $86\%$ for the MobileNetV1$^{\dagger}$ model.

\begin{figure*}[h!]
\centering\includegraphics[width=13cm]{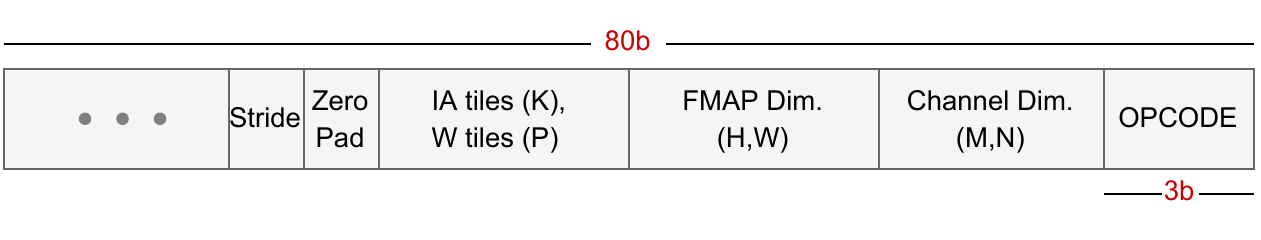}
\centering\caption{An illustration of an instruction.}
\label{fig:suppl_inst_set}
\end{figure*} 

\section{Hardware-aware balanced weight pruning}
\label{sec:hardware_aware_pruning_suppl}
We propose a hardware-aware balanced weight pruning technique to reduce memory storage and access and improve energy efficiency and throughput for the PW layer. The pruning is done during the training stage in an iterative manner to have minimum accuracy degradation \cite{Han2015}. Initially, the weights are divided into tiles of size $M\times n$ as shown in Fig. \ref{fig:w_pruning}(a), and for ease of explanation, we have considered $n=4$. A fixed number of weights are pruned in each tile row based on the magnitude, leading to structured sparsity, which can be efficiently exploited in our hardware. We employ the CSR scheme with modifications to store W in compressed form. The CSR scheme uses a set of non-zero values, bounds/row pointer, and index/column pointer to represent compressed information. The bounds/row pointer determines the number of non-zero elements present in a row of the sparse matrix, and the index provides the column index of the non-zero value. In our pruning scheme, since all the rows in a sparse matrix tile have the same number of non-zero elements, we don’t have to store bounds explicitly. It further eliminates the PE workload imbalance problem, and the start location of a particular tile in memory can be easily identified. The index in our implementation provides a non-zero column index for a specific tile and not the entire sparse matrix requiring $log_{2}(n)$ bits instead of $log_{2}(N)$ in the conventional CSR approach. In our implementation, $n$ is 16, requiring a 4b index, and the values of the weights are quantized to 8b. Thus, each non-zero weight element is represented by a 12b value-index pair.
\par The balanced pruning methodology ensures uniform zero/non-zero weight distribution across the weight tiles, thereby eliminating the workload imbalance problem. Without the balanced pruning strategy, the non-zero weights would be non-uniformly distributed across different weight tiles processed by different PEs resulting in workload imbalance, and the overall performance is limited by the PE with the heaviest workload. The PEs with low non-zero weight tile distribution completes their execution faster and has to be stalled for the slowest one (the PE with high non-zero weight tile distribution).
\par 
Fig. \ref{fig:w_pruning}(b) shows the balanced weight pruning strategy employed in RAMAN with $n=16$ for different pruning ratios.
\begin{figure*}[h!]
  \centering
  \begin{subfigure}[b]{0.35\textwidth}
    \includegraphics[width=\textwidth]{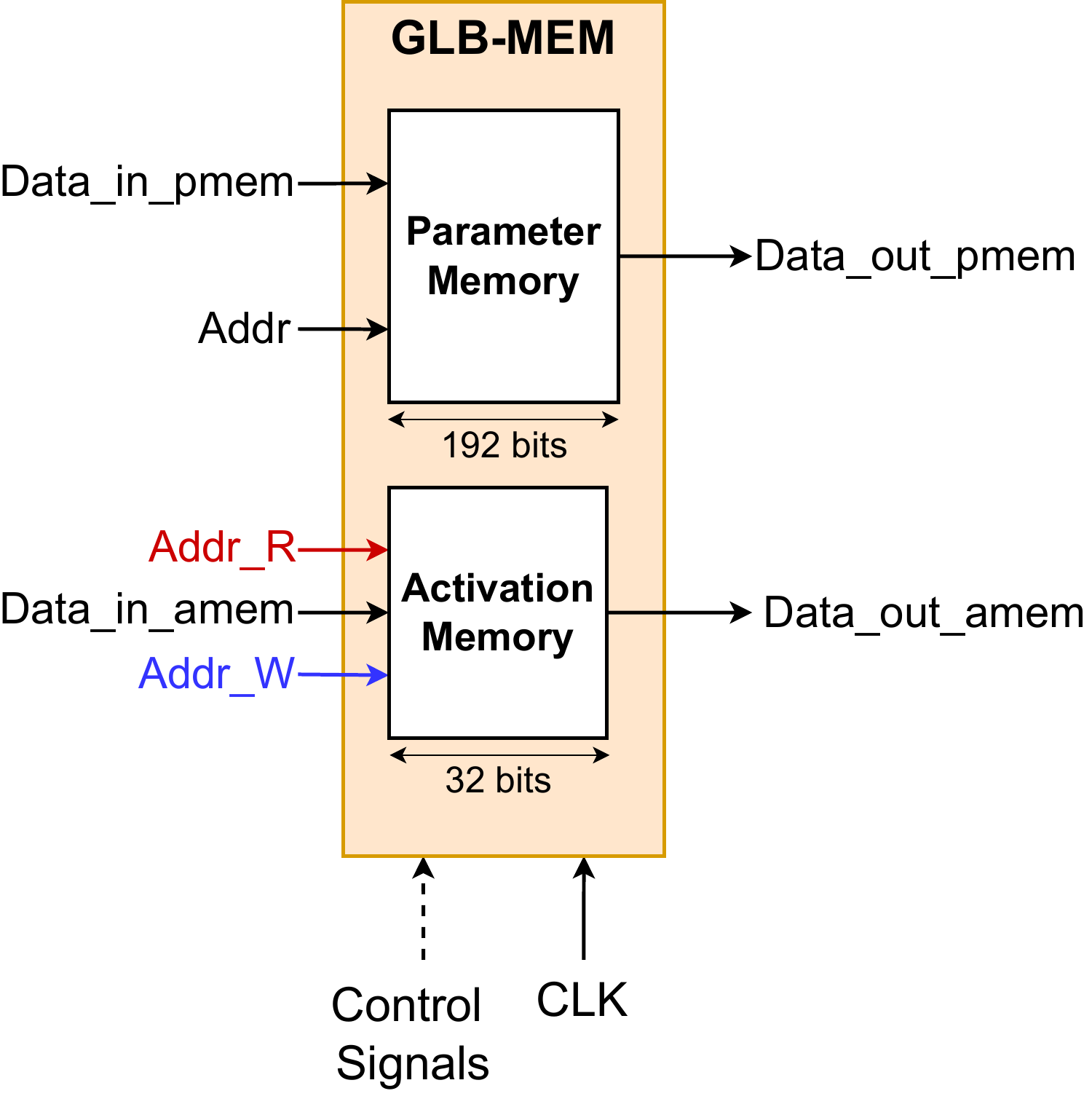}
    \caption{ }
  \end{subfigure}
    \begin{subfigure}[b]{0.3\textwidth}
    \includegraphics[width=\textwidth]{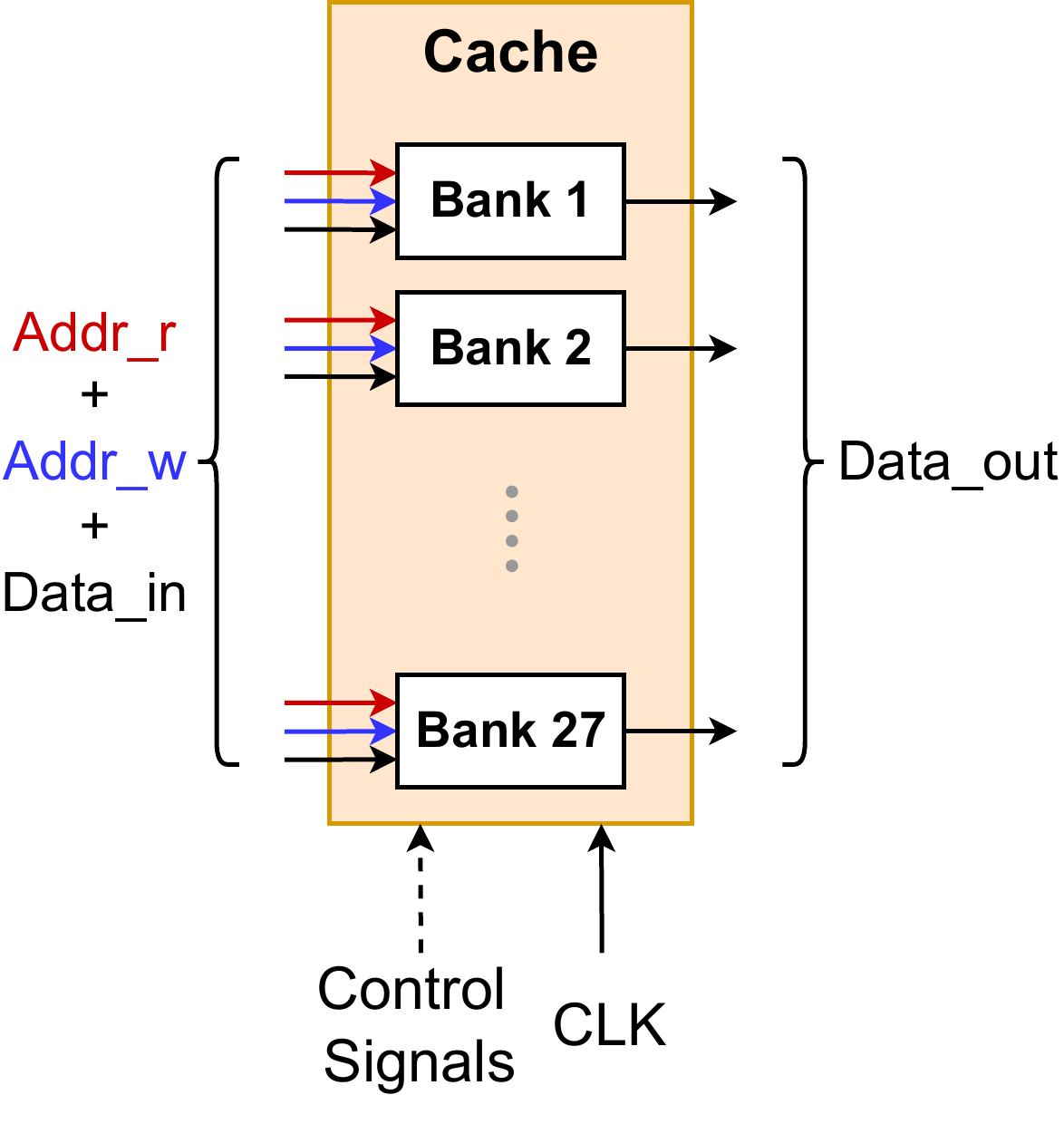}
    \caption{}
  \end{subfigure}
  \caption{Memory blocks of RAMAN: (a) Global Memory and (b) Activation and parameter cache. }
  \label{fig:suppl_glb_cache_mem}
\end{figure*}
\begin{figure*}[h!]
	\begin{center}
		\includegraphics[width=12cm]{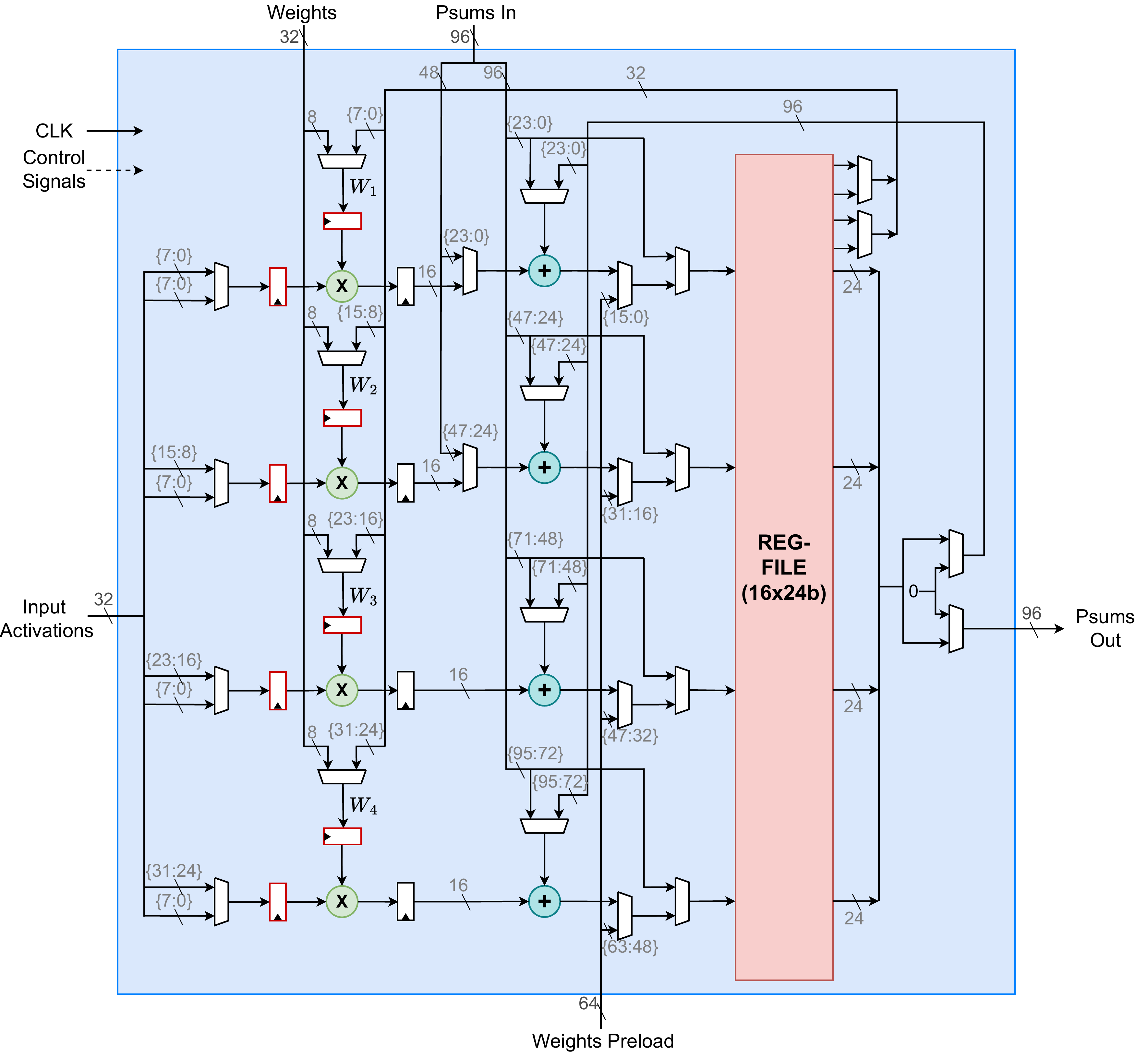}
	\end{center}
	\caption{Processing Element (Type-1)} 
	\label{fig:suppl_pe1}
\end{figure*}


\begin{figure*}[h]
	\begin{center}
		\includegraphics[width=14cm]{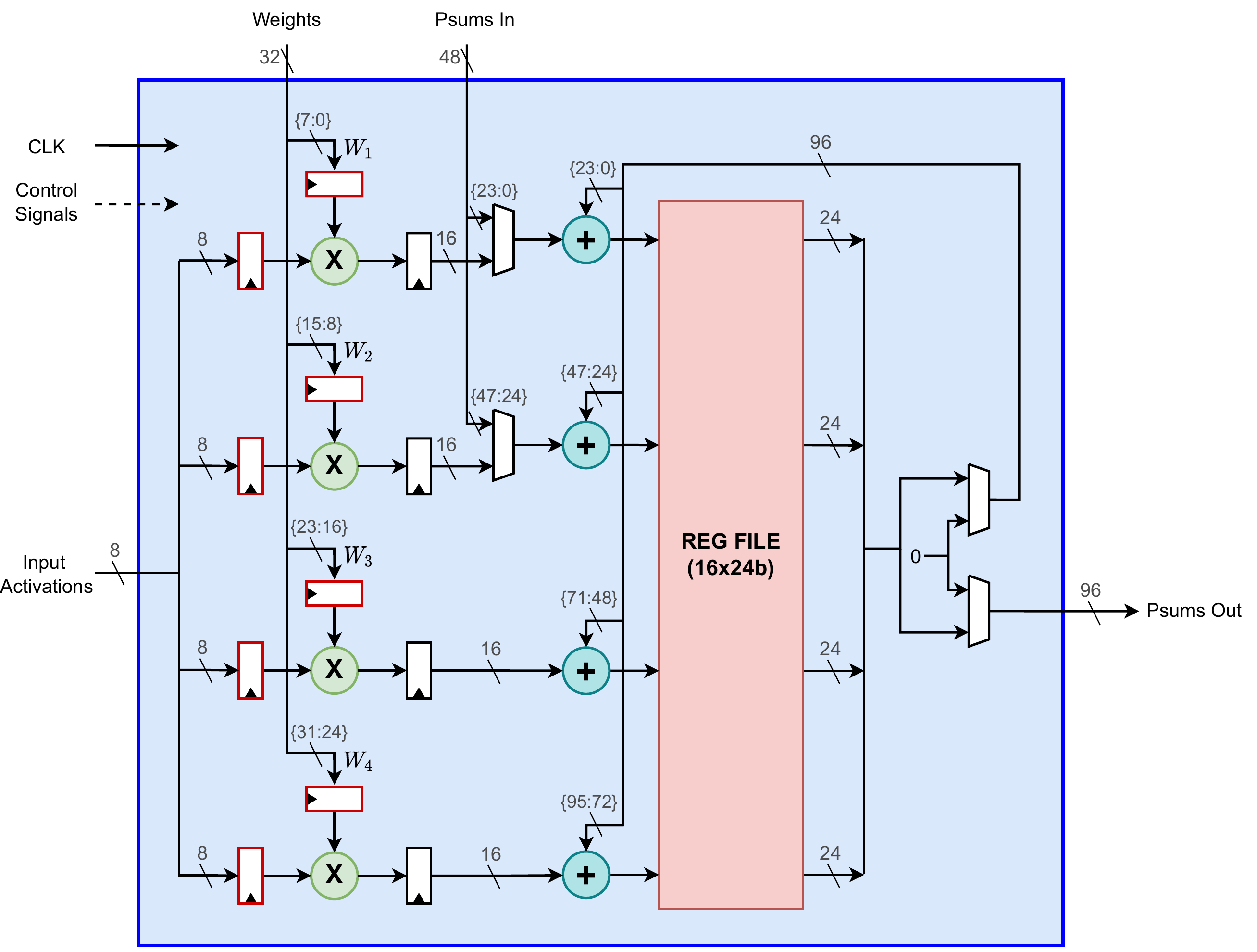}
	\end{center}
	\caption{Processing Element (Type-2)} 
	\label{fig:suppl_pe2}
\end{figure*}

\begin{figure*}[h!]
	\begin{center}
		\includegraphics[width=14cm]{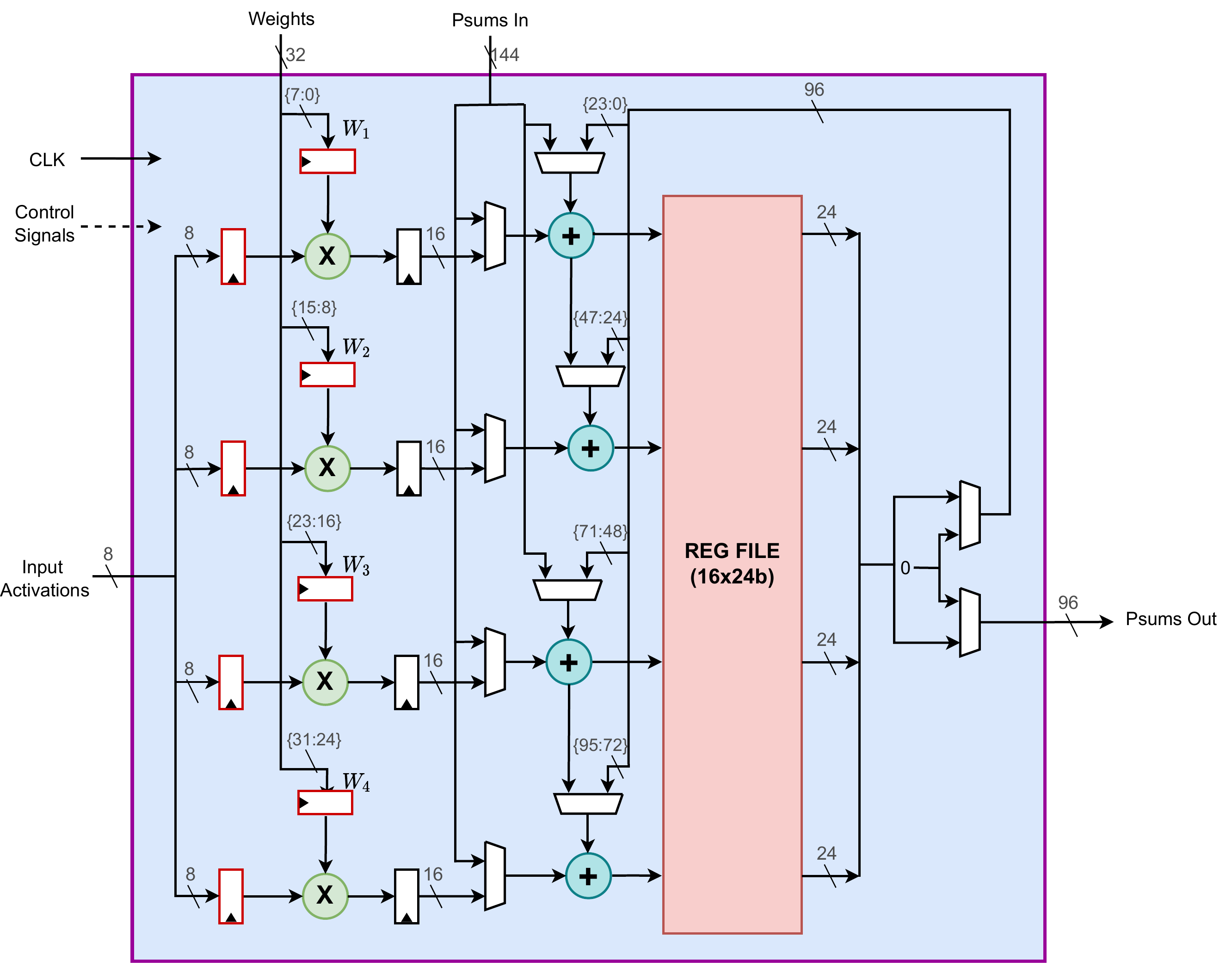}
	\end{center}
	\caption{Processing Element (Type-3)} 
	\label{fig:suppl_pe3}
\end{figure*}

\section{Top-level controller Implementation}
\label{sec:controller}
The top-level controller employs a master-slave hierarchy to implement a layer execution. The slave controller carries out instruction execution, whereas the master controller is in charge of fetching and decoding instructions. In the instruction fetch stage, a layer-specific instruction scheduled for execution is fetched from the instruction memory and sent to the top-level master controller. In the instruction decode stage, the fetched instruction is subsequently decoded to retrieve the opcode and layer configuration data. Control is then handed to the slave controller based on the opcode of a given instruction, which activates one of the five slave controllers according to the layer type specified by the opcode. Following the instruction execution, the master controller is activated to repeat the instruction fetch, decode and execute stages for the next instruction. 
\par 
A RAMAN instruction is 80 bits wide as shown in Fig. \ref{fig:suppl_inst_set}, with a 3b opcode, and the rest of the bits are dedicated to storing the information related to each layer, like the zero pad, stride, IA dimension, channel dimension, etc.

 \section{Global and cache memory }
The global and cache memory block diagrams are shown in Fig. \ref{fig:suppl_glb_cache_mem}. A description of the architecture is provided in the main document.

 \section{Processing Element}
The detailed architectures of the processing elements of type-1 to type-3 are shown in Figs. \ref{fig:suppl_pe1}-\ref{fig:suppl_pe3}.  A description of the PE architecture is provided in the main document.

\section{Dataset and model statistics}
\label{sec:suppl_ms}

We employ RAMAN for auditory and computer vision (CV) tasks.  The dataset used for the auditory task is the Speech Command dataset \cite{Warden2018Speech} for Keyword spotting (KWS) application. The dataset is a collection of one-second utterances of simple keywords. The DS-CNN$^{\dagger}$ model shown in Table \ref{model_stats_table}(a) is trained on the dataset for 12 classes as listed in Table \ref{tab:class_table}(a). 
\par The dataset used for the CV task is the Visual Wake Word (VWW) dataset \cite{Chowdhery2019VWW} for person/no-person detection in an image. The dataset is a set of \begin{math} 224 \times 224 \end{math} RGB images which are down-sampled to \begin{math} 96 \times 96 \end{math} and converted to grey-scaled for hardware deployment. The VWW dataset is used to train MobileNetV1$^{\dagger}$ model shown in Table \ref{model_stats_table}(b) with two classes listed in Table \ref{tab:class_table}(b).
\par The model statistics of the targeted applications that gave maximum test accuracy in software are shown in Table \ref{model_stats_table}. DS-CNN$^{\dagger}$ model consists of eight depth-wise Separable (DWS) layers comprising depth-wise and point-wise layers with the input cochleagram of size $30\times 32$ extracted from cascade of asymmetric resonators \cite{Xucar2018}. The global average pooling (GAP) layer reduces the input to the fully connected (FC) layer where the classification takes place. It is to be noted that the DS-CNN$^{\dagger}$ model is much smaller in size as compared to the  MobileNetV1$^{\dagger}$ model which has 13 DWS layers sandwiched between CONV and FC layer. The input to the MobileNetV1$^{\dagger}$ model is a pre-processed image of size $96 \times 96$ image derived from a $224 \times 224$ RGB image which is downsized and greyscaled as part of the pre-processing. It is also worthwhile to note that the pre-processing is being done offline and not at run-time.

\begin{table}[H]
\caption{Classes of the target applications}
\renewcommand{\arraystretch}{1.2}
\begin{subtable}[b]{0.45\textwidth}
\centering
\begin{tabular}[b]{|cccc|}
\hline
\multicolumn{4}{|c|}{\textbf{KWS Classes}}                                                                                        \\ \hline
\multicolumn{1}{|c|}{\textbf{Class}} & \multicolumn{1}{c|}{\textbf{Keyword}} & \multicolumn{1}{c|}{\textbf{Class}} & \textbf{Keyword} \\ \hline
\multicolumn{1}{|c|}{0}              & \multicolumn{1}{c|}{Unknown}        & \multicolumn{1}{c|}{6}              & Left           \\ \hline
\multicolumn{1}{|c|}{1}              & \multicolumn{1}{c|}{Silence}        & \multicolumn{1}{c|}{7}              & Right          \\ \hline
\multicolumn{1}{|c|}{2}              & \multicolumn{1}{c|}{Yes}            & \multicolumn{1}{c|}{8}              & On             \\ \hline
\multicolumn{1}{|c|}{3}              & \multicolumn{1}{c|}{No}             & \multicolumn{1}{c|}{9}              & Off            \\ \hline
\multicolumn{1}{|c|}{4}              & \multicolumn{1}{c|}{Up}             & \multicolumn{1}{c|}{10}             & Stop           \\ \hline
\multicolumn{1}{|c|}{5}              & \multicolumn{1}{c|}{Down}           & \multicolumn{1}{c|}{11}             & Go             \\ \hline
\end{tabular}
\caption{KWS}
\label{kws_class_table}
\end{subtable}
\begin{subtable}[b]{0.45\textwidth}
\centering
\begin{tabular}[b]{|cc|}
\hline
\multicolumn{2}{|c|}{\textbf{VWW Classes}}                   \\ \hline
\multicolumn{1}{|c|}{\textbf{Class}} & \textbf{Status}       \\ \hline
\multicolumn{1}{|c|}{0}              & Person \\ \hline
\multicolumn{1}{|c|}{1}              & Not a person     \\ \hline
\end{tabular}
\caption{VWW} 
\label{vww_class_table}
\end{subtable}
\label{tab:class_table}
\end {table}

\begin{table}[H]
\caption{Model statistics implemented on RAMAN.}
\renewcommand{\arraystretch}{1.2}
\begin{subtable}[b]{0.45\textwidth}
\centering
\begin{tabular}{|c|c|c|}
\hline
\textbf{Layer}  & \textbf{$H_{in}$ $\times$ $W_{in}$ $\times$ $M$} & \textbf{$H_{out}$ $\times$ $W_{out}$ $\times$ $N$} \\ \hline
\textbf{CONV} & 30 $\times$ 32 $\times$ 1           & 28 $\times$ 30  $\times$ 64  \\ \hline
\textbf{DW}     & 28 $\times$ 30 $\times$ 64           & 28 $\times$ 30  $\times$ 64  \\ \hline
\textbf{PW}     & 28 $\times$ 30 $\times$ 64             & 28 $\times$ 30 $\times$ 64 \\ \hline
\textbf{DW}     & 28 $\times$ 30 $\times$ 64          & 14 $\times$ 15 $\times$ 64   \\ \hline
\textbf{PW}     & 14 $\times$ 15 $\times$ 64             & 14 $\times$ 15 $\times$ 64  \\ \hline
\textbf{DW}     & 14 $\times$ 15 $\times$ 64           & 7 $\times$ 8 $\times$ 64  \\ \hline
\textbf{PW}     & 7 $\times$ 8 $\times$ 64            & 7 $\times$ 8 $\times$ 64  \\ \hline
\textbf{DW}     & 7 $\times$ 8 $\times$ 64           & 4 $\times$ 4 $\times$ 64  \\ \hline
\textbf{PW}     & 4 $\times$ 4 $\times$ 64            & 4 $\times$ 4 $\times$ 64  \\ \hline
\textbf{DW $\times$ 4}     & 4 $\times$ 4 $\times$ 64           & 4 $\times$ 4 $\times$ 64  \\ \hline
\textbf{PW $\times$ 4}     & 4 $\times$ 4 $\times$ 64            & 4 $\times$ 4 $\times$ 64   \\ \hline
\textbf{GAP}     & 4 $\times$ 4 $\times$ 64            & 1 $\times$ 1 $\times$ 64    \\ \hline
\textbf{FC}     & 1 $\times$ 1 $\times$ 64            & 1 $\times$ 1 $\times$ 12  \\ \hline
\end{tabular}
\caption{DS-CNN$^{\dagger}$}
\label{kws_stats_table}
\end{subtable}
\begin{subtable}[b]{0.45\textwidth}
\centering
\begin{tabular}{|c|c|c|}
\hline
\textbf{Layer} &  $H_{in}$ $\times$ $W_{in}$ $\times$ $M$ & $H_{out}$ $\times$ $W_{out}$ $\times$ $N$  \\ \hline
\textbf{CONV} & 96 $\times$ 96 $\times$ 1             & 47  $\times$  47 $\times$ 16  \\ \hline
\textbf{DW}     & 47  $\times$  47 $\times$ 16             & 47  $\times$  47 $\times$ 16   \\ \hline
\textbf{PW}     & 47  $\times$  47 $\times$ 16              & 47  $\times$  47 $\times$ 16    \\ \hline
\textbf{DW}     & 47  $\times$  47 $\times$ 16              & 24  $\times$  24 $\times$ 16   \\ \hline
\textbf{PW}     & 24  $\times$  24 $\times$ 16              & 24  $\times$  24 $\times$ 32  \\ \hline
\textbf{DW}     & 24  $\times$  24 $\times$ 32              & 24  $\times$  24 $\times$ 32   \\ \hline
\textbf{PW}     & 24  $\times$  24 $\times$ 32              & 24  $\times$  24 $\times$ 32  \\ \hline
\textbf{DW}     & 24  $\times$  24 $\times$ 32              & 12  $\times$  12 $\times$ 32   \\ \hline
\textbf{PW}     & 12  $\times$  12 $\times$ 32              & 12  $\times$  12 $\times$ 64   \\ \hline
\textbf{DW}     & 12  $\times$  12 $\times$ 64              & 12  $\times$  12 $\times$ 64  \\ \hline
\textbf{PW}     & 12  $\times$  12 $\times$ 64              & 12  $\times$  12 $\times$ 64      \\ \hline
\textbf{DW}     & 12  $\times$  12 $\times$ 64              & 12  $\times$  12 $\times$ 64     \\ \hline
\textbf{PW}     & 12  $\times$  12 $\times$ 64              & 12  $\times$  12 $\times$ 128    \\ \hline
\textbf{DW $\times$ 5}     & 12  $\times$  12 $\times$ 128              & 12  $\times$  12 $\times$ 128     \\ \hline
\textbf{PW $\times$ 5}     & 12  $\times$  12 $\times$ 128              & 12  $\times$  12 $\times$ 128 \\ \hline
\textbf{DW}     & 12 $\times$ 12 $\times$ 128              & 6 $\times$ 6 $\times$ 128       \\ \hline
\textbf{PW}     & 6 $\times$ 6 $\times$ 128                & 6 $\times$ 6 $\times$ 256      \\ \hline
\textbf{DW}     & 6 $\times$ 6 $\times$ 256                & 6 $\times$ 6 $\times$ 256        \\ \hline
\textbf{PW}     & 6 $\times$ 6$\times$ 256                & 6 $\times$ 6$\times$ 256     \\ \hline
\textbf{GAP}    & 6 $\times$ 6$\times$ 256                & 1 $\times$ 1$\times$ 256\\ \hline
\textbf{FC}     & 1 $\times$ 1$\times$ 256                & 1 $\times$ 1$\times$ 2    \\ \hline
\end{tabular}
\caption{MobileNetV1$^{\dagger}$} 
\label{vww_stats_table}
\end{subtable}
\label{model_stats_table}
\end {table}

\newpage
\section{Performance Analysis}
\label{sec:pa_suppl}
In this section, we analyze memory accesses throughout the hierarchy separately as reads and writes in both DS-CNN$^{\dagger}$ and MobileNetV1$^{\dagger}$ models. Also we try to analyze the reduction of reads and writes with and without IA sparsity, with and without parameter pruning. Additionally, we try to analyze the effects of IA sparsity and parameter pruning on MAC operations. To benchmark our results, we have taken 110 audio signatures and 110 images as inputs to average the operations (OPs) and memory accesses involved in inferring them. In our work, a read or write to the GLB-MEM or to the cache bank is considered to be a memory access. 

\begin{table}[H]
\caption{Activation Global Memory Access in KB}

\renewcommand{\arraystretch}{1.2}
\begin{subtable}[b]{0.45\textwidth}
\centering
\begin{tabular}{|c|cc|}
\hline
\multirow{2}{*}{\textbf{DS-CNN$^{\dagger}$}} & \multicolumn{2}{c|}{\textbf{\begin{tabular}[c]{@{}c@{}}Activation Global \\ Memory (in KB)\end{tabular}}} \\ \cline{2-3} 
                                & \multicolumn{1}{c|}{\textbf{Read}}                            & \textbf{Write}                            \\ \hline
\textbf{Total CONV}           & \multicolumn{1}{c|}{0.96}                                     & 53.76                                      \\ \hline
\textbf{Total DW}               & \multicolumn{1}{c|}{128.64}                                   & 75.904                                     \\ \hline
\textbf{Total PW}               & \multicolumn{1}{c|}{75.904}                                    & 75.904                                     \\ \hline
\textbf{Total GAP}              & \multicolumn{1}{c|}{1.024}                                        & 0.064                                    \\ \hline
\textbf{Total FC}               & \multicolumn{1}{c|}{0.064}                                   & 0.012                                    \\ \hline
\textbf{Total}                  & \multicolumn{1}{c|}{206.592}                                   & 205.644                                   \\ \hline
\end{tabular}
\caption{DS-CNN$^{\dagger}$}
\label{kws_act_gmem}
\end{subtable}
\hspace*{0.2mm}%
\begin{subtable}[b]{0.45\textwidth}
\centering
\begin{tabular}{|c|cc|}
\hline
\multirow{2}{*}{\textbf{MobileNetV1$^{\dagger}$}} & \multicolumn{2}{c|}{\textbf{\begin{tabular}[c]{@{}c@{}}Activation Global \\ Memory (in KB)\end{tabular}}} \\ \cline{2-3} 
                                & \multicolumn{1}{c|}{\textbf{Read}}                            & \textbf{Write}                            \\ \hline
\textbf{Total CONV}           & \multicolumn{1}{c|}{9.216}                                        & 35.344                                      \\ \hline
\textbf{Total DW}               & \multicolumn{1}{c|}{245.792}                                   & 192.016                                    \\ \hline
\textbf{Total PW}               & \multicolumn{1}{c|}{334.864}                                   & 219.664                                 \\ \hline
\textbf{Total GAP}              & \multicolumn{1}{c|}{9.216}                                        & 0.256                                      \\ \hline
\textbf{Total FC}               & \multicolumn{1}{c|}{0.256}                                     & 0.002                                     \\ \hline
\textbf{Total}                  & \multicolumn{1}{c|}{599.344}                                 & 447.282                                    \\ \hline
\end{tabular}
\caption{MobileNetV1$^{\dagger}$} 
\label{vww_act_gmem}
\end{subtable}
\label{act_gmem_table}
\end {table}

DS-CNN$^{\dagger}$ being a statiscally smaller model as compared to MobileNetV1$^{\dagger}$, the activation GLB-MEM reads and writes in Table \ref{kws_act_gmem} are much lower than that shown in Table \ref{vww_act_gmem}.  It can also be observed that the depth-wise separable (DWS) layers constitute maximum activation GLB-MEM accesses (both reads and writes) which justfies our focus on optimising DWS layer dataflow and the hardware architecture to the maximum extent possible.     



\begin{table}[H]
\caption{Activation Cache Access in KB}
\renewcommand{\arraystretch}{1.2}
\begin{subtable}[b]{0.45\textwidth}
\centering
\begin{tabular}{|cccc|}
\hline
\multicolumn{4}{|c|}{\textbf{Activation Cache Access (in KB)}}                                                                                                                                                                                                                                                                                              \\ \hline
\multicolumn{2}{|c|}{\textbf{Read}}                                                                                                                                                     & \multicolumn{2}{c|}{\textbf{Write}}                                                                                                                               \\ \hline
\multicolumn{1}{|c|}{\textbf{\begin{tabular}[c]{@{}c@{}}Without\\ IA Sparsity\end{tabular}}} & \multicolumn{1}{c|}{\textbf{\begin{tabular}[c]{@{}c@{}}With\\ IA Sparsity\end{tabular}}} & \multicolumn{1}{c|}{\textbf{\begin{tabular}[c]{@{}c@{}}Without\\ IA Sparsity\end{tabular}}} & \textbf{\begin{tabular}[c]{@{}c@{}}With\\ IA Sparsity\end{tabular}} \\ \hline
\multicolumn{1}{|c|}{359.232}                                                                 & \multicolumn{1}{c|}{219.099}                                                              & \multicolumn{1}{c|}{205.568}                                                                 & 124.355                                                              \\ \hline
\end{tabular}
\caption{DS-CNN$^{\dagger}$}
\label{kws_cache_act_table}
\end{subtable}
\hspace*{0.2mm}%
\begin{subtable}[b]{0.45\textwidth}
\centering

\begin{tabular}{|cccc|}
\hline
\multicolumn{4}{|c|}{\textbf{Activation Cache Access (in KB)}}                                                                                                                                                                                                                                                                                              \\ \hline
\multicolumn{2}{|c|}{\textbf{Read}}                                                                                                                                                     & \multicolumn{2}{c|}{\textbf{Write}}                                                                                                                               \\ \hline
\multicolumn{1}{|c|}{\textbf{\begin{tabular}[c]{@{}c@{}}Without\\ IA Sparsity\end{tabular}}} & \multicolumn{1}{c|}{\textbf{\begin{tabular}[c]{@{}c@{}}With\\ IA Sparsity\end{tabular}}} & \multicolumn{1}{c|}{\textbf{\begin{tabular}[c]{@{}c@{}}Without\\ IA Sparsity\end{tabular}}} & \textbf{\begin{tabular}[c]{@{}c@{}}With\\ IA Sparsity\end{tabular}} \\ \hline
\multicolumn{1}{|c|}{978.848}                                                                 & \multicolumn{1}{c|}{538.343}                                                              & \multicolumn{1}{c|}{590.128}                                                                 & 323.36                                                              \\ \hline
\end{tabular}
\caption{MobileNetV1$^{\dagger}$} 
\label{vww_cache_act_table}
\end{subtable}
\label{cache_act_table}
\end {table}

The data shown in Table \ref{cache_act_table} are the total cache accesses made for activations. The writes are $\approx40\%$ less than reads which indicates extensive reuse of activations in the cache before being evicted out. $\approx40-45\%$ reduction in accesses is seen when IA sparsity is exploited using ASE with RAP threshold set to zero, out of which the majority of the accesses happen in DWS layers itself.

\begin{table}[H]
\caption{Activation Cache accesses breakdown}
\renewcommand{\arraystretch}{1.2}
\begin{subtable}[b]{0.5\textwidth}
\centering
\begin{tabular}{|c|cccc|}
\hline
\multirow{3}{*}{\textbf{Layer}} & \multicolumn{4}{c|}{\textbf{Activations Cache Access in KB}}                                                                               \\ \cline{2-5} 
                                & \multicolumn{2}{c|}{\textbf{With IA sparsity}}                           & \multicolumn{2}{c|}{\textbf{Without IA sparsity}}   \\ \cline{2-5} 
                                & \multicolumn{1}{c|}{\textbf{Read}} & \multicolumn{1}{c|}{\textbf{Write}} & \multicolumn{1}{c|}{\textbf{Read}} & \textbf{Write} \\ \hline
\textbf{Total CONV}           & \multicolumn{1}{c|}{0.479}        & \multicolumn{1}{c|}{0.171}         & \multicolumn{1}{c|}{2.688}         & 0.96         \\ \hline
\textbf{Total DW}               & \multicolumn{1}{c|}{171.029}      & \multicolumn{1}{c|}{76.656}        & \multicolumn{1}{c|}{280.512}      & 128.64       \\ \hline
\textbf{Total PW}               & \multicolumn{1}{c|}{47.463}       & \multicolumn{1}{c|}{47.463}        & \multicolumn{1}{c|}{75.904}        & 75.904         \\ \hline
\textbf{Total GAP}              & \multicolumn{1}{c|}{0}             & \multicolumn{1}{c|}{0}              & \multicolumn{1}{c|}{0}             & 0              \\ \hline
\textbf{Total FC}               & \multicolumn{1}{c|}{0.128}         & \multicolumn{1}{c|}{0.064}         & \multicolumn{1}{c|}{0.128}         & 0.064        \\ \hline
\textbf{Total}                  & \multicolumn{1}{c|}{219.099}      & \multicolumn{1}{c|}{124.355}       & \multicolumn{1}{c|}{359.232}      & 205.568       \\ \hline
\end{tabular}
\caption{DS-CNN$^{\dagger}$}
\label{kws_cache_act_breakdown_table}
\end{subtable}
\hspace*{0.2mm}%
\begin{subtable}[b]{0.5\textwidth}
\centering
\begin{tabular}{|c|cccc|}
\hline
\multirow{3}{*}{\textbf{Layer}} & \multicolumn{4}{c|}{\textbf{Activations Cache Access in KB}}                                                                               \\ \cline{2-5} 
                                & \multicolumn{2}{c|}{\textbf{With IA sparsity}}                           & \multicolumn{2}{c|}{\textbf{Without IA sparsity}}   \\ \cline{2-5} 
                                & \multicolumn{1}{c|}{\textbf{Read}} & \multicolumn{1}{c|}{\textbf{Write}} & \multicolumn{1}{c|}{\textbf{Read}} & \textbf{Write} \\ \hline
\textbf{Total CONV}           & \multicolumn{1}{c|}{9.852}        & \multicolumn{1}{c|}{6.708}         & \multicolumn{1}{c|}{13.536}         & 9.216         \\ \hline
\textbf{Total DW}               & \multicolumn{1}{c|}{355.661}      & \multicolumn{1}{c|}{143.822}        & \multicolumn{1}{c|}{630.192}      & 245.792      \\ \hline
\textbf{Total PW}               & \multicolumn{1}{c|}{172.574}       & \multicolumn{1}{c|}{172.574}        & \multicolumn{1}{c|}{334.864}        & 334.864         \\ \hline
\textbf{Total GAP}              & \multicolumn{1}{c|}{0}             & \multicolumn{1}{c|}{0}              & \multicolumn{1}{c|}{0}             & 0              \\ \hline
\textbf{Total FC}               & \multicolumn{1}{c|}{0.256}         & \multicolumn{1}{c|}{0.256}         & \multicolumn{1}{c|}{0.256}         & 0.256         \\ \hline
\textbf{Total}                  & \multicolumn{1}{c|}{538.343}      & \multicolumn{1}{c|}{323.36}       & \multicolumn{1}{c|}{978.848}      & 590.128      \\ \hline
\end{tabular}
\caption{MobileNetV1$^{\dagger}$} 
\label{vww_cache_act_breakdown_table}
\end{subtable}
\label{cache_act_breakdown_table}
\end {table}

Table \ref{cache_act_breakdown_table} shows the breakdown of Table \ref{cache_act_table}. Majority of the accesses are evidently done during the execution of DWS layers.
\begin{table}[H]
\caption{Number of Operations in  Million}
\renewcommand{\arraystretch}{1.2}
\begin{subtable}[b]{0.45\textwidth}
\centering
\begin{tabular}{|ccc|}
\hline
\multicolumn{3}{|c|}{\textbf{\#Operations (in Million)}}                                                                                                                                                                                                              \\ \hline
\multicolumn{1}{|c|}{\textbf{\begin{tabular}[c]{@{}c@{}}Weight Sparsity in \\ PW layer\end{tabular}}} & \multicolumn{1}{c|}{\textbf{\begin{tabular}[c]{@{}c@{}}Without \\ IA sparsity\end{tabular}}} & \textbf{\begin{tabular}[c]{@{}c@{}}With \\ IA sparsity\end{tabular}} \\ \hline
\multicolumn{1}{|c|}{\textbf{0\%}}                                                                    & \multicolumn{1}{c|}{12.052}                                                                   & 7.099                                                             \\ \hline
\multicolumn{1}{|c|}{\textbf{25\%}}                                                                   & \multicolumn{1}{c|}{9.623}                                                                   & 5.581                                                                \\ \hline
\multicolumn{1}{|c|}{\textbf{50\%}}                                                                   & \multicolumn{1}{c|}{7.194}                                                                   & 4.062                                                                \\ \hline
\multicolumn{1}{|c|}{\textbf{75\%}}                                                                   & \multicolumn{1}{c|}{4.765}                                                                   & 2.543                                                                 \\ \hline
\end{tabular}
\caption{DS-CNN$^{\dagger}$}
\label{kws_mac_ops}
\end{subtable}
\hspace*{0.2mm}%
\begin{subtable}[b]{0.45\textwidth}
\centering
\begin{tabular}{|ccc|}
\hline
\multicolumn{3}{|c|}{\textbf{\#Operations (in Million)}}                                                                                                                                                                                                              \\ \hline
\multicolumn{1}{|c|}{\textbf{\begin{tabular}[c]{@{}c@{}}Weight Sparsity in \\ PW layer\end{tabular}}} & \multicolumn{1}{c|}{\textbf{\begin{tabular}[c]{@{}c@{}}Without \\ IA sparsity\end{tabular}}} & \textbf{\begin{tabular}[c]{@{}c@{}}With \\ IA sparsity\end{tabular}} \\ \hline
\multicolumn{1}{|c|}{\textbf{0\%}}                                                                    & \multicolumn{1}{c|}{41.802}                                                                   & 21.055                                                              \\ \hline
\multicolumn{1}{|c|}{\textbf{25\%}}                                                                   & \multicolumn{1}{c|}{32.378}                                                                   & 16.385                                                               \\ \hline
\multicolumn{1}{|c|}{\textbf{50\%}}                                                                   & \multicolumn{1}{c|}{22.953}                                                                   & 11.716                                                                \\ \hline
\multicolumn{1}{|c|}{\textbf{75\%}}                                                                   & \multicolumn{1}{c|}{13.528}                                                                   & 7.046                                                                 \\ \hline
\end{tabular}
\label{vww_mac_table}
\caption{MobileNetV1$^{\dagger}$}
\end{subtable}
\label{mac_ops_table}
\end {table}

\begin{figure*}[h]
\begin{subfigure}{1\textwidth}
  \centering
    \includegraphics[width=13cm]{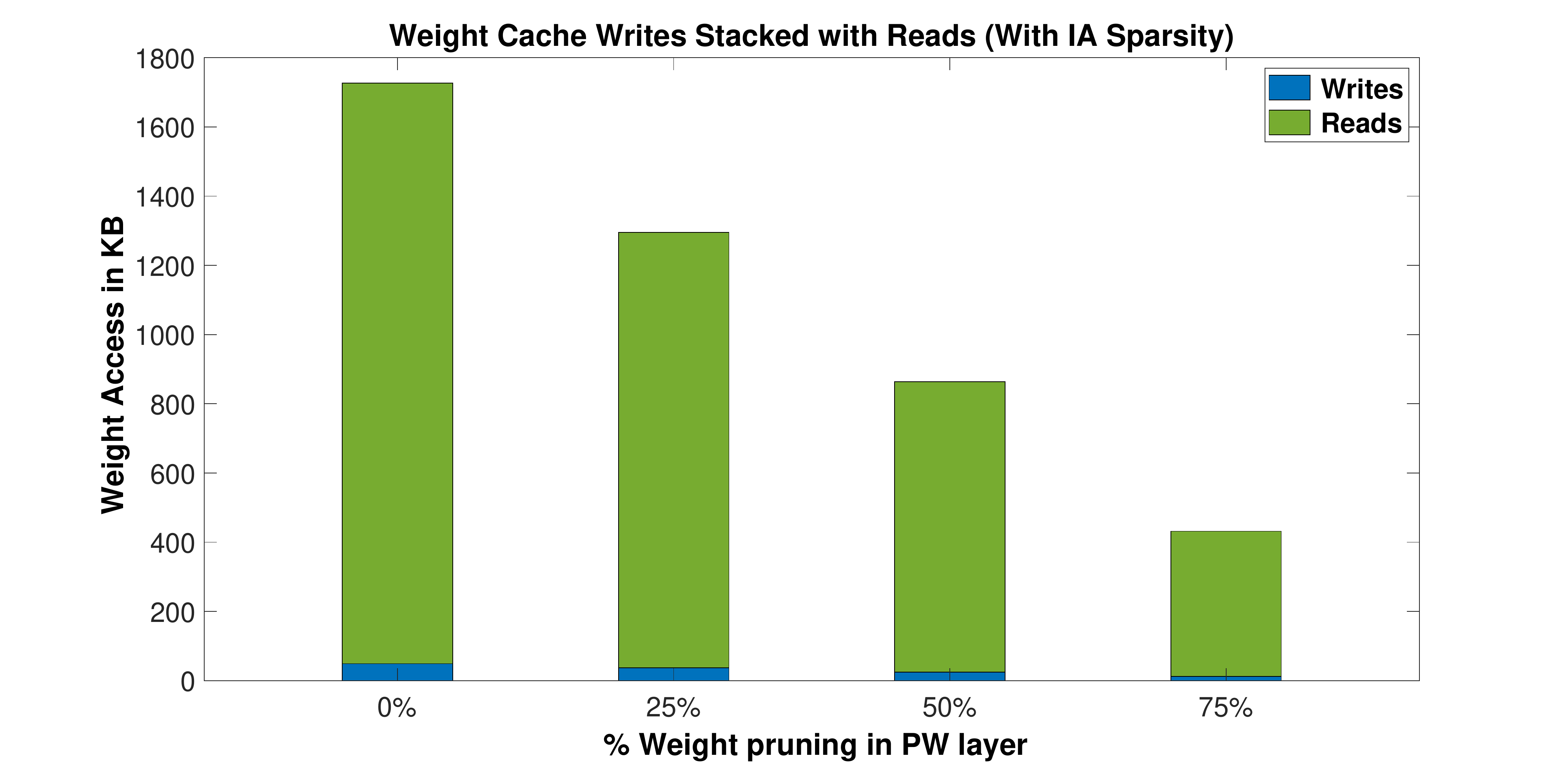}
    \caption{DS-CNN$^{\dagger}$}
    \label{cache_wgt_rw}
\end{subfigure}
\begin{subfigure}{1\textwidth}
  \centering
    \includegraphics[width=13cm]{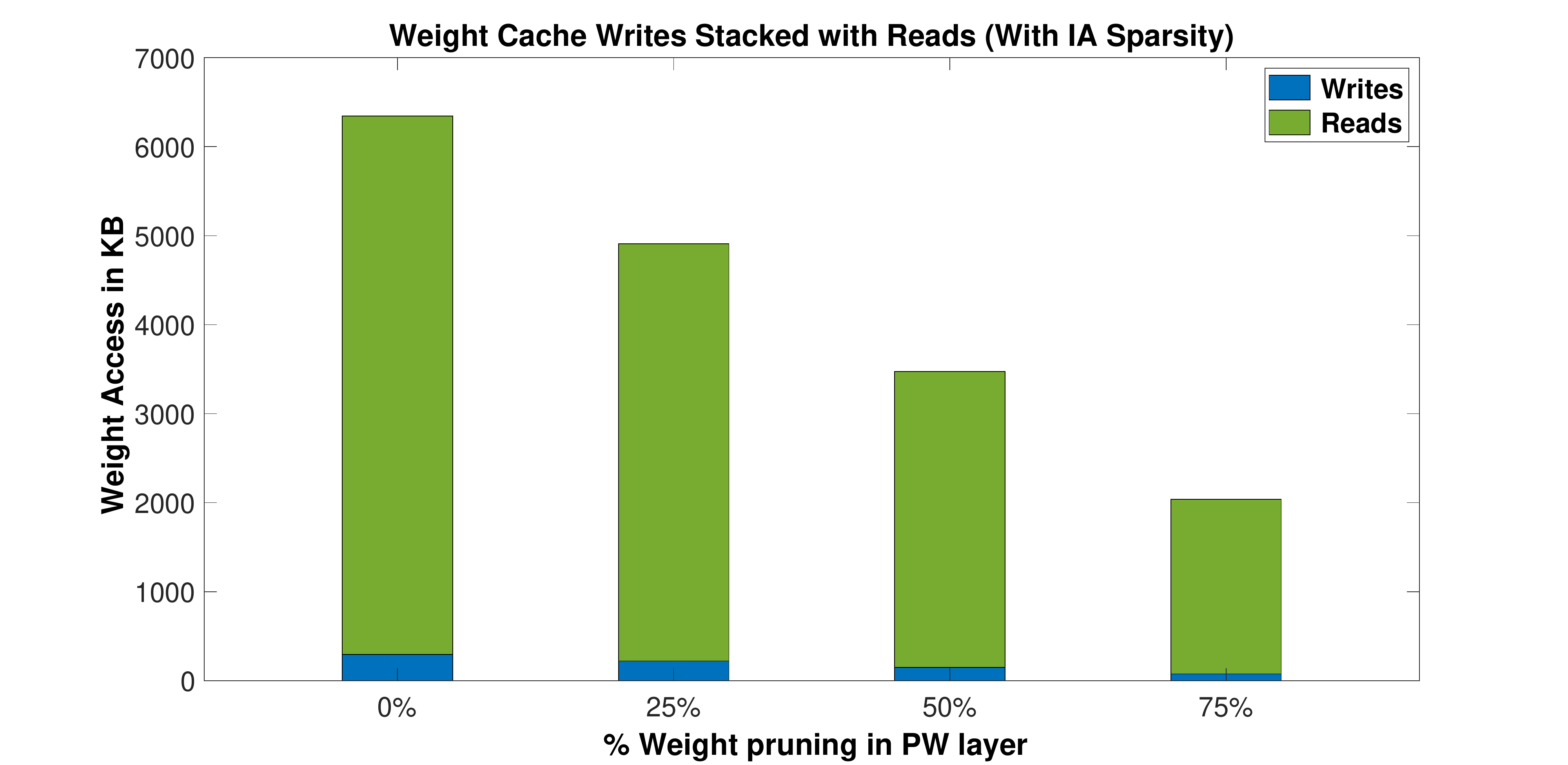}
    \caption{MobileNetV1$^{\dagger}$}
    \label{cache_wgt_rw}
\end{subfigure}
\caption{Weight Cache Access in KB}
\label{cache_wgt_rw_kws_vww}
\end{figure*}

Table \ref{mac_ops_table} shows the number of operations for both cases, with and without input sparsity. The baseline again would be a model without exploiting input and weight sparsity. We find a significant reduction of operations with IA sparsity when coupled with parameter pruning. We find that operations reduce by $\approx54\%$ when the parameters are 25\% pruned (12.05 Mil. vs 5.58 Mil.) whereas it further reduces by  another $\approx55\%$ when parameters are 75\% pruned (5.58 Mil. vs 2.54 Mil.). It should be noted that even without IA sparsity, just parameter pruning also gives considerable reductions in operations. For example, operations reduce by 21\% when compared to a 25\% pruned model (9.62 Mil. vs 12.05 Mil.) and 61\% when compared to a 75\% pruned model (4.76 Mil. vs 12.05 Mil.). PW layers dominate other layers regarding the number of operations and contribute to nearly 90\% 


Fig. \ref{cache_wgt_rw_kws_vww} depicts the cache accesses (reads and writes) involved in the parameter space. According to the dataflow described in the main document, cache bank is used only to store PW layer parameters. Hence, the plot are accesses related to PW layers in the model. The reads for any given pruning percentage is much higher than writes, which implies that there is considerable reuse of weights present in the cache. Also, the reads and writes reduce as the pruning percentage increases. 

\begin{figure*}[h]
\begin{subfigure}{1\textwidth}
  \centering
     \includegraphics[width = 13cm]{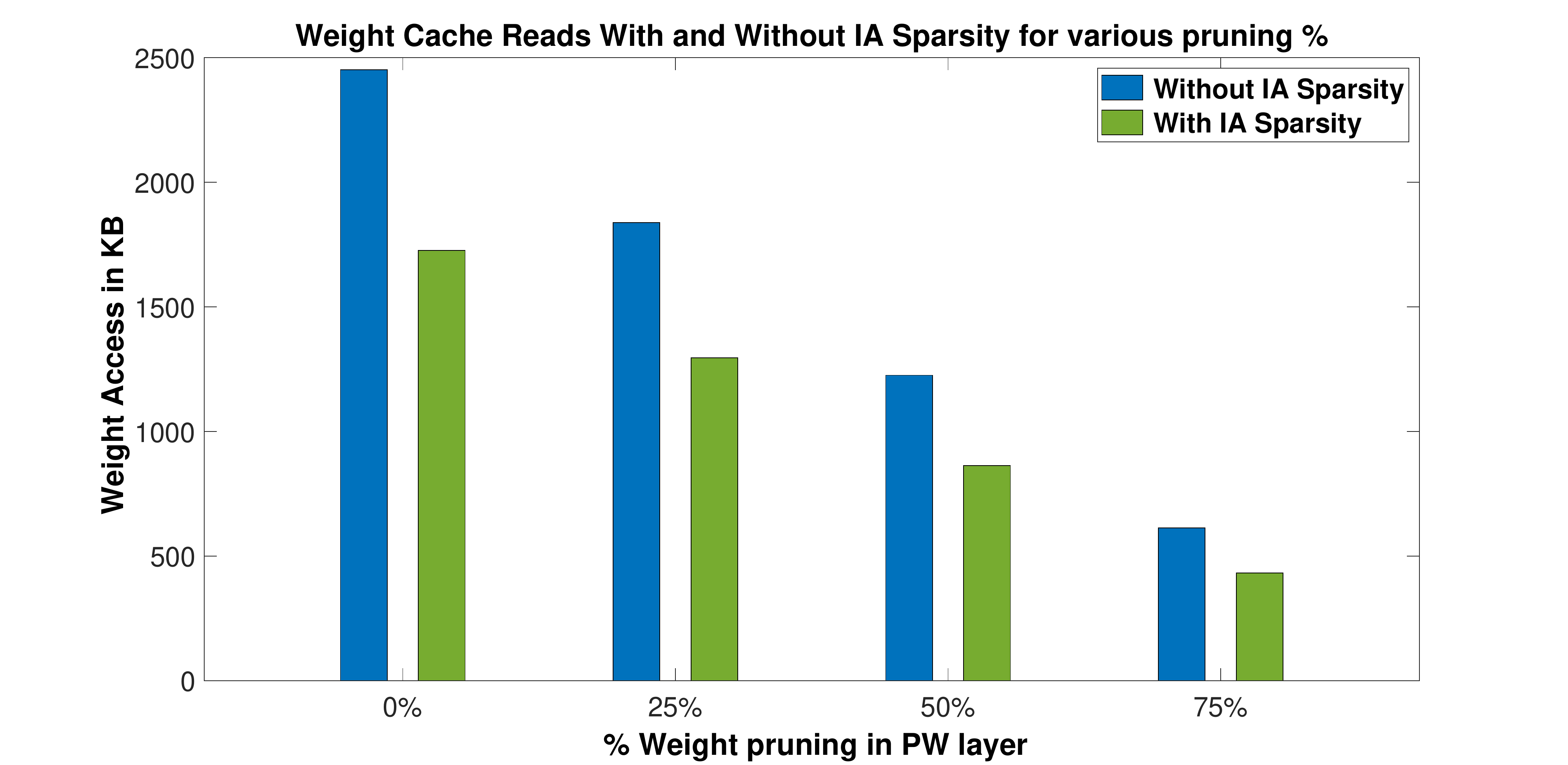}
    \caption{DS-CNN$^{\dagger}$}
    \label{cache_wgt_ips_kws}
\end{subfigure}
\begin{subfigure}{1\textwidth}
  \centering
    \includegraphics[width = 13cm]{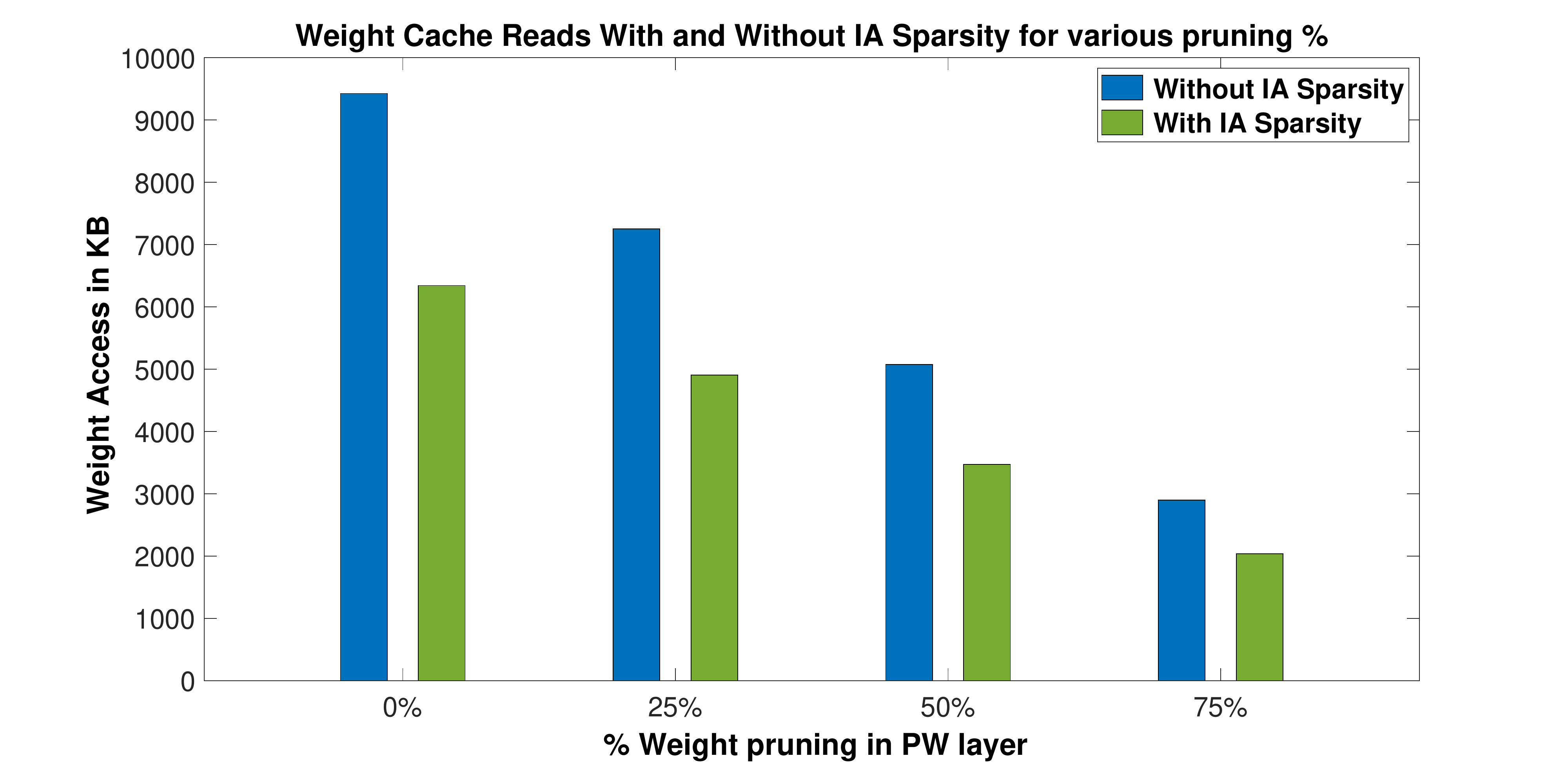}
    \caption{MobileNetV1$^{\dagger}$}
    \label{cache_wgt_ips_vww}
\end{subfigure}
\caption{Weight Cache Reads in KB}
\label{cache_wgt_ips_kws_vww}
\end{figure*}

\begin{figure}[h]
  \centering
    \includegraphics[width=\columnwidth]{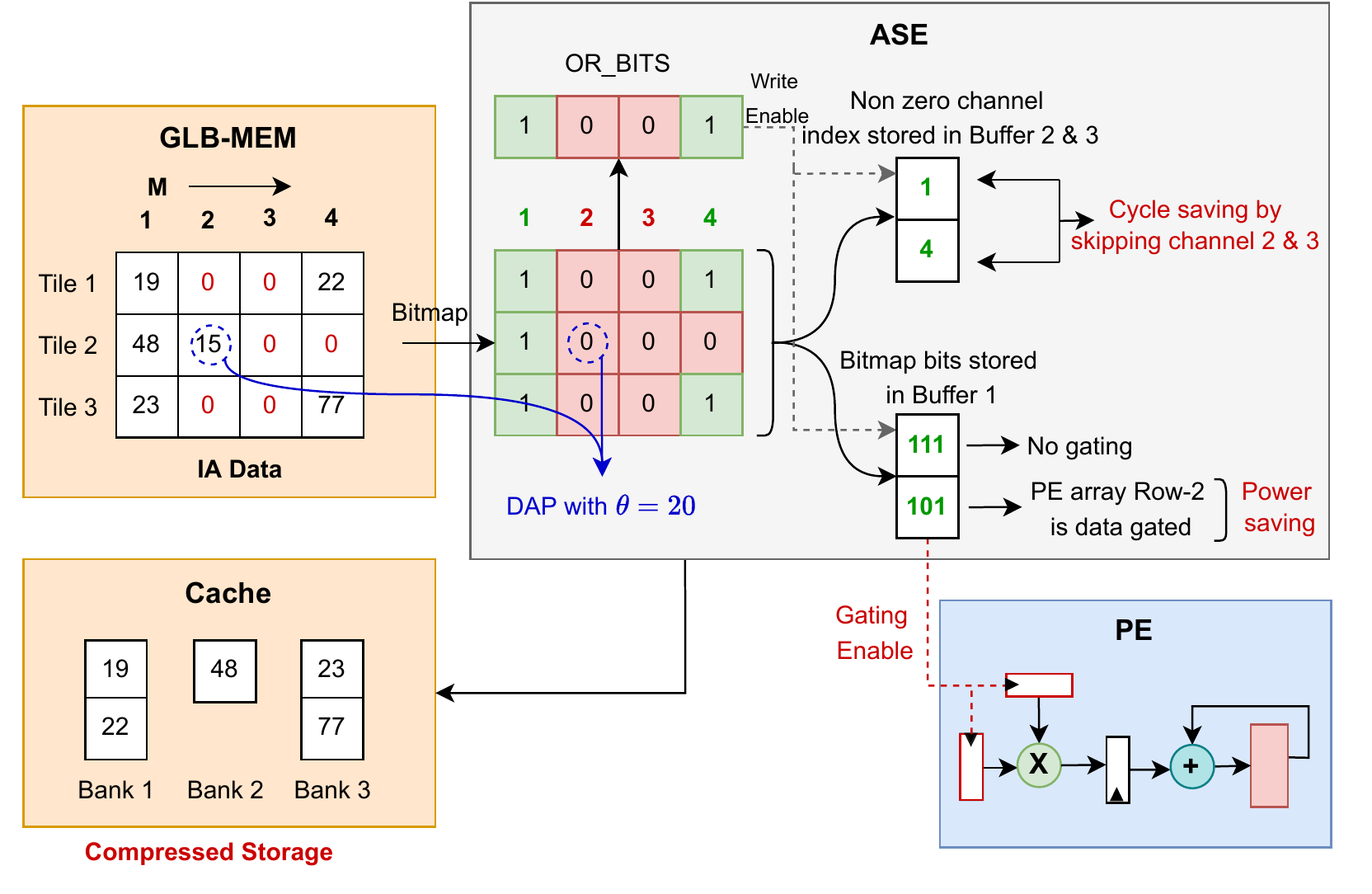}

  \caption{ASE illustration for the PW layer.}
  \label{fig:ase_illustration_pw}
\end{figure}

Fig. \ref{cache_wgt_ips_kws_vww} depicts the effect of ASE on the memory accesses in parameter reads. The parameters are not read from the cache when all the elements of a column are zero (for instance columns 2 and 3 in Fig. \ref{fig:ase_illustration_pw}). This reduces the parameter reads to a great extent leading to considerable latency and power saving. This plot shows the effectiveness of ASE in not only exploiting input activation sparsity but also reducing the parameter reads to a great extent.

\begin{figure*}[h!]
  \centering

  \begin{subfigure}[b]{0.33\textwidth}
    \includegraphics[width=\textwidth]{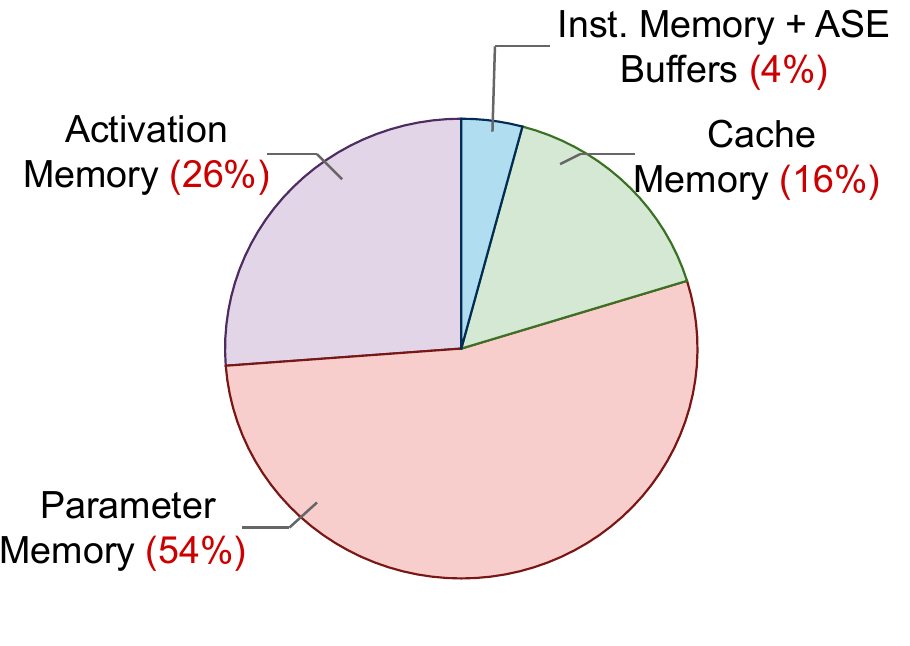}
    \caption{ }
  \end{subfigure}
    \begin{subfigure}[b]{0.36\textwidth}
    \includegraphics[width=\textwidth]{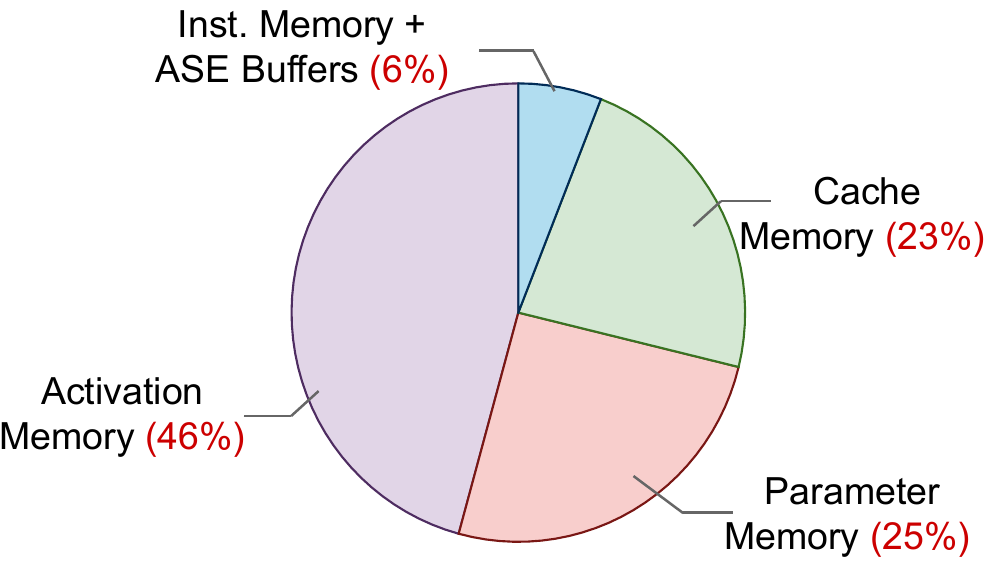}
    \caption{}
  \end{subfigure}

  \caption{Memory breakdown of RAMAN for (a) MobileNetV1$^{\dagger}$ model and (b) DS-CNN$^{\dagger}$ model for $75\%$ weight pruning in the PW layers. }
  \label{fig:mem_breakdown}
\end{figure*}
\begin{figure*}[h!]
  \centering

  \begin{subfigure}[b]{0.4\textwidth}
    \includegraphics[width=\textwidth]{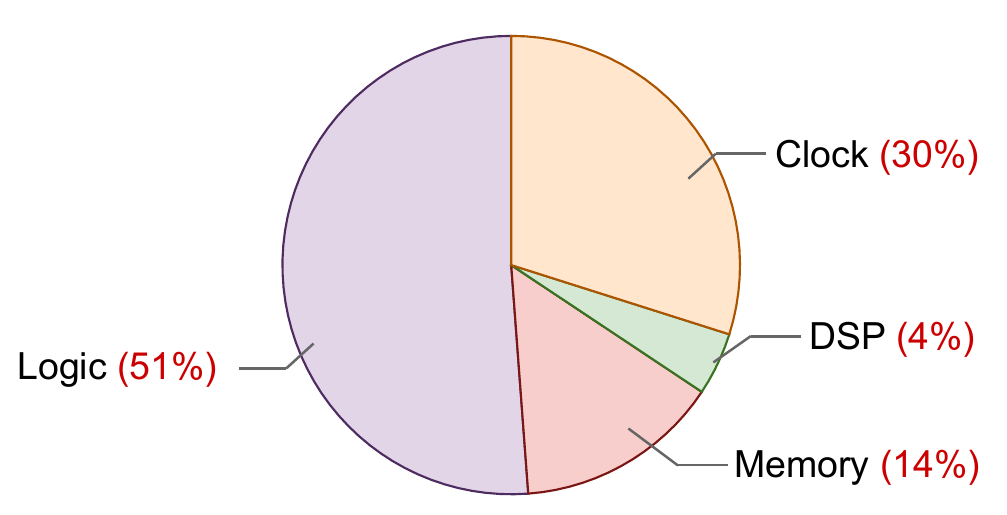}
    \caption{ }
  \end{subfigure}
    \begin{subfigure}[b]{0.4\textwidth}
    \includegraphics[width=\textwidth]{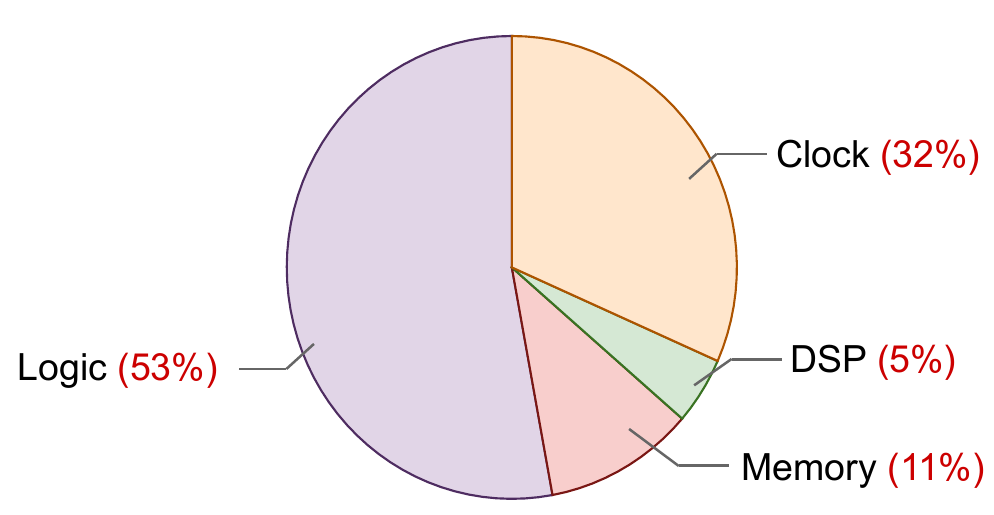}
    \caption{}
  \end{subfigure}

  \caption{Power breakdown of RAMAN for (a) MobileNetV1$^{\dagger}$ model and (b) DS-CNN$^{\dagger}$ model for $75\%$ weight pruning in the PW layers. }
  \label{fig:pwr_breakdown}
\end{figure*}

\par The memory and power breakdown of the RAMAN architecture is shown in Fig.\ref{fig:mem_breakdown} and Fig.\ref{fig:pwr_breakdown} considering $75\%$ weight pruning in the PW layers. The parameter memory ($54\%$ util.) dominates the activations ($26\%$ util.) for the MobileNetV1$^{\dagger}$ model, whereas the activation memory ($46\%$ util.)  dominates the parameters ($25\%$ util.) for the DS-CNN$^{\dagger}$ model. The cache and Instruction Memory+ASE buffers account for $16-23\%$  and $4-6\%$, respectively. The power breakdown is as follows; logic+clock account for $80\%$, and the memory+DSP account for the remaining $20\%$.

\subsection{Model-wise Analysis:}
The number of operations and memory accesses of DS-CNN$^{\dagger}$ and MobileNetV1$^{\dagger}$ vary significantly based on the sparsity of the given input. Also because the model statistics (as shown in Table \ref{model_stats_table}) vary significantly between both the models, it is important to analyze RAMAN's performance for both the models separately.

\begin{table*}[]
\caption{Performance Breakdown of DS-CNN$^{\dagger}$ model with 8 DWS layers. The RAP threshold was set to zero and the parameter pruning in the PW layers was set to 50\% }
\begin{tabular}{|c|c|c|c|c|c|c|c|}
\hline
\textbf{Layer}  & \textbf{\begin{tabular}[c]{@{}c@{}}Activation GLB-MEM \\ accesses (KB)\end{tabular}} & \textbf{\begin{tabular}[c]{@{}c@{}}Parameter GLB-MEM\\ accesses (KB)\end{tabular}} & \textbf{\begin{tabular}[c]{@{}c@{}}Activation Cache \\ accesses (KB)\end{tabular}} & \textbf{\begin{tabular}[c]{@{}c@{}}Parameter Cache\\ accesses (KB)\end{tabular}} & \textbf{\begin{tabular}[c]{@{}c@{}}OPs \\ (in Mil.)\end{tabular}}  & \textbf{\begin{tabular}[c]{@{}c@{}}Num. of \\Active PEs\end{tabular}}  & \textbf{\begin{tabular}[c]{@{}c@{}}Avg. IA Sparsity\\ (\%)\end{tabular}} \\ \hline
\textbf{CONV} & 54.72                                                                          & 0.96                                                                           & 0.651                                                                            & 0                                                                                & 0.172                                                                                                                            & 11 (91.67\%)     & 82.2                                                                     \\ \hline
\textbf{DW}     & 204.544                                                                          & 7.68                                                                         & 247.684                                                                               & 0                                                                                & 0.849                                                                                                                                & 11 (91.67\%)  & 22.7                                                                     \\ \hline
\textbf{PW}     & 151.808                                                                            & 27.648                                                                          &  94.927                                                                             &   888.044                                                                         & 3.038                                                                                                                             & 12 (100\%)    & 37.5                                                                     \\ \hline
\textbf{Pooling}    & 1.088                                                                           & 0                                                                               & 0                                                                                 & 0                                                                                & 0.001                                                                                                                               & -  & 72.5                                                                     \\ \hline
\textbf{FC}     & 0.076                                                                           & 1.056                                                                           & 0.192                                                                               & 0                                                                                & 0.002                                                                                                                              & 12 (100\%)    & 66.8                                                                     \\ \hline
\textbf{TOTAL}  &  412.236                                                                           & 37.344                                                                          & 343.454                                                                         &  888.044                                                                          & 4.062                                                                                                                               & -   & 28.6                                                                     \\ \hline
\end{tabular}
\label{dscnn_perf}
\end{table*}

\begin{table*}[h!]
\caption{Performance Breakdown of MobileNetV1$^{\dagger}$ model which has 13 DWS layers. The RAP threshold was set to zero and the parameter pruning in the PW layers was set to 50\%}
\begin{tabular}{|c|c|c|c|c|c|c|c|}
\hline
\textbf{Layer}  & \textbf{\begin{tabular}[c]{@{}c@{}}Activation GLB-MEM \\ accesses (KB)\end{tabular}} & \textbf{\begin{tabular}[c]{@{}c@{}}Parameter GLB-MEM\\ accesses (KB)\end{tabular}} & \textbf{\begin{tabular}[c]{@{}c@{}}Activation Cache \\ accesses (KB)\end{tabular}} & \textbf{\begin{tabular}[c]{@{}c@{}}Parameter Cache\\ accesses (KB)\end{tabular}} & \textbf{\begin{tabular}[c]{@{}c@{}}OPs \\ (in Mil.)\end{tabular}} & \textbf{\begin{tabular}[c]{@{}c@{}}Num. of \\Active PEs\end{tabular}} & \textbf{\begin{tabular}[c]{@{}c@{}}Avg. IA Sparsity\\ (\%)\end{tabular}} \\ \hline
\textbf{CONV} & 44.56                                                                           & 0.240                                                                        & 16.561                                                                             & 0                                                                                & 0.463                                                                                                                               & 11 (91.67\%) & 27.2                                                                     \\ \hline
\textbf{DW}     & 437.808                                                                           & 18.720                                                                        & 499.483                                                                            & 0                                                                                & 1.904                                                                                                                              & 11 (91.67\%)  & 35.1                                                                     \\ \hline
\textbf{PW}     & 554.528                                                                           & 157.536                                                                       & 345.147                                                                            & 3620.764                                                                         & 9.339                                                                                                                                & 12 (100\%) & 44.9                                                                     \\ \hline
\textbf{Pooling}    & 9.472                                                                             & 0                                                                               & 0                                                                                  & 0                                                                                & 0.009                                                                                                                                & - & 59.8                                                                     \\ \hline
\textbf{FC}     &  0.258                                                                             & 0.8                                                                          & 0.512                                                                                & 0                                                                                & 0.001                                                                                                                             & 12 (100\%)  & 7.1                                                                      \\ \hline
\textbf{TOTAL}  & 1046.626                                                                          & 177.296                                                                         & 861.704                                                                            & 3620.764                                                                        & 11.716                                                                                                                            & - & 39.5                                                                     \\ \hline
\end{tabular}
\label{mobilenet_perf}
\end{table*}

\subsubsection{DS-CNN$^{\dagger}$}
Table \ref{dscnn_perf} summarizes the performance of RAMAN for the DS-CNN$^{\dagger}$ model, which consists of eight DWS layers with 50$\%$ weight pruning in the PW layers. It is observed that the pre-processed audio input (cochleagram) fed to the CONV layer is highly sparse (82.2\%). The activation cache accesses total just 0.651KB (0.479KB reads and 0.171KB writes), as the cache memory doesn't store zeros. These accesses translate to 0.172M OPs which is ${\approx}80\%$ lower when IA Sparsity is not exploited. CONV employs the DW dataflow and doesn't require parameter cache accesses; thus, the parameters are directly accessed from the GLB-MEM. 
\par
DWS layers follow the CONV layer. From Table \ref{dscnn_perf}, it is evident that activation cache access is 247.684  KB,  which is ${\approx}21\%$ higher than the activation GLB-MEM accesses. With an average IA sparsity of $22.7\%$, there is a reduction of ${\approx}38\%$ (0.849 Mil.) in the operations when IA Sparsity is exploited.
\par The PW layer employs the RD dataflow, as discussed in the main document. The parameter cache access (888.044) dominates the GLB-MEM accesses (27.648KB), leading to effective temporal cache reuse of about 32x. There is a 3x reduction in operations with IA Sparsity (37.5\%) and $50\%$ weight pruning.
\par
The global average pooling  (GAP) layer reduces each input channel to a single value. In RAMAN, during the execution of the GAP layer, the entire PE array is gated to reduce dynamic power consumption, and the entire execution takes place in the PPM. GAP constitutes a tiny percentage of operations and memory accesses. 
\par
The FC layer leverages IA reuse; hence, we observe that the activation cache accesses are ${\approx}2.5\times$ more than the activation GLB-MEM accesses. The parameters are directly accessed from the GLB-MEM, bypassing the cache due to no reuse.
\par
A further ${\approx}10\%$  reduction in IA cache access and ${\approx}13\%$  reduction parameter cache reads were observed with run-time activation pruning with the threshold value of 30.

\subsubsection{MobileNetV1$^{\dagger}$}
MobileNetV1$^{\dagger}$ model has 13 DWS layers which account for higher memory access and operations as indicated by Table \ref{mobilenet_perf} in comparison with the DS-CNN$^{\dagger}$  model. The estimates shown are for $50\%$ weight pruning and leveraging IA sparsity. 
The DW and the PW layers dominate the GLB-MEM activation and parameter accesses, and the same trend is observed in the cache access. The weights are stored in the cache only in the PW dataflow leading to a temporal reuse of about 23x. Run-time activation pruning with the threshold value of 40 led to an additional $\approx8\%$ decrease in IA cache access and $\approx21\%$ reduction in parameter cache reads.

\bibliographystyle{IEEEtran}
\bibliography{ref.bib}